\documentclass{article}

\usepackage{microtype}
\usepackage{graphicx}
\usepackage{subfigure}
\usepackage{booktabs} 

\usepackage{hyperref}

\usepackage{amsmath}
\usepackage{amssymb}
\usepackage{dsfont}
\usepackage{caption}

\usepackage{xparse}
\renewcommand{\vec}[1]{\mathbf{#1}}
\renewcommand{\mid}{|}
\newcommand{\mat}[1]{\mathbf{#1}}
\newcommand*\diff{\mathop{}\!\mathrm{d}}
\NewDocumentCommand{\x}{g}{ \IfValueTF {#1} {x_{#1}} {\boldsymbol{x}} }
\NewDocumentCommand{\cx}{g}{ \IfValueTF {#1} {x_{c,#1}} {\boldsymbol{x}_c} }
\NewDocumentCommand{\ex}{g}{ \IfValueTF {#1} {x_{#1}} {\boldsymbol{x}} }
\NewDocumentCommand{\p}{g}{ \IfValueTF {#1} {p_{#1}} {\boldsymbol{p}} }
\NewDocumentCommand{\kx}{g}{ \IfValueTF {#1} {k_{#1}} {\boldsymbol{k}} }
\NewDocumentCommand{\CM}{g}{ \IfValueTF {#1} {C_{#1}} {\mat{C}} }
\NewDocumentCommand{\pr}{m g}{ p(#1 \IfValueTF {#2} {\mid #2}{} )}
\NewDocumentCommand{\prd}{m g}{ p(#1 \IfValueTF {#2} {\mid #2}{} )}
\NewDocumentCommand{\cpr}{m g}{ F(#1 \IfValueTF {#2} {\mid #2}{} )}
\NewDocumentCommand{\indicator}{m}{\mathds{1}_{#1}}
\NewDocumentCommand{\normalizedfrequency}{m m m}{\frac{(#1 + #2)_{#3}}{|| #1 + #2 ||_1}}

\NewDocumentCommand{\LS}{}{ \mathcal{L} }
\NewDocumentCommand{\US}{}{ \mathcal{U} }
\NewDocumentCommand{\ES}{}{ \mathcal{E} }

\DeclareMathOperator*{\EE}{\mathbb{E}}
\newcommand{\expec}[2]{\EE_{#1}\left[#2\right]}
\newcommand{\tpm}{$\pm$}

\newcommand{\cand}{\vec{x}_c}

\newcommand{\kc}{\vec{k}_{\cand}}

\newcommand{\1}{\vec{1}}

\NewDocumentCommand{\xgain}{}{ \mathrm{xgain} }
\NewDocumentCommand{\MultiNom}{m}{ \mathrm{Multinom} ( #1 ) }
\NewDocumentCommand{\Diriclet}{m}{ \mathrm{Dir}(#1) }
\NewDocumentCommand{\Cat}{m}{ \mathrm{Cat}(#1) }
\NewDocumentCommand{\prior}{g}{ \IfValueTF {#1} {\alpha_{#1}} {\boldsymbol{\alpha}} }
\NewDocumentCommand{\priorx}{g}{ \IfValueTF {#1} {\alpha_{#1}} {\boldsymbol{\alpha}} }
\NewDocumentCommand{\priorc}{g}{ \IfValueTF {#1} {\beta_{#1}} {\boldsymbol{\beta}} }

\DeclareMathOperator*{\E}{\mathbb{E}}
\newcommand{\Exp}[2]{\E_{#1} \left[#2\right]}

\DeclareMathOperator*{\argmax}{arg\,max}

\NewDocumentCommand{\kxn}{g}{ \IfValueTF {#1} {k_{#1}^{\LS^+}} {\boldsymbol{k}^{\LS^+}} }
\NewDocumentCommand{\normalizedfrequencyshort}{m m}{\frac{(#1)_{#2}}{|| #1 ||_1}}
\newcommand{\2}{\vec{2}}

\usepackage[textsize=small]{todonotes}
\usepackage{forloop}
\newcounter{loopc}

\newtheorem{definition}{Definition}
\newtheorem{theorem}{Theorem}

\usepackage{pifont}
\newcommand{\cmark}{\ding{51}}%
\newcommand{\xmark}{\ding{55}}%

\usepackage[accepted]{icml2020}

\begin{document}

\twocolumn[
\icmltitle{Toward Optimal Probabilistic Active Learning Using a Bayesian Approach}




\begin{icmlauthorlist}
\icmlauthor{Daniel Kottke}{ies}
\icmlauthor{Marek Herde}{ies}
\icmlauthor{Christoph Sandrock}{ies}
\icmlauthor{Denis Huseljic}{ies}
\icmlauthor{Georg Krempl}{uu}
\icmlauthor{Bernhard Sick}{ies}
\end{icmlauthorlist}

\icmlaffiliation{ies}{Intelligent Embedded Systems, University of Kassel, Germany}
\icmlaffiliation{uu}{Department of Information and Computing Sciences, Utrecht University, The Netherlands}

\icmlcorrespondingauthor{Daniel Kottke}{daniel.kottke@uni-kassel.de}

\icmlkeywords{Classification, Probabilistic Active Learning, Expected Error Reduction, Decision-theoretic Optimization}

\vskip 0.3in
]



\printAffiliationsAndNotice{}  

\begin{abstract} 
Gathering labeled data to train well-performing machine learning models is one of the critical challenges in many applications. Active learning aims at reducing the labeling costs by an efficient and effective allocation of costly labeling resources. In this article, we propose a decision-theoretic selection strategy that (1) directly optimizes the gain in misclassification error, and (2) uses a Bayesian approach by introducing a conjugate prior distribution to determine the class posterior to deal with uncertainties. By reformulating existing selection strategies within our proposed model, we can explain which aspects are not covered in current state-of-the-art and why this leads to the superior performance of our approach. Extensive experiments on a large variety of datasets and different kernels validate our claims.
\end{abstract}

\section{Introduction} \label{sec:introduction}

To train classifiers with machine learning algorithms in a supervised manner, we need labeled data. Whereas gathering unlabeled instances is easy, the annotation with class labels is often expensive, exhaustive, or time-consuming and needs, consequently, to be optimized. Active learning (AL) algorithms aim to reduce annotation costs efficiently and effectively \cite{settles2009active}. For that purpose, a selection strategy successively chooses the most useful labeling candidate from the pool of unlabeled instances and acquires the corresponding label from an oracle.

%
%

Our approach builds on three pillars: 
(1) We approximate the usefulness of one candidate on a representative subset, as mentioned in ``\textit{toward optimal} AL'' by \citet{roy2001eer}. 
(2) We estimate the usefulness by determining the decision-theoretic gain in performance, as mentioned in ``\textit{probabilistic} AL'' by \citet{kottke2016mcpal}. 
(3) We use a \textit{Bayesian approach} and introduce a conjugate prior distribution to calculate the predictive posterior distribution. Thereby, we consider the certainty of a classifier on its predictions \cite{murphy2006binomial}. 
As indicated in italic font, these pillars explain our choice of the title of this article. 

The contributions of this article are as follows:
\begin{itemize}
    \itemsep0pt
    \vspace{-1em}
    \item We propose a universal model for decision-theoretic AL, called xPAL, which calculates the gain in performance using a Bayesian approach.
    \item By simplifying our model, we prove equivalence to existing AL methods and show how this simplification affects the selection of candidates.
    \item Our experiments on 22 datasets confirm the superiority of our approach compared to several baselines and the robustness of our prior parameter.
\end{itemize}

The remainder of this article is structured as follows: First, we discuss related work in Sec.~\ref{sec:relatedwork}. In Sec.~\ref{sec:problem}, we define our problem and provide the foundations for our model. In Sec.~\ref{sec:method}, we propose our new method xPAL and show how it theoretically and empirically relates to state-of-the-art approaches in Sec.~\ref{sec:comparison}. We evaluate our results experimentally and discuss our key findings in Sec.~\ref{sec:experiments}. We close this article with a conclusion and an outlook on our future work in that field.

\section{Related Work} \label{sec:relatedwork}


The central component of an AL algorithm is the selection strategy. The most na\"ive one is to choose the next candidate randomly~\cite{settles2009active}. A common heuristic is uncertainty sampling~\cite{lewis1994uncertainty}. The idea is to use, e.\,g., the estimated class posteriors of probabilistic classifiers or the distance to the decision boundary to build a usefulness score~\cite{settles2012}. This \emph{exploits} the current classification hypothesis by labeling instances close to the decision boundary. 
In contrast to density-based approaches~\cite{NguyenSmeulders2004}, it ignores the representativeness of selected instances for the entire training set, and fails to perform \emph{exploration}~\cite{BonduLemaireBoulle2010,OsugiKimScott2005}. That is, it does not search the instance space for large regions with incorrect classifications.
This might lead to even worse performance compared to random sampling~\cite{settles2012}. Hence, there exist variants that add random sampling~\cite{zliobaite2014active,ThrunMoller1992}, use reinforcement learning~\cite{OsugiKimScott2005} or simulated annealing~\cite{ZollerBuhmann2000} to balance exploitation and exploration, or combine it with a density weight~\cite{donmez2007dual} and a variety of further factors, including sample diversity~\cite{WeiglEtal2015,XuAkellaZhang2007,Brinker2003} and class priors~\cite{CALMA201813}. 

Uncertainty sampling is a special case of adaptive submodular maximization~\cite{CuongLeeYe2014}, and several works have established links between submodularity and AL~\cite{CuongLeeYe2014,GolovinKrause2010,GuilloryBilmes2010}.
An example for a recent approach, built on these works, is filtered active submodular selection (FASS)~\cite{WeiIyerBilmes2015}. FASS combines  uncertainty sampling with a submodular data subset selection framework, capturing both sample informativeness and representativeness. 
For Gaussian Process classifiers, a Bayesian information theoretic AL approach is Bayesian Active Learning by Disagreement (BALD)~\cite{HoulsbyEtal2011}. BALD aims to select instances with the highest marginal uncertainty about the class label but simultaneously high confidence for the individual settings of the model's parameters. 

The query by committee (QBC) method~\cite{seung1992qbc} builds classifier ensembles and aims to reduce the disagreement between them. To improve balancing of exploration and exploitation in ensembles of active learners, \citet{BaramYanivLuz2004} proposed a formulation as a multi-armed bandit problem. Here, each active learner corresponds to one slot machine whose relative progress in performance is tracked over time, and on each trial one active learner is chosen for selecting an instance using the EXP4 algorithm.
Furthermore, reinforcement learning approaches have been proposed that learn a policy for selecting active learners, for example by modelling active learning as a Markov decision process~\cite{KonyushkovaSznitmanFua2018}.

In 2001, \citet{roy2001eer} proposed expected error reduction. As shortly addressed in the introduction, they aim to estimate the expected generalization error if a candidate gets an additional label. Thus, they simulate each label for each labeling candidate and evaluate the mean error using the unlabeled instances. To estimate the probabilities, they use the class posteriors provided by probabilistic classifiers. \citet{chapelle2005active} noticed that these estimates are highly unreliable (esp. at the beginning of the training) and therefore suggested the use of a beta prior. 

\citet{kottke2016mcpal} address the issue pointed out by Chapelle and named their approach probabilistic AL. They propose to use a distribution of the class posterior probability instead of using the classifier outputs directly. Calculating the expectation over this posterior leads to a decision-theoretic approach that gets rid of the parameter of Chapelle and leads to a mathematically sound approach. 

\section{Problem Formulation and Foundations}
\label{sec:problem}

In ``The Nature of Statistical Learning Theory,'' \citet{Vapnik1995} introduced a holistic concept on how to \textbf{learn from examples}. He defined three different components that take part in such a process, namely a generator, a supervisor, and a learning machine.\footnote{We adapt the terms and notation slightly. We use calligraphy for sets, bold font for vectors, and $p( \cdot )$ is either the probability density function or the probability mass of a discrete probability space. Please note the difference between $\p$ and $p(\cdot )$ (the latter is always a function). } The \textit{generator} creates random vectors $\x \in \mathbb{R}^D$ (D-dimensional feature space) independently drawn from a fixed but unknown probability distribution $\pr{\x}$. The \textit{supervisor} provides class labels $y \in \mathcal{Y} = \{1, \dots, C\}$ ($C$ is the number of classes) for every instance $\x$ according to a conditional distribution $\pr{y|\x}$ which is also fixed but unknown. In our case, the learning machine is a \textit{classifier} $f_{\boldsymbol{\theta}}(\x)$ with some parameters $\boldsymbol{\theta}$. The goal is to choose that learning machine that approximates the supervisor's response best.

We adopt the above definition for the active learning scenario by refining the role of the (omniscient) supervisor:
\begin{definition}[Supervisor]\label{def:supervisor}
    A supervisor consists of:
    \vspace{-1em}
	\begin{enumerate}
	    \itemsep0pt
		\item A \emph{ground truth} which is an unknown but fixed, deterministic function $t \colon \mathbb{R}^D \rightarrow [0,1]^C$ that maps an instance~$\x$ to a probability vector $\p = t(\x)$ with $\sum_{i = 1}^C \p{i} = 1$. 
		Each element describes the true probability for the corresponding class given the instance $\x$.
		\item An \emph{oracle} which provides a class label ${y \in \mathcal{Y}}$ for every instance $\x$ according to the ground truth ${\p = t(\x)}$. Hence, the label is sampled from a categorical distribution $y \sim \Cat{t(\x)}$.
	\end{enumerate}
\end{definition} 

\begin{figure}[h]
	\centering
	\includegraphics[scale=.8]{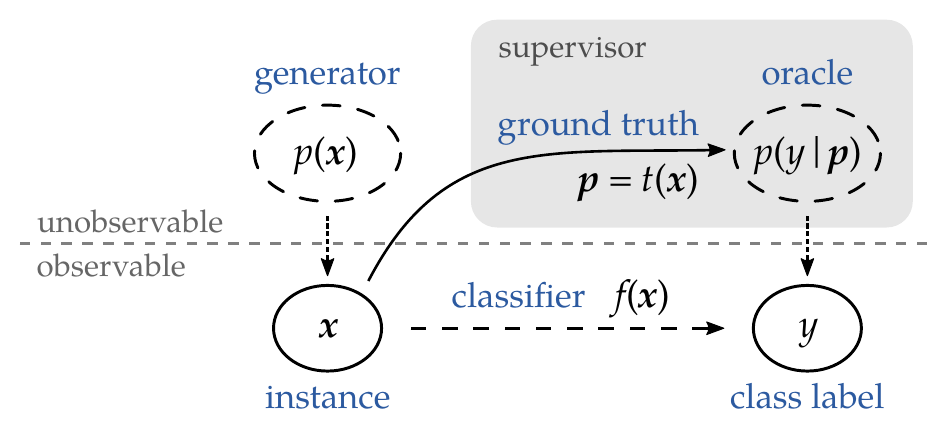}
	\caption{A Schematic illustration of how to learn from examples.
	}
	\label{fig:learn-from-ex}
\end{figure}

We visualize the learning process in Fig.~\ref{fig:learn-from-ex}. The generator provides instances $\x$ for which the oracle provides the class label $y$ based on the ground truth $t(\x) = \p = (p_1, \dots, p_C)$.  Unfortunately, we solely have information about the instance-label-pair $(\x,y)$ but not on the generator, the ground truth, or the oracle.

\newpage
In the technical community, the process of data generation is often described from a \textit{model-driven perspective}: Then it is assumed that each class $y$ has its own data generator $\pr{\x}{y}$. Hence, every instance $\x$ has exactly one label, which is also called ground truth. Due to noise during data generation, different classes might appear in the same region, but still, the true label exists. Our view (as given in Def.~\ref{def:supervisor} and Fig.~\ref{fig:learn-from-ex}) is purely \textit{data-driven}: Looking at the data, we do not know why there are different labels in the same region. It could be due to noise in the data generation or due to the imperfectness of the oracle. When learning a classifier, the reason does not matter: We only observe that the oracle provides different labels for similar instances according to some proportion $\p$ which we call \textit{ground truth}.

In the field of \textbf{active learning}, we assume to have an unlabeled dataset $\mathcal{U} = \{\x_1, \dots, \x_N\}$ (candidate pool) given by the generator. Labels are usually not available at the beginning but can be acquired from the oracle~\cite{settles2009active}, which chooses the label according to the ground truth. 

A \textbf{selection strategy} selects an instance $\x \in \mathcal{U}$, and we acquire the corresponding label $y \in \mathcal{Y}$ from the oracle. We remove the newly labeled instance from the candidate pool $\mathcal{U} \gets \mathcal{U} \setminus \{\x\}$, add the instance-label-pair to the labeled set $\mathcal{L} \gets \mathcal{L} \cup \{(\x,y)\}$, and retrain the classifier on $\mathcal{L}$.

We use a kernel-based classifier with \textbf{kernel} $K$ which describes the similarity of two instances $\x$ and $\x'$. 
In our experiments, we use three different kernels (see Sec.~\ref{sec:experiments}) but our method is not restricted to these kernels.

\begin{definition}[Kernel Frequency Estimate]\label{def:kernel-frequency-estimate}
    The kernel frequency estimate $\kx_{\x}^{\LS}$ of an instance $\x$ is determined using the set of labeled instances $\LS$. The $y$-th element of that $C$-dimensional vector describes the similarity-weighted number of labels of class $y$:\footnote{$\indicator{cond}$ denotes the indicator function which returns $1$ if $cond$ is true and $0$ otherwise. }
    \begin{align}
        \kx{\x,y}^{\LS} = \sum_{(\x', y') \in \LS} \indicator{y = y'} K(\x, \x') .
    \end{align}
\end{definition}

We denote $f^\LS$ as a \textbf{classifier} which uses the labeled data $\LS$ for training.\footnote{To simplify the notation, we do not mention the parameters~$\boldsymbol{\theta}$.} Similar to the Parzen Window Classifier (PWC) used in \citet{chapelle2005active}, the classifier $f^\LS$ predicts the \textbf{most frequent class}:
\begin{align}
    f^\LS(\x) = \argmax_{y \in \mathcal{Y}} \left( \kx{\x,y}^{\LS} \right).
\end{align}

Our method requires estimating kernel frequencies which is straight-forward for the PWC but also possible for other classifiers. For example, \citet{Beyer2015} estimates kernel frequencies (called label statistics) for Naive Bayes, $k$-Nearest Neighbour, and Tree-Based classifiers.

\section{Toward Optimal Probabilistic Active Learning using a Bayesian Prior} \label{sec:method}

The idea of our approach is to estimate the expected performance gain that a new instance would provide if we would acquire its label from the oracle. Then, we select the most promising instance for actual labeling. Within the next subsections, we explain the necessary steps towards the final method.

\subsection{Estimating the Risk}
In this article, we use the misclassification error as our performance measure (this can easily be changed). To optimize this performance, we minimize the estimated risk using the zero-one loss similarly to \citet{Vapnik1995}.

\begin{definition}[Risk, Zero-one Loss]\label{def:risk-loss}
    The risk describes the expected value of the loss $L$ with respect to the joint distribution $\pr{\x,y}$ given a classifier $f^\LS$:
    \begin{align}
    	R(f^\LS)
    	&= \Exp{\pr{\x,y}}{L(y, f^{\LS}(\x))}\\
    	&= \Exp{\pr{\x}}{
    		    \Exp{\pr{y \mid \x}}{
    			    L(y, f^{\LS}(\x))
    			}
    	    } .
    \end{align}
    The zero-one loss returns $0$ if the prediction of the classifier $f^\LS(\x)$ is equal to the true class $y$ and $1$ otherwise:
    \begin{align}
        L(y, f^\LS(\x)) = \indicator{f^\LS(\x) \neq y} .
    \end{align}
\end{definition}
As the generator $\pr{\x}$ is not observable, we use a Monte-Carlo integration using a set of instances $\ES$ which is able to represent the generator. For simplicity, we use the complete set of available instances, i.\,e. the labeled and the unlabeled data ($\ES = \{\x \colon (\x,y) \in \LS\} \cup \US$). Following the notation of \citet{japkowicz2011evaluating}, we calculate the empirical risk $R_{\ES}$ as follows:
\begin{align}
	R_{\ES}(f^\LS)
	&= \frac{1}{|\ES|} \sum_{\x \in \ES}
		\Exp{\pr{y \mid \x}}{
			L(y, f^\LS(\x))
		}\\
	&= \frac{1}{|\ES|} \sum_{\x \in \ES}
		\sum_{y \in \mathcal{Y}}
			\pr{y \mid \x} 
			    L(y, f^\LS(\x)) \label{eq:emp-risk}
\end{align}


\subsection{Introducing a Conjugate Prior}
The conditional class probability $p(y|\x)$ from Eq.~\eqref{eq:emp-risk} depends on the ground truth $t$ which is unknown (see Fig.~\ref{fig:learn-from-ex}):
\begin{align}
	\pr{y}{\x} &= \pr{y}{t(\x)} \label{eq:label-posterior}
	 = \pr{y}{\p}
	 = \Cat{y \mid \p}
	 = \p{y}.
\end{align}
As a consequence, the probability $\pr{y}{\x}$ is exactly the $y$-th element of the unknown ground truth vector $\p$. We can use the nearby labels from $\LS$ (represented in $\kx_{\x}^\LS$, Def.~\ref{def:kernel-frequency-estimate}) to estimate the ground truth $\p$ as the oracle provides the labels according to $\p$ (see Fig.~\ref{fig:learn-from-ex}). With increasing number of labels, our estimate converges to the correct ground truth. For estimation, we use a Bayesian approach by determining the posterior predictive distribution, i.\,e.~calculating the expected value over all possible ground truth values~$\p$ (see \citet{murphy2006binomial} for details on predictive distributions): 
\begin{align} \label{eq:est-label-posterior}
	\pr{y}{\x} 
	&\approx \pr{y}{\kx_{\x}^\LS}  
	 = \Exp{\pr{\p}{\kx_{\x}^{\LS}}}{\p{y}}
	 = \int \pr{\p}{\kx_{\x}^{\LS}} \, \p{y} \diff \p .
\end{align}
To determine the posterior probability $\pr{\p}{\kx_{\x}^{\LS}}$ of the ground truth $\p$ at instance $\x$, we use Bayes' theorem in Eq.~\eqref{eq:bayes}. 
The likelihood $\pr{\kx_{\x}^\LS}{\p}$ is a multinomial distribution as each label $y$ has been drawn from $Cat(y | \p)$ (see Fig.~\ref{fig:learn-from-ex}).\footnote{Normally, the multinomial distribution only allows non-negative integers as observations. Hence, we use it as an analogy. As our probability is normalized, we can also calculate its density for real-valued observations $\kx_{\x}^\LS$.} We \textbf{introduce a prior} $\pr{\p}$ which we choose to be a Dirichlet distribution with parameter $\prior \in \mathbb{R}^C$ as this is the conjugate prior of the multinomial distribution. 
We choose an indifferent prior and 
set each element to the same value ($\prior{1} = \ldots = \prior{C} \in \mathbb{R}^{>0}$) such that none of the classes is favoured. 
Using this prior can be seen as adding $\prior{y}$ pseudo-instances to every class~$y$~\citep[p.~77]{bishop2006pattern}.
This means that in case of high values of $\prior$, we need many labeled instances (i.\,e., high frequency estimates $\kx_{\x}^\LS$) to get distinct posterior probabilities.

As we use the conjugate prior of the multinomial likelihood, there exists an analytic solution for the posterior which is a Dirichlet distribution~\cite{murphy2006binomial}.
\begin{align}
	\pr{\p}{\kx_{\x}^\LS} 
	&= \frac{\pr{\kx_{\x}^\LS}{\p} \pr{\p}}{\pr{\kx_{\x}^\LS}} \label{eq:bayes}\\
	&= \frac{\MultiNom{\kx_{\x}^\LS \mid \p} \cdot \Diriclet{\p \mid \prior}}
			{\int \MultiNom{\kx_{\x}^\LS \mid \p} \cdot \Diriclet{\p \mid \prior} \diff \p} \\
	&= \Diriclet{\p \mid \kx_{\x}^\LS + \prior}
\end{align}
Now, we determine the conditional class probability $\pr{y}{\kx_{\x}^\LS}$ from Eq.~\eqref{eq:est-label-posterior} by calculating the expected value of the Dirichlet distribution~\cite{murphy2006binomial}:
\begin{align}
	&\pr{y}{\kx_{\x}^\LS} 
	 = 
	   \Exp{\Diriclet{\p | \kx_{\x}^\LS + \prior}}{\p{y}} \\
	&\qquad = \int \Diriclet{\p \mid \kx_{\x}^\LS + \prior} \, \p{y} \diff \p 
	= \normalizedfrequency{\kx_{\x}^\LS}{\prior}{y} . \label{eq:class-cond-prob-prior}
\end{align}
The last term describes the $y$-th element of the normalized vector $\kx_{\x}^\LS + \prior$. For normalization, we use the sum of all elements denoted as the 1-norm $|| \cdot ||_1$.

\subsection{Risk Difference Using the Conjugate Prior}

We insert Eq.~\eqref{eq:class-cond-prob-prior} into the empirical risk (Eq.~\eqref{eq:emp-risk}). As we approximate $\pr{y}{\x}$ with $\pr{y}{\kx_{\x}^\LS}$, this is an approximation of the empirical risk based on the labeled data $\LS$. Hence, we add $\LS$ as an argument of the estimated empirical risk:
\begin{align}
\hat{R}_{\ES}(f^{\LS}, \LS)
 = \frac{1}{|\ES|} \sum_{\x \in \ES}
			\sum_{y \in \mathcal{Y}}
				\normalizedfrequency{\kx_{\x}^\LS}{\prior}{y} L(y, f^\LS(\x)).
\end{align}

We now assume that we add a new labeled candidate $(\x_c, y_c)$ to the labeled set $\LS$ and denote the new set ${\LS^+ = \LS \cup \{(\x_c, y_c)\}}$. To determine how much this new instance-label-pair improved the performance of our classifier $f$, we estimate the gain in terms of risk difference under the current observations $\kx_{\x}^{\LS^+}$:
\begin{align}
    &\Delta \hat{R}_{\ES}(f^{\LS^+}, f^{\LS}, \LS^+) =
    \hat{R}_{\ES}(f^{\LS^+}, \LS^+) - \hat{R}_{\ES}(f^{\LS}, \LS^+)
\end{align}
\vspace{-2em}
\begin{align}
    \begin{split}
        &= \frac{1}{|\ES|} \sum_{x \in \ES} \sum_{y \in \mathcal{Y}} \normalizedfrequency{\kx_{\x}^{\LS^+}}{\prior}{y} \\
        &\quad \cdot \left(L(y, f^{\LS^+}(\x)) - L(y, f^\LS(\x))\right) .
    \end{split} \label{eq:risk-difference}
\end{align}

\subsection{The Expected Probabilistic Gain}

If we reduce the error under the new model $\LS^+$, the risk difference in Eq.~\eqref{eq:risk-difference} becomes negative. Therefore, we negate this term as we aim to maximize the gain in Def.~\ref{def:pgain}.



\begin{definition}[Expected Probabilistic Gain]\label{def:pgain}
	The probabilistic gain describes the expected change in classification risk $R$ when acquiring the label $y_c$ of candidate $\cx \in \US$. As the label $y_c$ and the corresponding ground truth $t(\cx)$ are unknown, we estimate $\pr{y_c}{\x_c}$ with $\pr{y_c}{\kx_{\x_c}^\LS}$ according to Eq.~\eqref{eq:class-cond-prob-prior} using $\Diriclet{\priorc}$ as prior. We write $\LS^+ = \LS \cup \{(\x_c, y_c)\}$.
	\begin{align}
		&\xgain(\cx ,\LS, \ES) = \expec{\pr{y_c}{\kx_{\x_c}^\LS}}{ - \Delta \hat{R}_{\ES}(f^{\LS^+}, f^{\LS}, \LS^+)} \\
		\begin{split} \label{eq:xgain}
    		&= -\sum_{y_c \in \mathcal{Y}} \normalizedfrequency{\kx_{\x_c}^{\LS}}{\priorc}{y_c}
    		\cdot \frac{1}{|\ES|} \sum_{\x \in \ES} \sum_{y \in \mathcal{Y}} \\ 
    		&\quad \normalizedfrequency{\kx_{\x}^{\LS^+}}{\priorx}{y} 
    		 \cdot \left(L(y, f^{\LS^+}(\x)) - L(y, f^\LS(\x))\right)
		\end{split}
    \end{align}
    For simplicity, we set $\priorc = \priorx$.
\end{definition}


We define the selection strategy xPAL to choose the candidate that optimizes the $\mathrm{xgain}$ score.
\begin{definition}[Selection Strategy: xPAL]\label{def:xpal}
	The selection strategy xPAL (Expected Probabilistic Gain for AL) chooses this candidate $\x_c^* \in \US$ with:
	\begin{align}
		\x_c^* = \argmax_{\x_c \in \US} \big( \xgain(\cx ,\LS, \ES) \big).
    \end{align}
\end{definition}

\section{Theoretical and Qualitative Comparison}
\label{sec:comparison}

\begin{figure*}[t!]
    \centering
    \includegraphics[width=.32\textwidth]{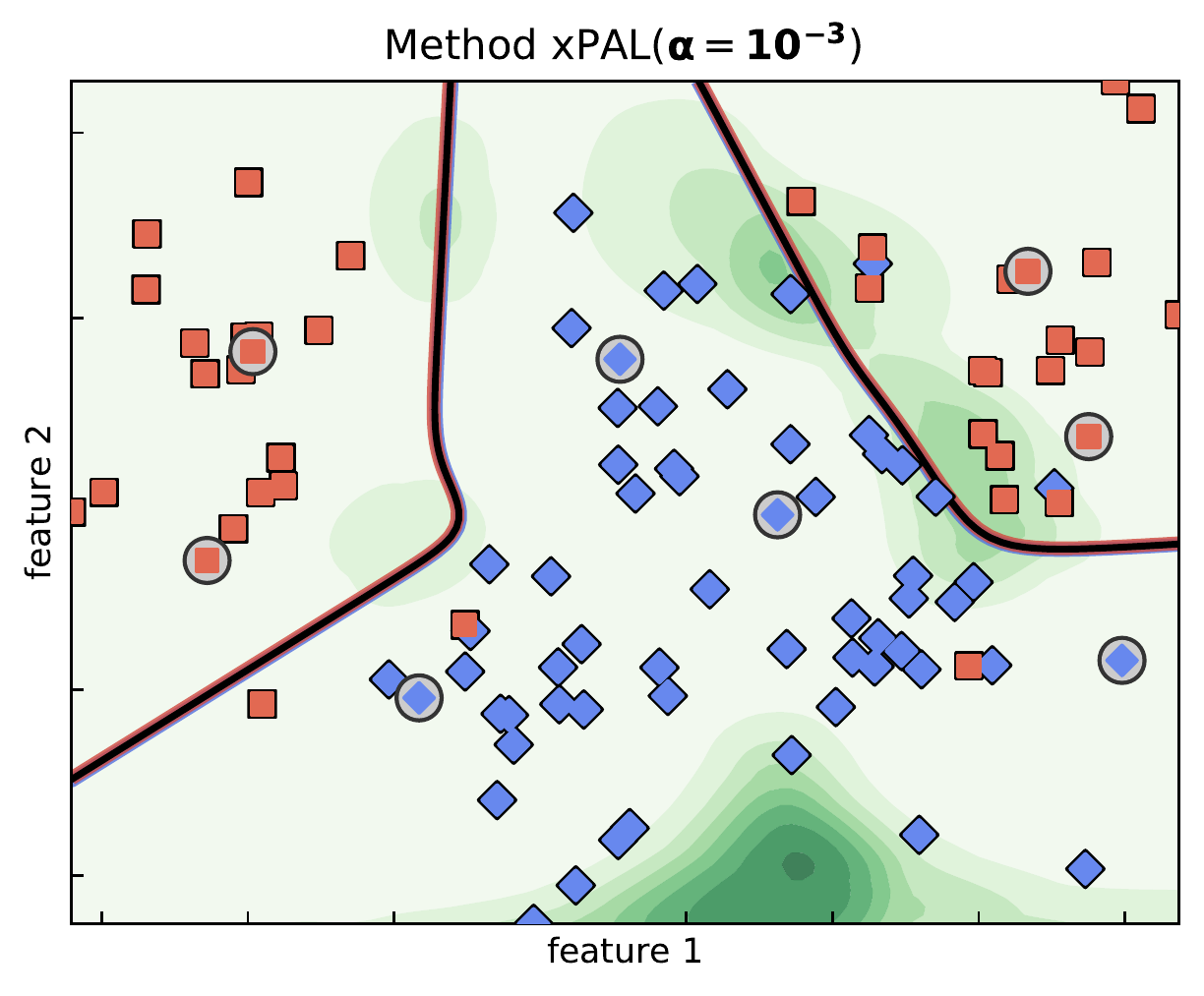}
    \includegraphics[width=.32\textwidth]{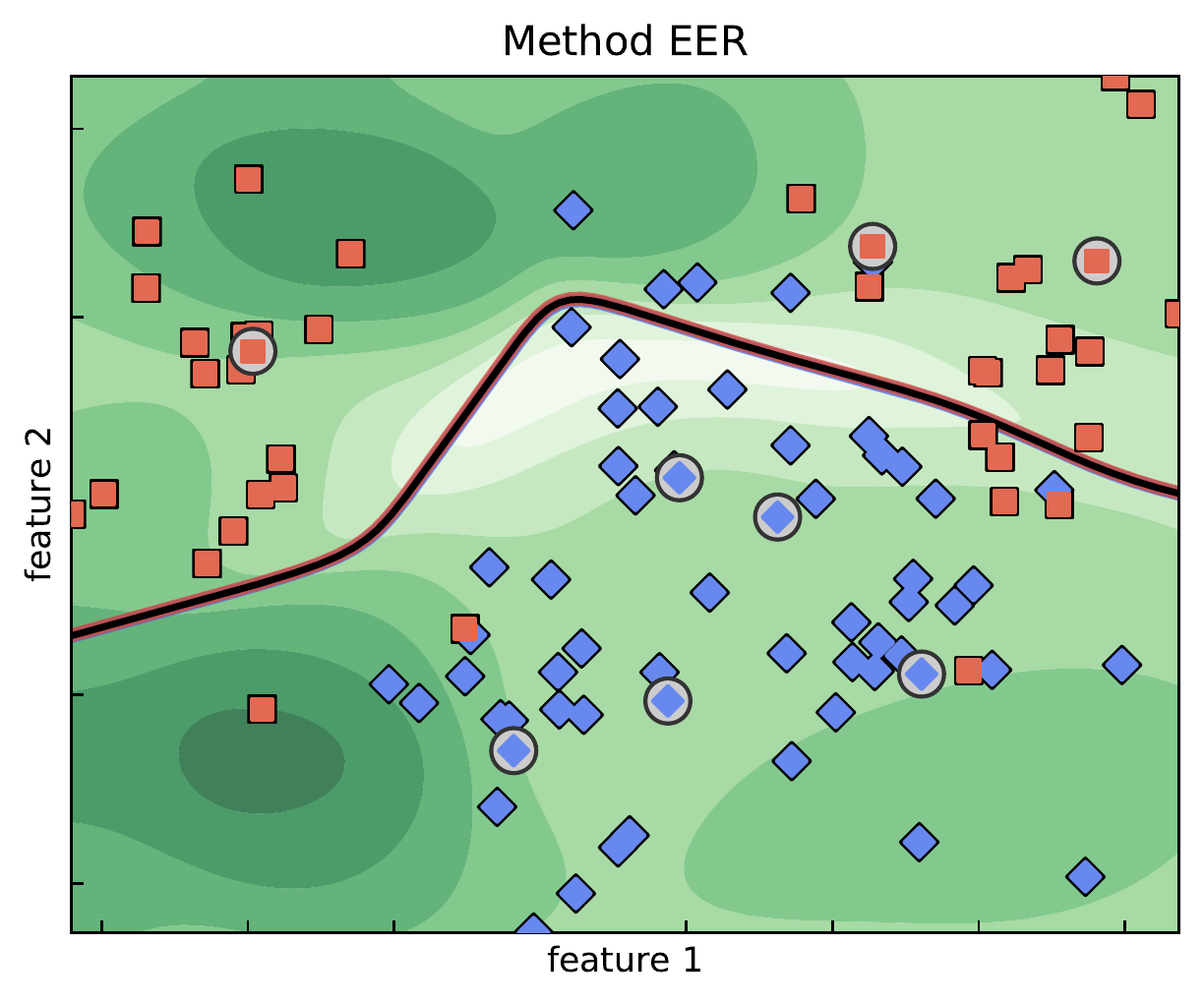}
    \includegraphics[width=.32\textwidth]{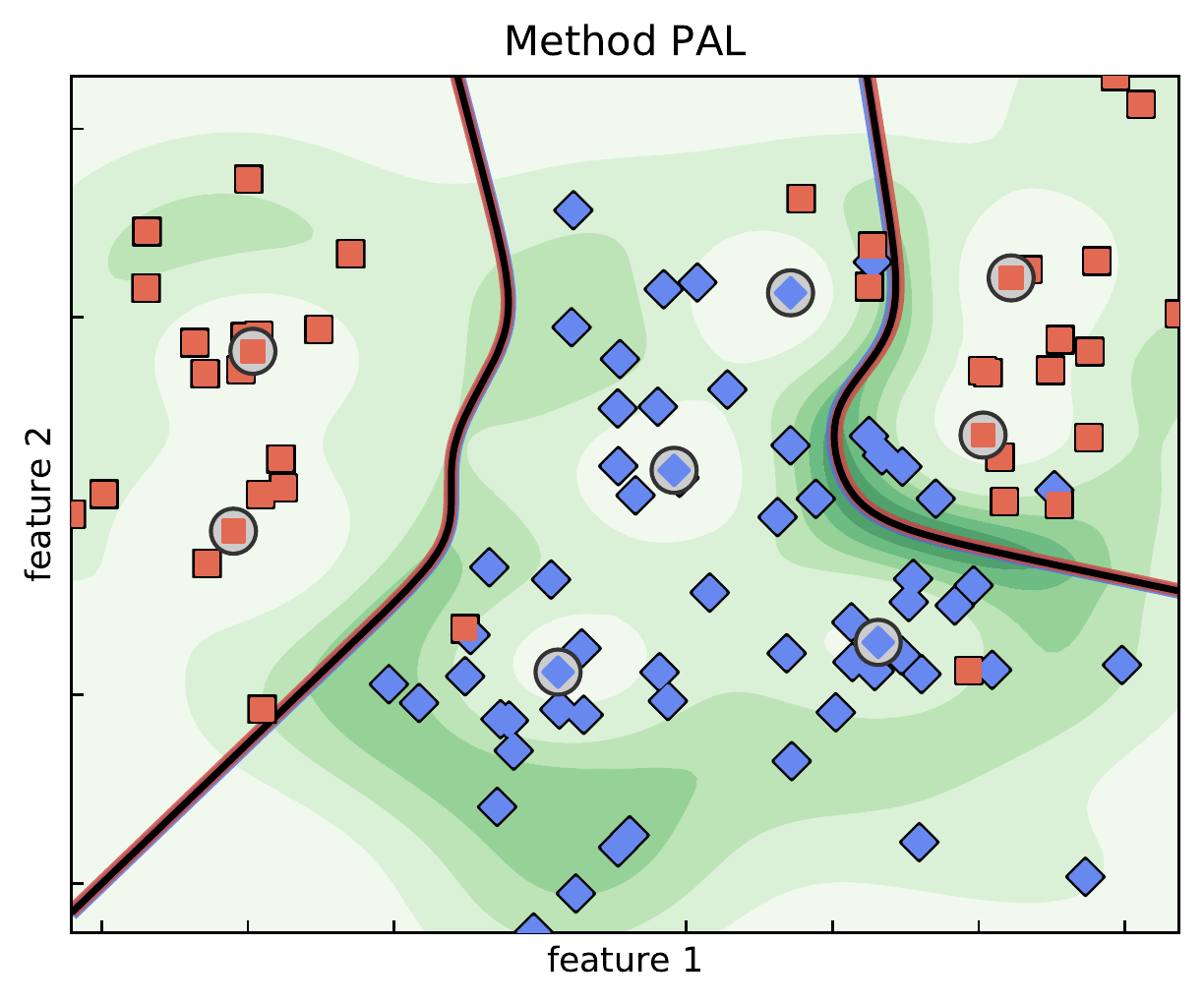}\\
    \medskip
    \includegraphics[width=.32\textwidth]{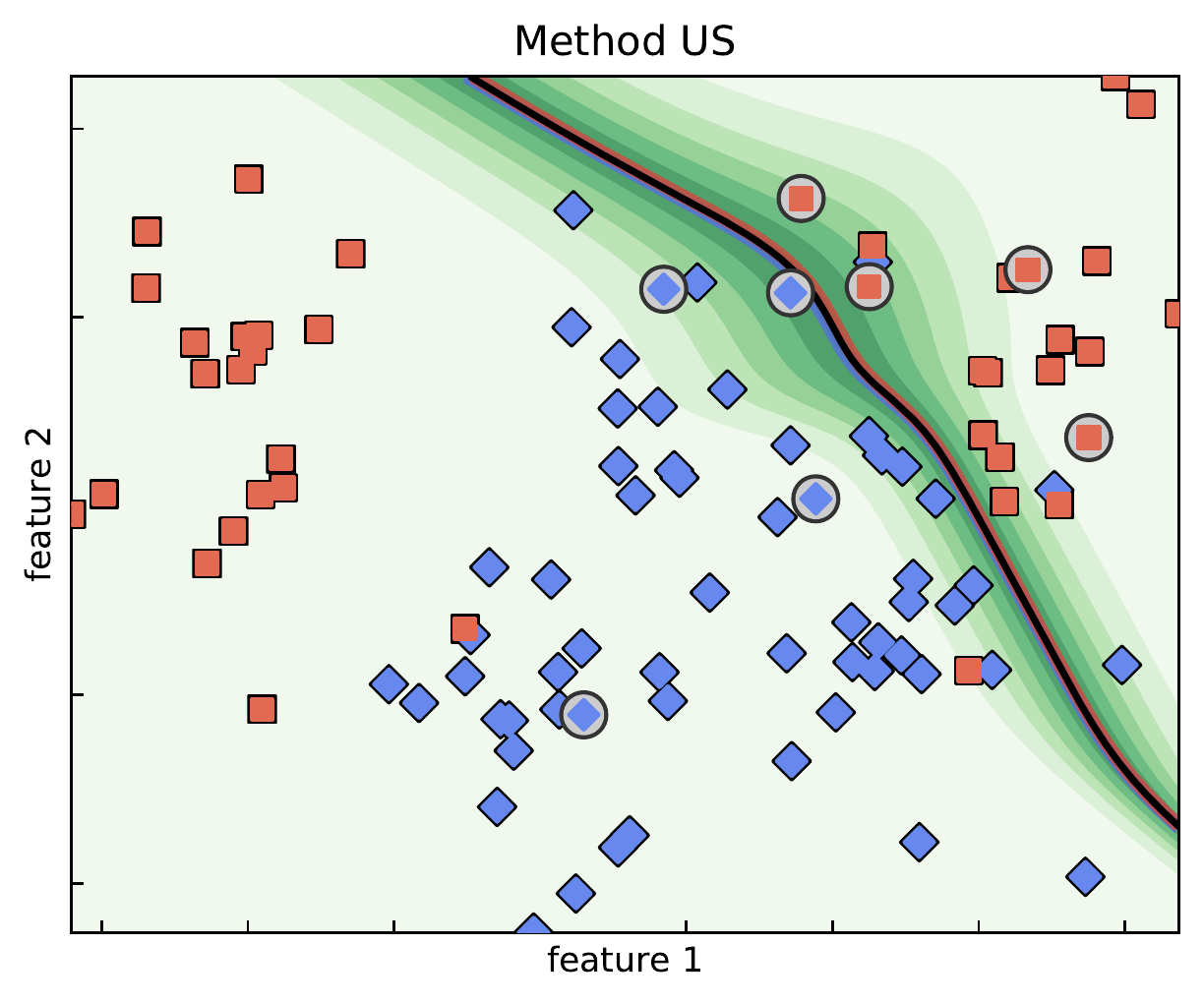}
    \includegraphics[width=.32\textwidth]{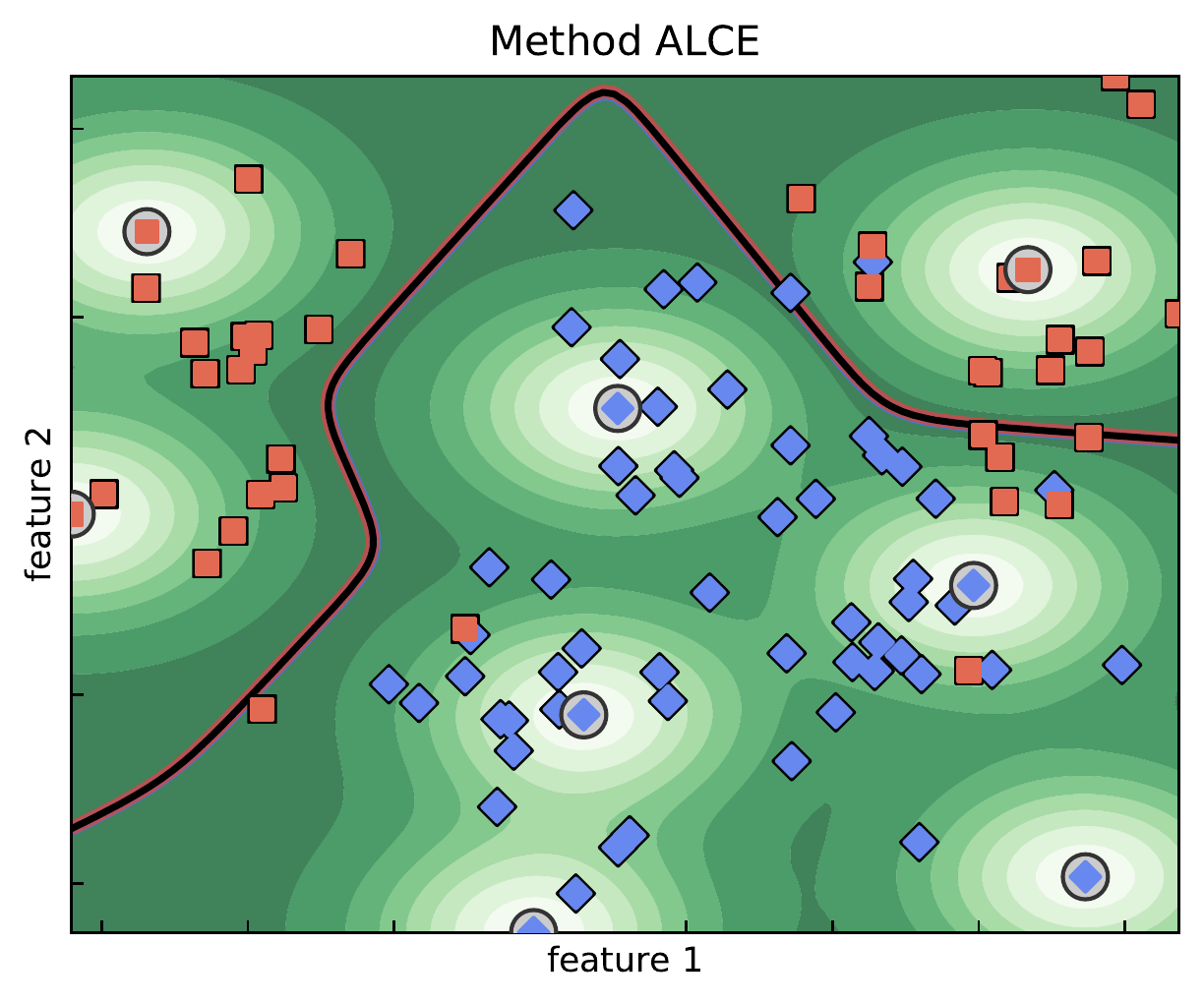}
    \includegraphics[width=.32\textwidth]{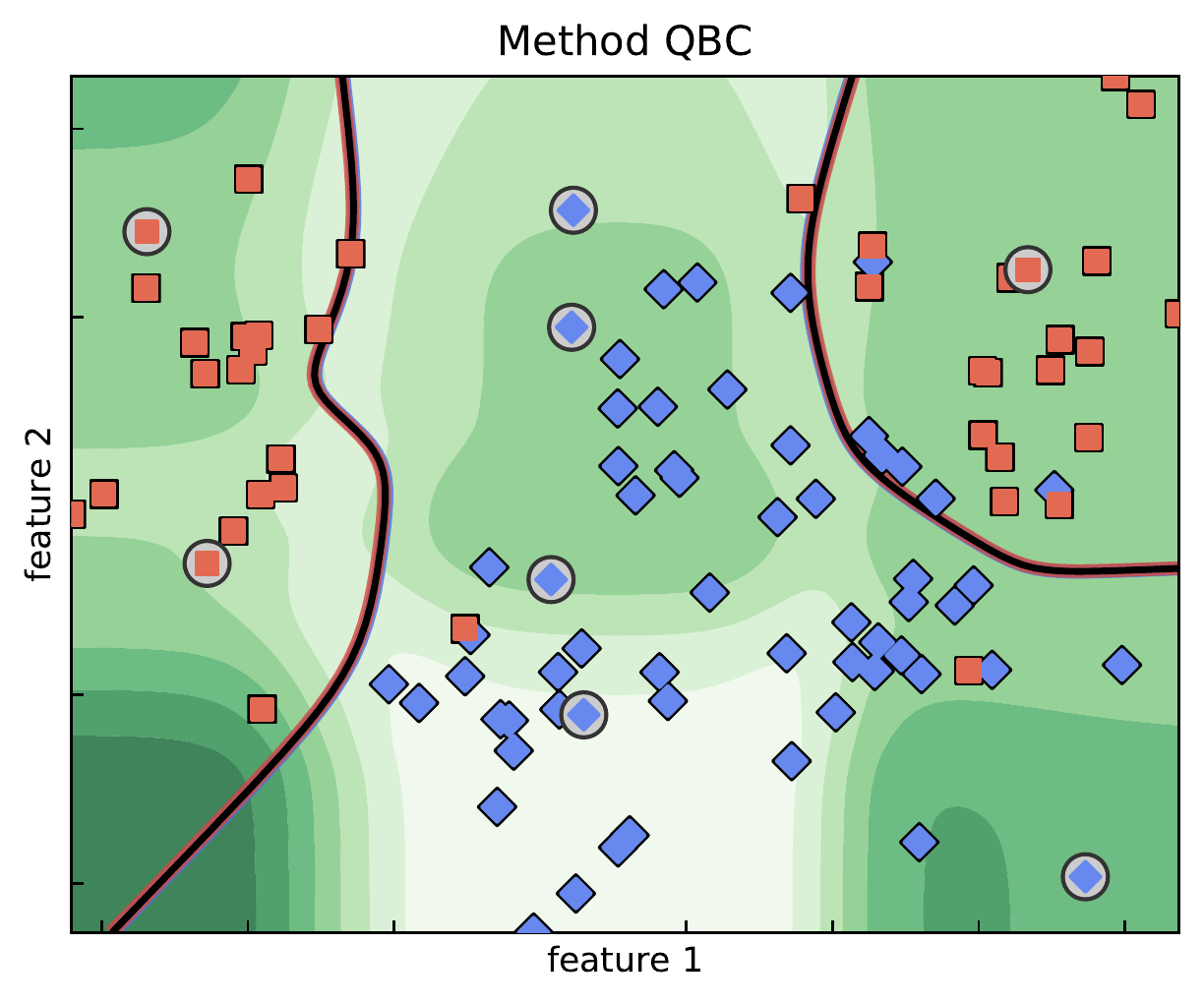}\\
    \medskip
    \includegraphics[width=.9\textwidth, trim=-3.7mm 0 -1.5mm 0]{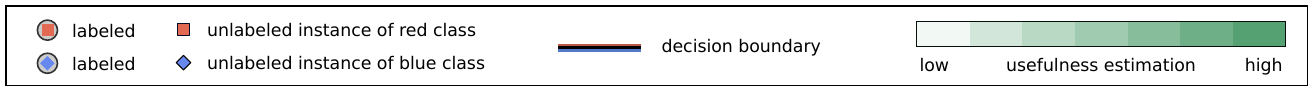}
    \caption{Visualization of acquisition behavior for different selection strategies. The green color indicates how useful a selection strategy considers a region. The usefulness depends on the selection criterion of the strategy. The eight labeled instances have been selected by the corresponding selection strategy. Thereby, one can see where the selection strategy selected instances in the past and how the usefulness is spatially distributed to select the next instance for labeling.}
    \label{fig:usefulness}
\end{figure*}

To provide an understanding of how the xPAL selection strategy works, we compare our new method to the most similar selection strategies by reformulating their approaches within our mathematical framework wherever possible. We provide the proofs for all theorems in the supplemental material.
In Tab.~\ref{tab:time_complexity}, we summarize the primary differences and show the computational complexity.

In Fig.~\ref{fig:usefulness}, we illustrate how the theoretical differences affect the actual choice of eight candidates on a toy dataset with two classes (blue diamonds and red rectangles).
For classification, we use the same setup as in Sec.~\ref{sec:experiments}. 
The first eight labeled instances, chosen by the selection strategy, are marked with a gray circle.
The background color shows how the respective selection strategy rates the usefulness of an area -- darker areas are considered more useful than brighter areas.

\subsection{Expected Probabilistic Gain for AL (xPAL)}

As seen in Fig.~\ref{fig:usefulness}, the currently labeled set $\mathcal{L}$ of xPAL is evenly spaced across the input space. 
That is, xPAL queried representative samples of the data set in the more explorative phase at the beginning, which leads to a rather good decision boundary with only eight labels.
Focusing on the current usefulness scores indicated by green background color, we see that regions close to the decision boundary and regions with very few labels (green area at the bottom) are preferred. Moreover, we notice more usefulness at the right decision boundary compared to the left one as this area is seen as being more relevant (due to the higher density).

\subsection{Expected Error Reduction (EER)}


\begin{theorem} The selection criterion of expected error reduction (EER) by \citet{roy2001eer} can be written as follows. The extension of adding a beta-prior $\vec{\epsilon}$ proposed by \citet{chapelle2005active} is given in blue color.
\begin{align}
        \begin{split} \label{eq:eer}
            \mathrm{eer}(\x_c, \LS, \US)
    		&= \sum_{y_c \in \mathcal{Y}} \normalizedfrequency{\kx_{\x_c}^{\LS}}{{\color{blue}\vec{\epsilon}}}{y_c}
    		\cdot \frac{1}{|{\color{orange}\US}|} \sum_{\x \in {\color{orange}\US}} \\
    		& \quad \sum_{y \in \mathcal{Y}}  \normalizedfrequency{\kx_{\x}^{\LS^+}}{{\color{blue}\vec{\epsilon}}}{y} 
    		    \cdot {\color{orange} L(y, f^{\LS^+}(\x)) }
		\end{split}
\end{align}
\end{theorem}

Comparing Eq.~\eqref{eq:eer} to Eq.~\eqref{eq:xgain}, we see that there are only a few differences highlighted in orange color. 
The main difference is the optimization objective as expected error reduction tries to query instances that minimize the expected error instead of the expected gain as in xPAL.
Second, EER neglects the labeled instances $\LS$ as it only uses $\US$ for Monte-Carlo integration. They assume that the unlabeled instances approximate the generator $p(\x)$ sufficiently well.
In the original version, \citet{roy2001eer} point out that the posterior estimates need to be reliable. Later, \citet{chapelle2005active} addresses this limitation by introducing a beta-prior~$\vec{\epsilon}$ (highlighted in blue), which serves a similar goal as our prior~$\prior$.

Although the theoretical differences of the two strategies are small, we see a clear difference in the acquired instances and in the usefulness estimation in Fig.~\ref{fig:usefulness}. Interestingly, the region close to the decision boundary is considered the least useful. Accordingly, EER neglects information there. 


\begin{table}[t!]
    \centering
    \caption{Summary of differences between xPAL and the four most similar methods evaluated on four criteria: (1) Is the usefulness estimated on a representative subset? (2) Does the method consider the performance gain? (3) Is some sort of prior included to handle uncertainties? (4) What is the asymptotic time complexity for determining the usefulness of one candidate sample?}
    \vskip 0.15in
    \begin{tabular}{lcccc}
        \toprule
        Method   & $\ES$ & Gain & Prior & $\mathcal{O}(\cdot)$\\
        \midrule
        xPAL     & \cmark & \cmark &  \cmark   & $|\ES| \cdot |\mathcal{Y}|^2$  \\
        PAL      & \xmark & \cmark &  \cmark   & $|\mathcal{Y}|^2$     \\
        EER      & (\cmark) & \xmark & (\cmark)  & $|\US| \cdot |\mathcal{Y}|^2$  \\
        US       & \xmark & \xmark &  \xmark   & $|\mathcal{Y}|$      \\
        \bottomrule
    \end{tabular}
    \label{tab:time_complexity}
\end{table}

\subsection{Probabilistic Active Learning (PAL)}

\begin{theorem} The selection criterion of (multi-class) probabilistic active learning (PAL) by \citet{kottke2016mcpal} can be written as follows.
\begin{align}
        \begin{split} \label{eq:pal}
            \mathrm{pal}(\x_c, \LS)
    		&= -{\color{orange}\hat{p}(\x_c}) \sum_{y_c \in \mathcal{Y}} \normalizedfrequency{\kx_{\x_c}^{\LS}}{{\color{orange}\vec{1}}}{y_c}
    	        \sum_{y \in \mathcal{Y}} \normalizedfrequency{\kx_{{\color{orange}\x_c}}^{\LS^+}}{{\color{orange}\vec{1}}}{y} \\
    		&\quad \cdot \left(L(y, f^{\LS^+}({\color{orange}\x_c})) - L(y, f^\LS({\color{orange}\x_c}))\right)
		\end{split}
\end{align}
\end{theorem}

The probabilistic active learning approach by \citet{kottke2016mcpal} does not consider a set $\ES$ for risk estimation but estimates the risk locally only for the candidate $\x_c$. Hence, we set $\ES = \{\x_c\}$. Instead, they include an estimated density weight $\hat{p}(\x_c)$ for their local gain. As a prior distribution, they use the indifferent prior $\vec{1}$. The original method is non-myopic. As xPAL is myopic, we ignored this for the theoretical discussion.



In general, we see a similar acquisition behavior of PAL and xPAL (see Fig.~\ref{fig:usefulness}). We see areas of high usefulness near the decision boundary and in sparely labeled regions. It seems that xPAL is more sensitive to the actual position of the instances as it considers the set $\ES$, and PAL only approximates this by using the density $\hat{p}(\x_c)$. Hence, the influence of a new label on the complete classification task is only approximated in PAL.


\subsection{Uncertainty Sampling (US)}
\begin{theorem} The selection criterion of confidence-based uncertainty sampling (US) by \citet{lewis1994uncertainty} can be written as follows.
\begin{align}
        \begin{split}
            \mathrm{us}(\x_c, \LS)
    		&= \sum_{y_c \in \mathcal{Y}} \normalizedfrequency{\kx_{\x_c}^{\LS}}{{\color{orange}\vec{0}}}{y_c} \cdot {\color{orange}L(y, f^\LS(\x_c))}
		\end{split}
\end{align}
\end{theorem}


Uncertainty sampling does not consider a set for risk estimation, but it solely estimates the error at the candidate $\x_c$ based on the current observations without any prior. Hence, it completely relies on the class posterior estimates from the classifier. Therefore, it might overestimate its certainty.

We observe this problem in Fig.~\ref{fig:usefulness} as US only finds one decision boundary and sticks at exploiting this. As it is not aware that the class posteriors on the left are highly unreliable (no labeled data here), it will only consider this region if the labels of all other candidates have been acquired. We notice a lack of exploration.



\newpage
\subsection{Active Learning with Cost Embedding (ALCE)}
The approach proposed by \citet{7837927} uses an embedding with some special distance measure in a hidden space with non-metric multidimensional scaling. As this follows an entirely different way of approaching the problem, it is not possible to transfer this algorithm to our framework. As shown in Fig.~\ref{fig:usefulness}, this approach explores the data space quite uniformly and is rather exploratory than exploitative.

\subsection{Query by committee (QBC)}
Query by committee \cite{seung1992qbc} uses an ensemble of classifiers that are trained on bootstrapped replicates of the labeled set $\LS$. 
With few labels, the strategy explores the dataset due to high randomness in the subsets (see Fig.~\ref{fig:usefulness}). Later, it starts exploiting more.


\newpage
\section{Experimental Evaluation} \label{sec:experiments}
To evaluate the quantitative performance of xPAL, we conduct experiments on real-world datasets.\footnote{Code: \url{https://github.com/dakot/probal}}
We provide information on the used datasets, algorithms, and the experimental setup.
We compare xPAL to state-of-the-art methods and show how the prior parameter affects the results.

\subsection{Datasets and Competitors}
We selected 27 datasets from the openML library~\cite{vanschoren2013openml} and two pre-processed text datasets from \citet{HernandezGonzales2018} with TF-IDF features. 
For the latter, we assigned the majority vote as the true class.
In the supplemental material, we list all used datasets with their openML-identifier and show specific characteristics such as the number of instances, features, and instances per class.

Next to xPAL, we use multi-class probabilistic AL (PAL) by \citet{kottke2016mcpal}, confidence-based uncertainty sampling (US) by \citet{lewis1994uncertainty}, active learning with cost embedding (ALCE) by \citet{7837927}, query by committee (QBC) by \citet{seung1992qbc}, expected error reduction (EER) by \citet{chapelle2005active}, and a random selector. We set all parameters according to the default values in the paper. For QBC, the disagreement within the randomly drawn sets, measured by the Kulback-Leibler divergence, describes the usefulness of a candidate. We use 25 classifiers as the committee and each of them is trained on a bootstrapped version of $\mathcal{L}$ with only a selection of features according to~\cite{Shi2008}. 

Additionally, we implemented a baseline that has additional access to \textit{all} labels of the unlabeled set $\US$. It successively (\textit{greedily}) selects the candidate, which minimizes the true empirical risk on $\US$ and $\LS$, called GREEDY-ALL.
It is equal to xPAL where the estimated class probability from Eq.~\ref{eq:label-posterior} is set to one for the true class.

\subsection{Experimental Setup}
To evaluate our experiments, we randomly split each dataset into a training set consisting of $60 \%$ of the instances and a test set containing the remaining $40 \%$ and repeat that $100$ times. 
As we start without any labeled instances, $\US$ contains the whole training set at the beginning, and $\LS$ is empty.
We acquire $200$ labels for every dataset or stop when $\US$ is empty.

For classification, we use the Parzen window classifier for all selection strategies. 
We applied three different kernels depending on the type of data. 
For numerical data, we z-standardize all features and use a radial basis function (RBF) kernel with bandwidth $\gamma$ which is defined as follows: 
\begin{align} \label{eq:kernel-sphere}
    K_{\text{rbf}}(\x, \x') = \exp\left( - \gamma || \x - \x' ||^2 \right).
\end{align}
We set the bandwidth of the kernel ($\gamma = 1/(2s^2)$) according to the mean criterion proposed by \cite{8215749} with $\sigma_p = 1$:
\begin{align}
    s = \sqrt{\frac{2N \sum_{j=1}^D \sigma_p^2}{(N-1) \ln{\frac{N-1}{\delta^2}}}}, \quad 
    \begin{split}
        \delta &= \sqrt{2} \cdot 10^{-6}, \\
        N &= \min\left( {|\US \cup \LS|, 200} \right).
    \end{split}
\end{align}

For categorical data, we use the hamming-distance kernel proposed by~\citet{Hutter2014} :
\begin{align} \label{eq:hamming-kernel}
    K_{\text{ham}}(\x, \x') = \exp\left( - \gamma \sum_{d=1}^{D} \indicator{x_d = x^\prime_d} \right),
\end{align}
where the hyperparameter $\gamma$ is again determined through the mean bandwidth criterion.

For the text datasets which contain TF-IDF features, we apply the cosine similarity kernel
\begin{align} \label{eq:cosine-kernel}
    K_{\text{cos}}(\x, \x') = \frac{\x^\mathrm{T} \cdot \x^\prime}{||\x||_2 \cdot ||\x^\prime||_2}.
\end{align}

\subsection{Comparison Between xPAL and Competitors}
\begin{figure*}[p!]
	\centering
	\includegraphics[width=.33\textwidth]{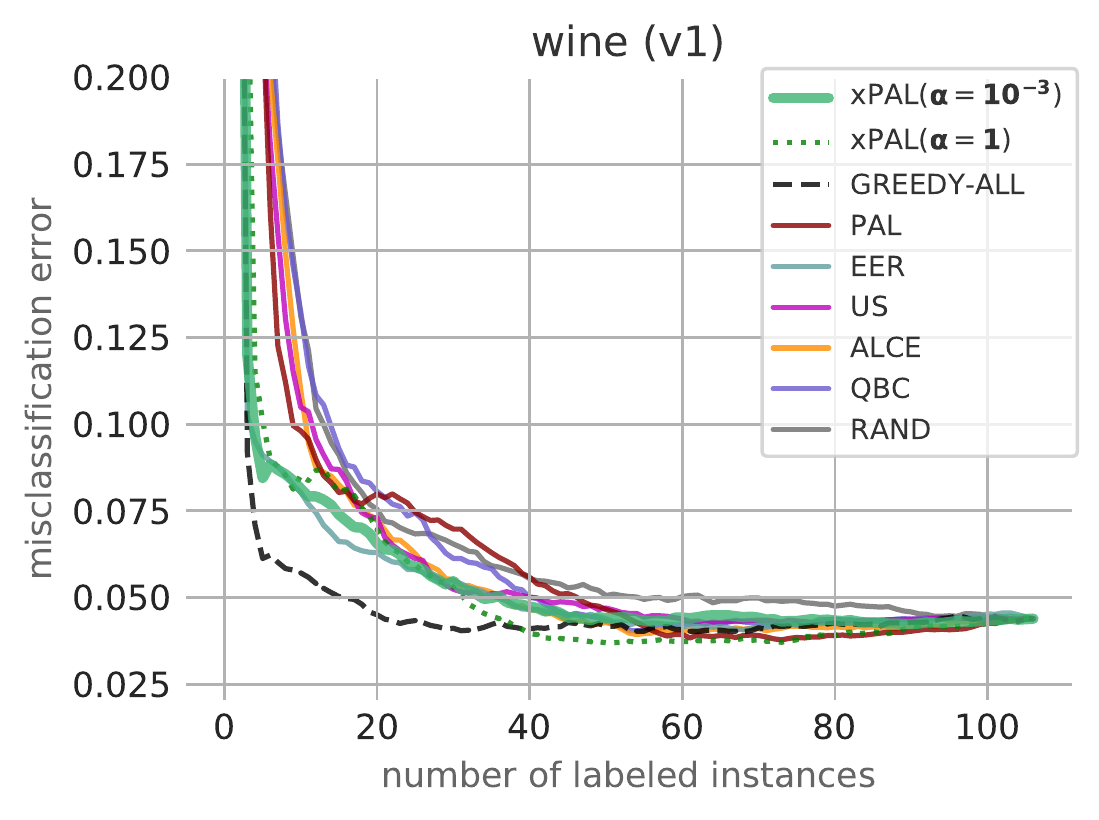}  %
	\includegraphics[width=.33\textwidth]{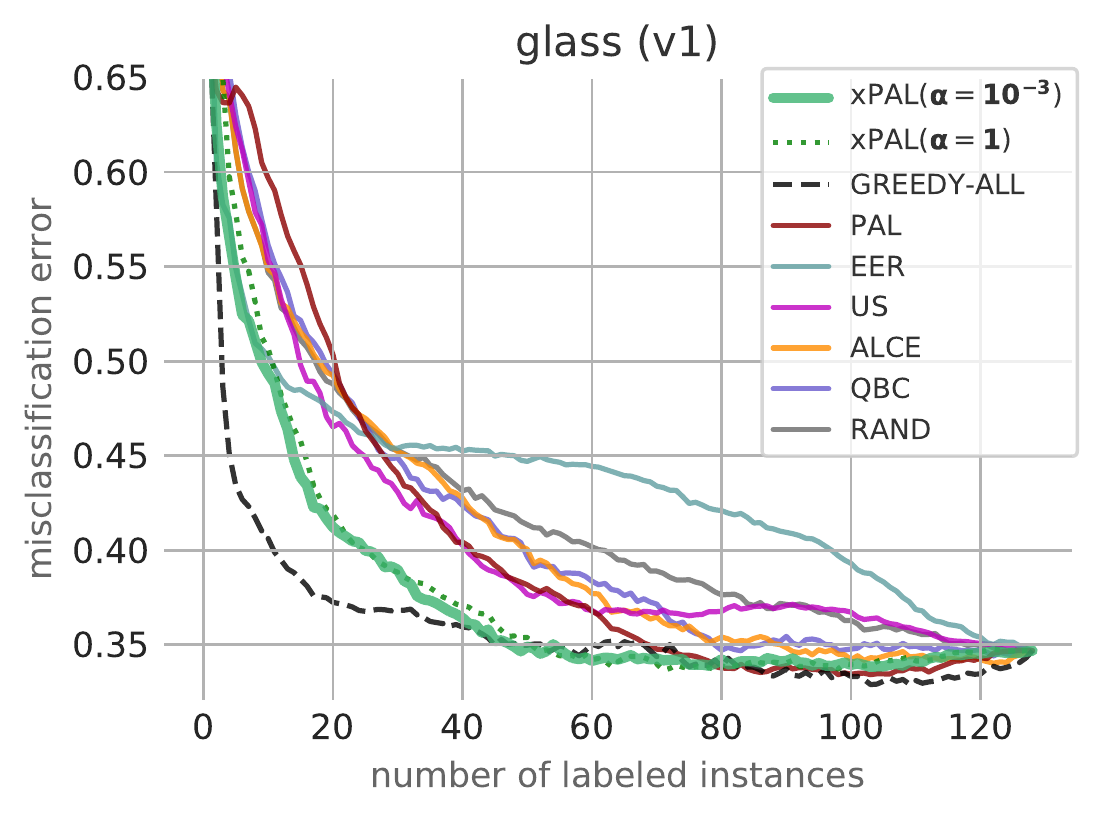}  %
	\includegraphics[width=.33\textwidth]{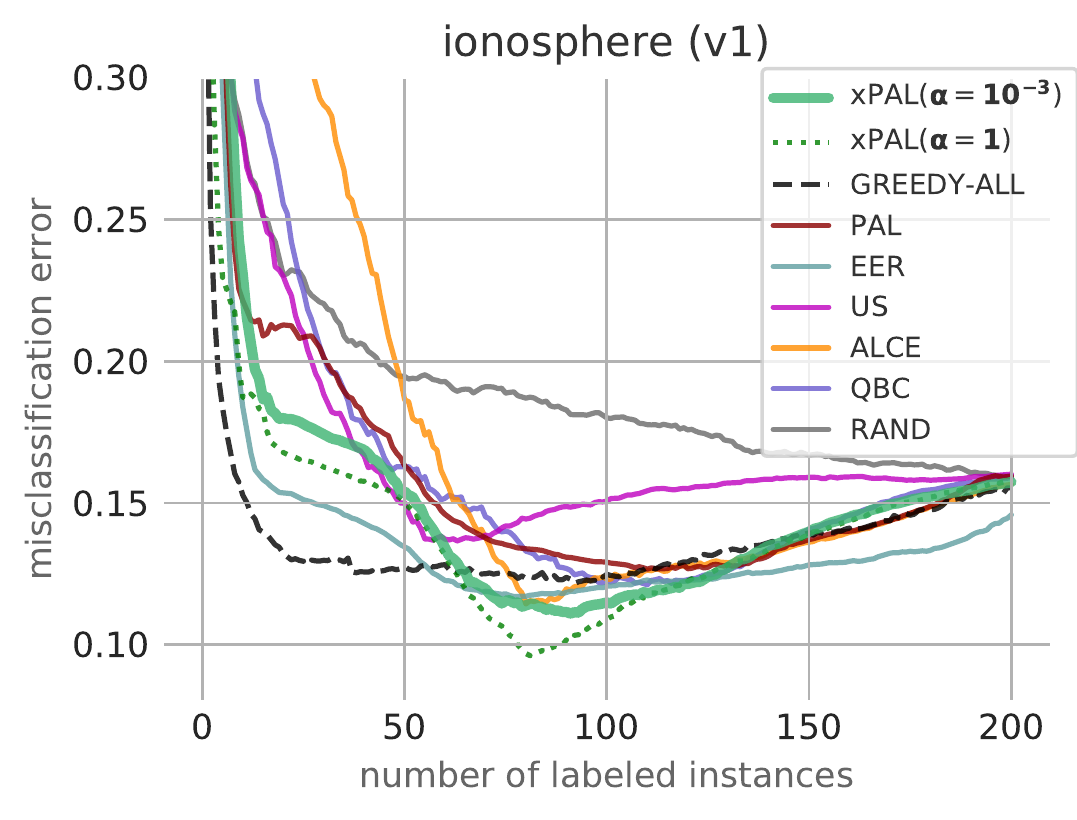}  %
	\includegraphics[width=.33\textwidth]{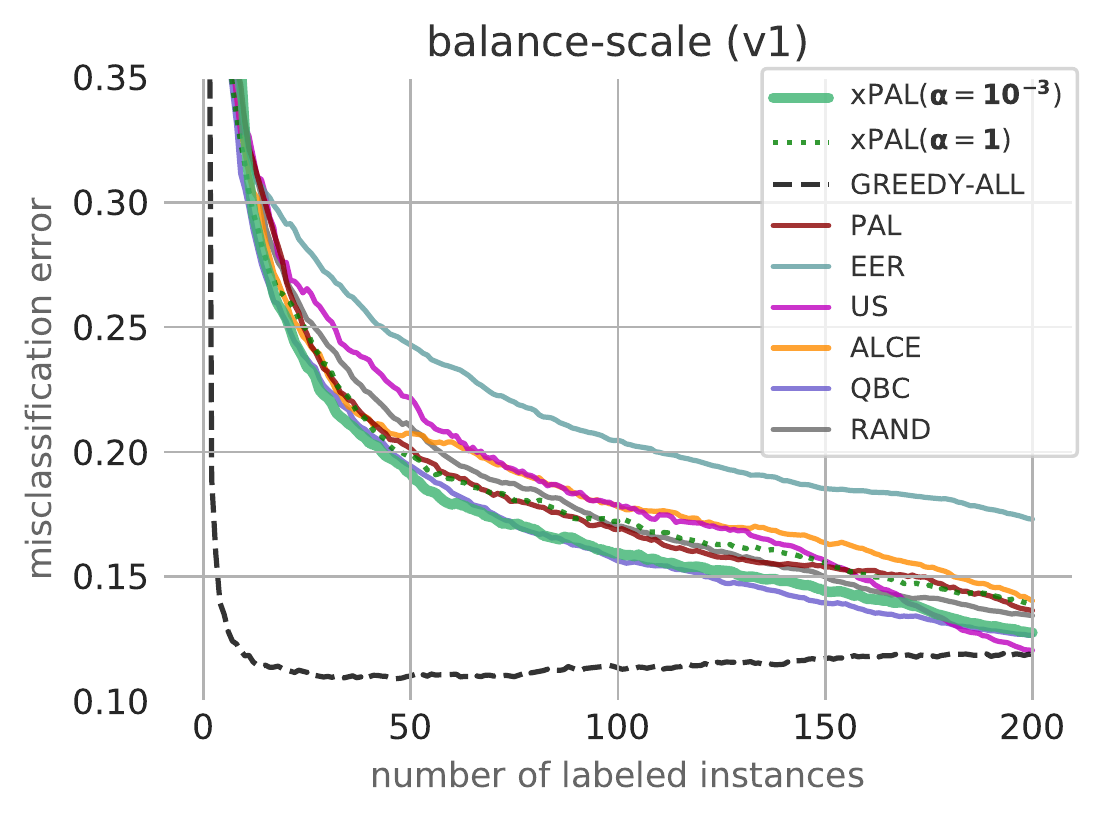} %
	\includegraphics[width=.33\textwidth]{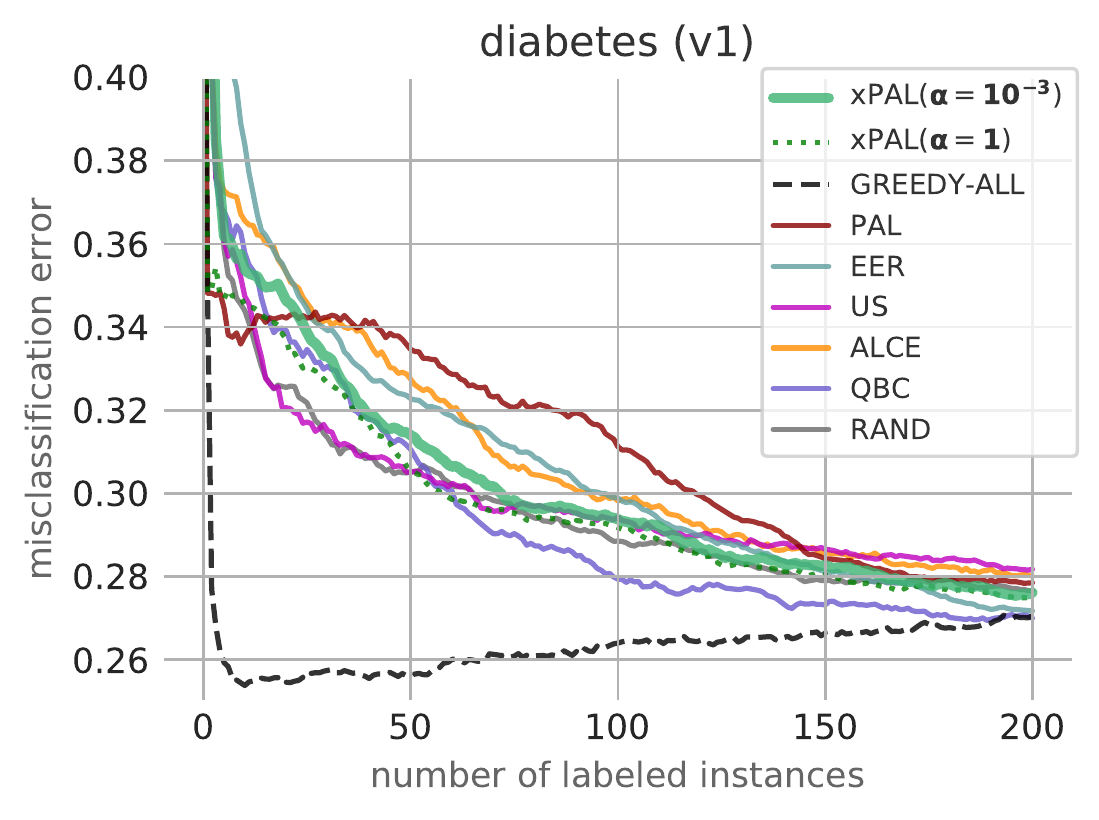}  %
	\includegraphics[width=.33\textwidth]{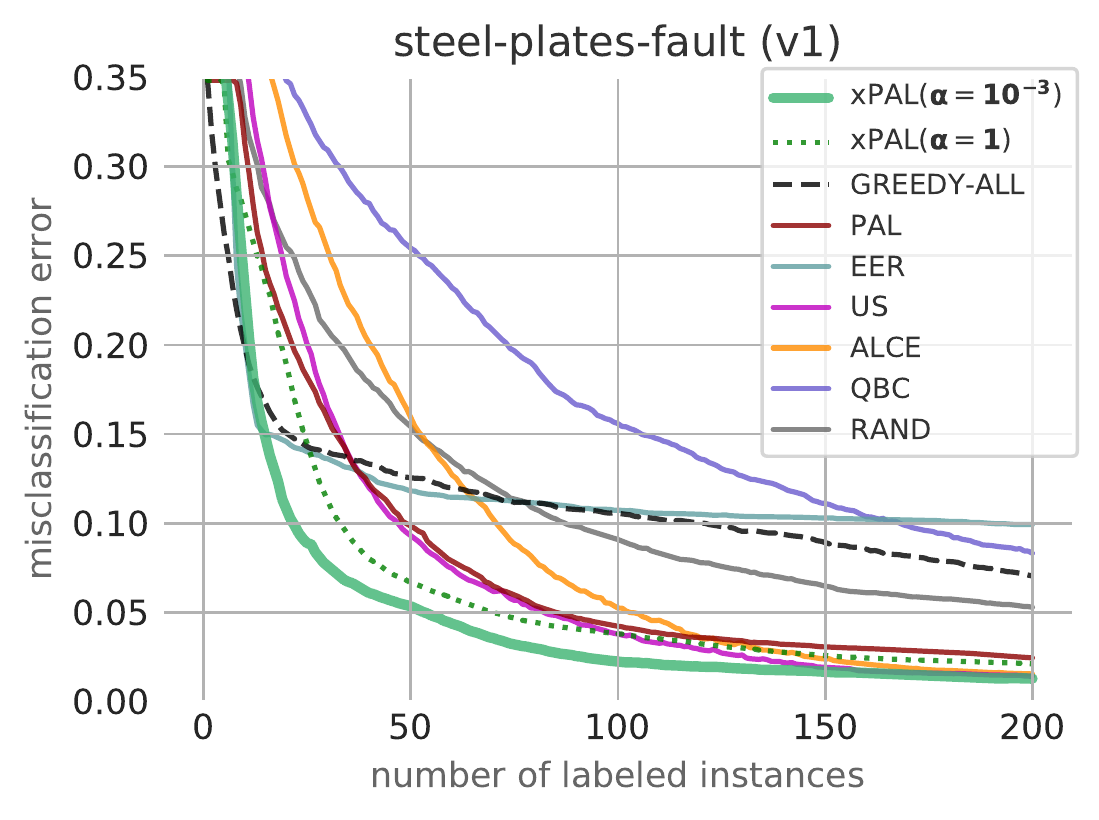}  %
	\vspace{-1em}
	\caption{Learning curves for six selected datasets. Each plot shows the misclassification error of xPAL and the competing algorithms w.\,r.\,t. the number of acquired labels. The learning curve that reaches a low error fast is considered best. The plots of the remaining 16 datasets are given in the supplemental material. }
	\label{fig:learning-curves}
\end{figure*}

We visualize our results using learning curves in Fig.~\ref{fig:learning-curves} and rank statistics in Fig.~\ref{fig:rankings}, \ref{fig:results_categorical}, and \ref{fig:results_text}. More results are given in the appendix. The learning curves show the misclassification error (averaged over the 100 repetitions) on the test set after each label acquisition for every combination of an algorithm and a dataset. The learning curve that reaches a low error fast is considered best.

\begin{figure*}[p!]
	\centering
	\vspace{1em}
	\includegraphics[scale=0.16]{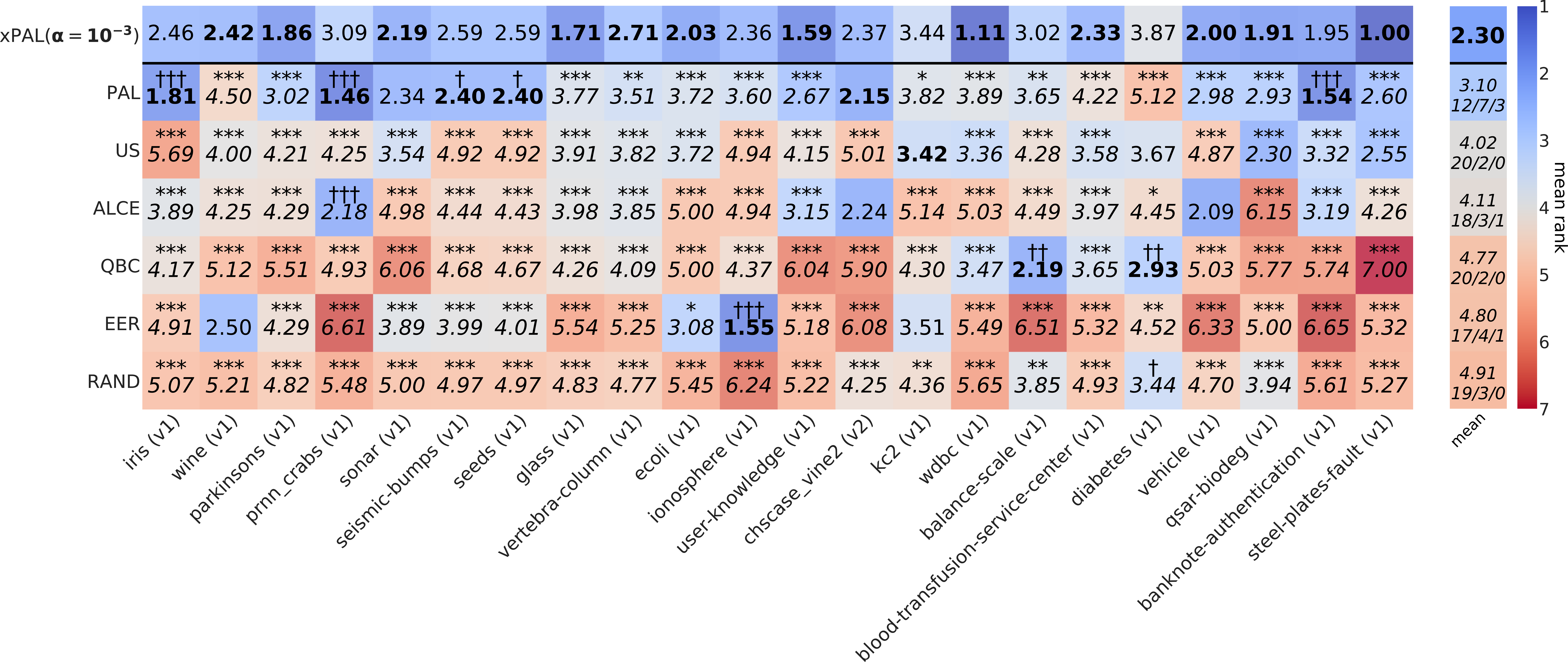}
	\caption{The mean rank for all combinations of selection strategies and numerical datasets (\textbf{RBF kernel}) across 100 repetitions. The best strategy is printed in bold. 
    Three stars (***) indicate significantly better results of xPAL with $p$-value $.001$, two stars (**) indicate a $p$-value of $0.01$ and one star (*) of $.05$. Analogously, significantly better performance of a competitor is shown with $\dag$.
	}
	\label{fig:rankings}
\end{figure*}
Almost all \textbf{learning curves} show that the supervised baseline (GREEDY-ALL) performs perfectly in an early phase. This is not surprising as it knows all labels (even from the unlabeled set $\US$) to optimize the error on the training set. As seen in \textit{steel-plates-fault}, this baseline does not achieve the best performance in all cases because of the greedy selection (no look-ahead). In that example, an optimal baseline would need to create a strategy for more than just the upcoming candidate.
Also, the xPAL approach (green, bold line) with $\prior = \vec{10^{-3}}$ performs well. For convenience, we plotted the xPAL also with $\prior = \vec{1}$ as another alternative. The differences between both curves are rather small. 

\begin{figure*}
    \begin{minipage}[b][7.5cm][t]{.36\textwidth}
    	\centering
    	\includegraphics[scale=0.16]{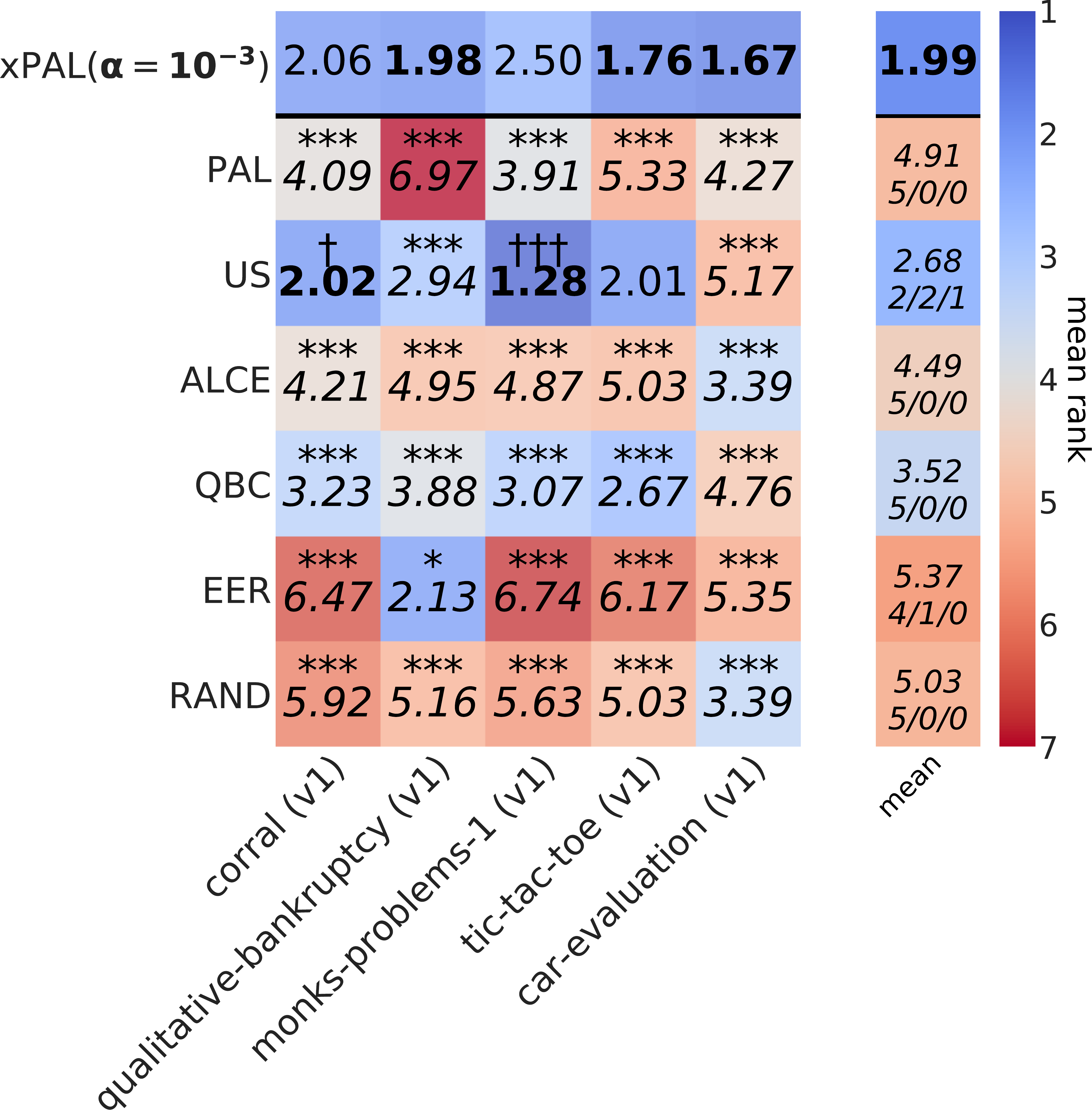}
    	\vspace{-5em}
    	\vfill
    	\captionof{figure}{The mean rank of selection strategies on datasets with categorical features using the \textbf{hamming kernel}.}
    	\label{fig:results_categorical}
    \end{minipage}
    \hfill
    \begin{minipage}[b][7.5cm][t]{.25\textwidth}
    	\centering
    	\includegraphics[scale=0.16]{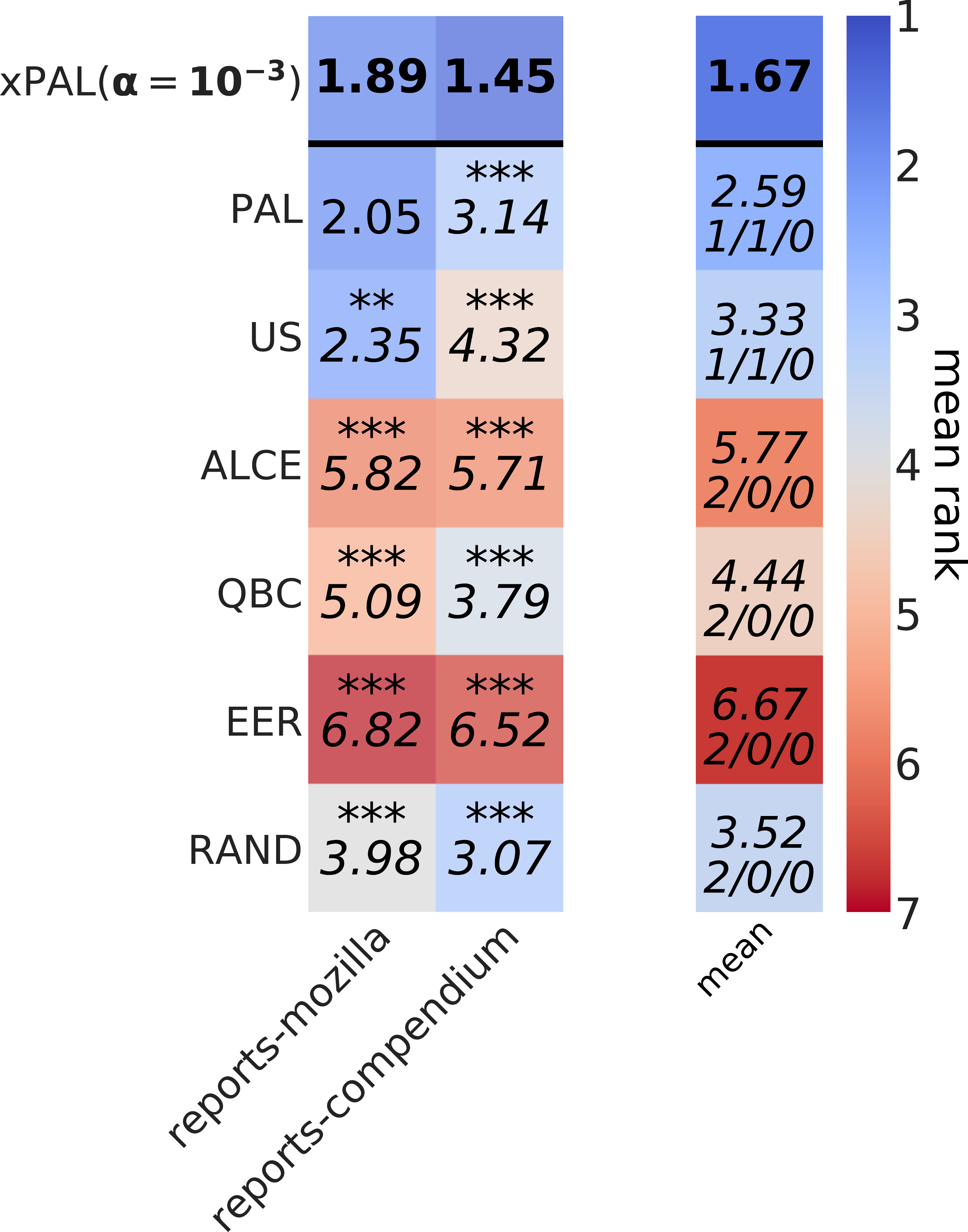}
    	\vfill
    	\captionof{figure}{The mean rank of selection strategies on text datasets using the \textbf{cosine kernel}.}
    	\label{fig:results_text}
    \end{minipage}
    \hfill
    \begin{minipage}[b][7.5cm][t]{.32\textwidth}
    	\centering
    	\includegraphics[width=.9\textwidth]{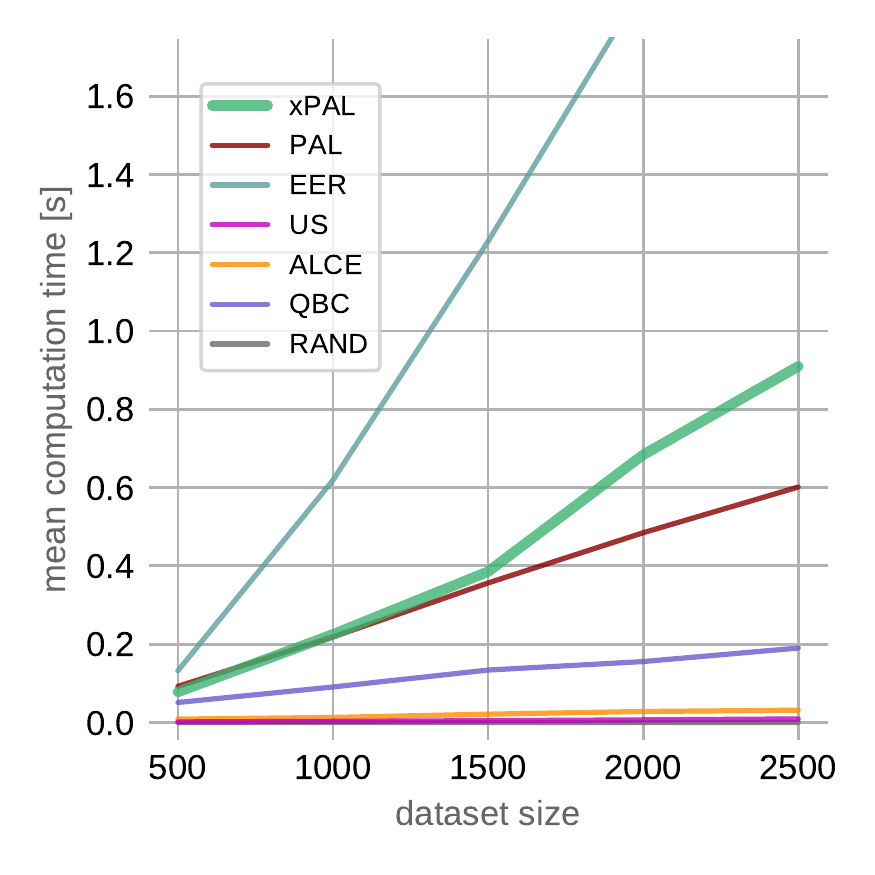} 
    	\vfill
    	\captionof{figure}{Mean computation time per label acquisitions on artificial data (2, 4, 6 classes) with varying dataset size.}
    	\label{fig:computational-complexity}    
    \end{minipage}
\end{figure*}

As it remains difficult to quantitatively assess the performance due to the large amount of datasets, we provide the \textbf{mean rank plot} in Fig.~\ref{fig:rankings}, \ref{fig:results_categorical}, and \ref{fig:results_text}. For this purpose, we calculated the rank of the area under the learning curve for each of the 100 repetitions and average this rank for every combination of a selection strategy and a dataset. We use color to visualize the performance: blue color means good rank, and red color indicates bad performance. The rank of the best algorithm is printed in bold. 
Moreover, we performed a Wilcoxon-signed-rank test to assess if the pairwise differences between xPAL and its competitors are significant. 
Three stars (***) indicate significantly better results of xPAL with a $p$-value of $.001$, two stars (**) indicate a $p$-value of $0.01$ and one star (*) of $.05$. Analogously, significantly better performance of a competitor is shown with $\dag$.
We yield the mean column (right) by averaging the ranks over all datasets. The pattern (a/b/c) in the second row of each cell summarizes a) the number of highly significant wins, c) the number of highly significant losses, and b) neither of both. 

We separated the ranking plots w.\,r.\,t.\, the kernel function. Figure~\ref{fig:rankings} shows results with the RBF kernel, Fig.~\ref{fig:results_categorical} with the hamming-distance kernel, and Fig.~\ref{fig:results_text} with the cosine similarity kernel.
One can observe that xPAL has the lowest mean rank for all kernels and is always printed in blueish color across the datasets. 
No other algorithm performs as robust. 
The strongest competitor is PAL.
But on the categorical data, we observe a clear performance difference between PAL and xPAL.
One reason might be the difficulty of obtaining a reliable density estimation for categorical data.



\subsection{Robustness of Prior Parameter}

\begin{figure}[b!]
	\centering
	\includegraphics[width=.44\textwidth]{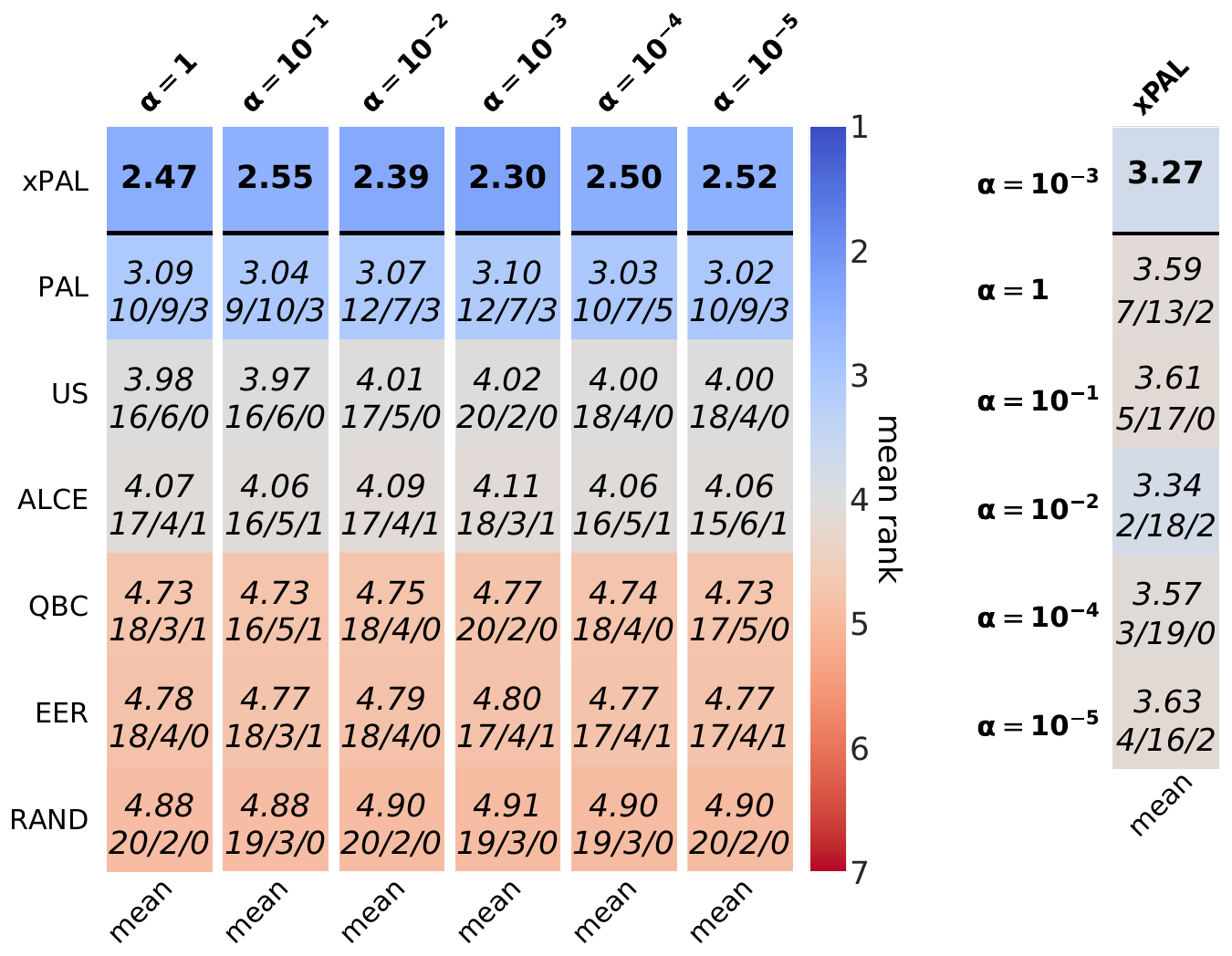} 
	\caption{The mean rank averaged over all numerical datasets for different parameters $\prior$. The 2nd row shows (wins/ties/losses) based on the results of the Wilcoxon-signed-rank test.}
	\label{fig:ranking-prior}
\end{figure}

In Fig.~\ref{fig:ranking-prior}, we show the mean ranking over all numerical datasets for different choices of priors $\prior$. Compared to the other strategies (left image), there is only a small difference across all choices. Comparing xPAL with $\prior=\vec{10^{-3}}$ to the other priors (right image), we see that there are datasets where the selected xPAL is significantly outperformed but in general, the effect is neglectable. Also, all mean ranks are between $3.27$ and $3.63$, which validates the robustness of our parameter. We propose to use $\prior=\vec{10^{-3}}$ as default.

\subsection{Computation Time}

In Tab.~\ref{tab:time_complexity}, we already showed the theoretical time complexity. 
In this section, we now show the actual computation time which of course also depends on the efficiency of the implementation. 
Therefore, we artificially generated datasets with $500, 1000, \dots, 2500$ instances and $2, 4, 6$ classes. 
With every selection strategy, we acquired 200 labels and report the mean computation time on a personal computer in Fig.~\ref{fig:computational-complexity}. 
We clearly see the exponential behavior of EER which is also visible for xPAL. 
As xPAL only needs to calculate the loss difference on instances, where the decision actually changes, we can reduce the computation time to a significant amount. 
Because of the inefficient optimization in PAL, we are even comparably fast to PAL for dataset with less than 1000 instances.


\section{Conclusion}
In this article, we moved toward optimal probabilistic AL by proposing xPAL. It is a decision-theoretic approach that determines the expected performance gain for labeling a candidate using a conjugate prior. We used this model to show the similarities and differences to the most related approaches and compared them by showing how each method selects their instances in a synthetic example. Moreover, we provide an exhaustive experimental evaluation indicating the superiority of xPAL and the robustness of its prior parameter.

In future work, we aim to apply this idea to other cost-sensitive loss functions and for error-prone annotators as this is a current limitation of this article. Moreover, we research possibilities to use the concept of xPAL to define a stopping criterion and to apply it for other classifier types.
The combination of xPAL with methods of deep learning is also promising. However, several challenges need to be addressed, such as unreliable estimates of the class probabilities and the estimation of the vector $\kx_{\x_c}^{\LS}$. 
The former might be solvable by using techniques that improve the returned probabilities (e.\,g.,~by using Bayesian neural networks). The latter could be addressed by transforming samples into a latent representation (e.\,g.,~by using variational autoencoders). The resulting features would allow for a kernel density estimation.
To extend this idea to regression problems, it will be necessary to combine the normally distributed output with a conjugate prior distribution (e.\,g.,~Gaussian-Wishart). This would allow for an analytic solution of the posterior which enables reliable estimation of the risk.


\bibliography{references}

\begin{thebibliography}{36}
\providecommand{\natexlab}[1]{#1}
\providecommand{\url}[1]{\texttt{#1}}
\expandafter\ifx\csname urlstyle\endcsname\relax
  \providecommand{\doi}[1]{doi: #1}\else
  \providecommand{\doi}{doi: \begingroup \urlstyle{rm}\Url}\fi

\bibitem[Baram et~al.(2004)Baram, Yaniv, and Luz]{BaramYanivLuz2004}
Baram, Y., Yaniv, R.~E., and Luz, K.
\newblock Online choice of active learning algorithms.
\newblock \emph{Journal of Machine Learning Research}, 5\penalty0
  (Mar):\penalty0 255--291, 2004.

\bibitem[Beyer et~al.(2015)Beyer, Krempl, and Lemaire]{Beyer2015}
Beyer, C., Krempl, G., and Lemaire, V.
\newblock How to select information that matters: A comparative study on active
  learning strategies for classification.
\newblock In \emph{Proceedings of the 15th International Conference on
  Knowledge Technologies and Data-Driven Business}, i-KNOW ’15, New York, NY,
  USA, 2015. Association for Computing Machinery.

\bibitem[Bishop(2006)]{bishop2006pattern}
Bishop, C.~M.
\newblock \emph{Pattern recognition and machine learning}.
\newblock Springer, 2006.

\bibitem[Bondu et~al.(2010)Bondu, Lemaire, and Boullé]{BonduLemaireBoulle2010}
Bondu, A., Lemaire, V., and Boullé, M.
\newblock Exploration vs. exploitation in active learning : A {Bayes}ian
  approach.
\newblock In \emph{Int. Joint Conf. on Neural Networks (IJCNN)}, pp.\  1–7.
  IEEE, 2010.

\bibitem[Brinker(2003)]{Brinker2003}
Brinker, K.
\newblock Incorporating diversity in active learning with support vector
  machines.
\newblock In \emph{Proc. of the 20th Int. Conf. on Machine Learning (ICML)},
  pp.\  59--66, 2003.

\bibitem[Calma et~al.(2018)Calma, Reitmaier, and Sick]{CALMA201813}
Calma, A., Reitmaier, T., and Sick, B.
\newblock Semi-supervised active learning for support vector machines: A novel
  approach that exploits structure information in data.
\newblock \emph{Information Sciences}, 456:\penalty0 13 -- 33, 2018.

\bibitem[Chapelle(2005)]{chapelle2005active}
Chapelle, O.
\newblock Active learning for parzen window classifier.
\newblock In \emph{Proc. of the 10th Int. Workshop on Artificial Intelligence
  and Statistics (AISTATS)}, volume~5, pp.\  49--56, 2005.

\bibitem[{Chaudhuri} et~al.(2017){Chaudhuri}, {Kakde}, {Sadek}, {Gonzalez}, and
  {Kong}]{8215749}
{Chaudhuri}, A., {Kakde}, D., {Sadek}, C., {Gonzalez}, L., and {Kong}, S.
\newblock The mean and median criteria for kernel bandwidth selection for
  support vector data description.
\newblock In \emph{Int. Conf. on Data Mining Workshops (ICDMW)}, pp.\
  842--849. IEEE, Nov 2017.

\bibitem[Cuong et~al.(2014)Cuong, Lee, and Ye]{CuongLeeYe2014}
Cuong, N.~V., Lee, W.~S., and Ye, N.
\newblock Near-optimal adaptive pool-based active learning with general loss.
\newblock In \emph{Proc. of the 30th Conf. on Uncertainty in Artificial
  Intelligence (UAI)}, pp.\  122--131, 2014.

\bibitem[Donmez et~al.(2007)Donmez, Carbonell, and Bennett]{donmez2007dual}
Donmez, P., Carbonell, J.~G., and Bennett, P.~N.
\newblock Dual strategy active learning.
\newblock In \emph{Proc. of the European Conf. on Machine Learning (ECML)},
  pp.\  116--127. Springer, 2007.

\bibitem[Golovin \& Krause(2010)Golovin and Krause]{GolovinKrause2010}
Golovin, D. and Krause, A.
\newblock Adaptive submodularity: A new approach to active learning and
  stochastic optimization.
\newblock In Kalai, A.~T. and Mohri, M. (eds.), \emph{Proc. of the 23rd Conf.
  on Algorithmic Learning Theory (ALT)}, pp.\  333--345, 2010.

\bibitem[Guillory \& Bilmes(2010)Guillory and Bilmes]{GuilloryBilmes2010}
Guillory, A. and Bilmes, J.
\newblock Interactive submodular set cover.
\newblock In \emph{Proc. of the 27th Int. Conf. on Machine Learning (ICML)},
  2010.

\bibitem[Hernández-González et~al.(2018)Hernández-González, Rodriguez,
  Inza, Harrison, and Lozano]{HernandezGonzales2018}
Hernández-González, J., Rodriguez, D., Inza, I., Harrison, R., and Lozano,
  J.~A.
\newblock Two datasets of defect reports labeled by a crowd of annotators of
  unknown reliability.
\newblock \emph{Data in Brief}, 18:\penalty0 840 -- 845, 2018.

\bibitem[Houlsby et~al.(2011)Houlsby, Huszar, Ghahramani, and
  Lengyel]{HoulsbyEtal2011}
Houlsby, N., Huszar, F., Ghahramani, Z., and Lengyel, M.
\newblock {Bayes}ian active learning for classification and preference
  learning.
\newblock \emph{Computing Research Repository (CoRR)}, abs/1112.5745, 2011.

\bibitem[{Huang} \& {Lin}(2016){Huang} and {Lin}]{7837927}
{Huang}, K. and {Lin}, H.
\newblock A novel uncertainty sampling algorithm for cost-sensitive multiclass
  active learning.
\newblock In \emph{Proc. of the 16th Int. Conf. on Data Mining (ICDM)}, pp.\
  925--930. IEEE, Dec 2016.

\bibitem[Hutter et~al.(2014)Hutter, Xu, Hoos, and Leyton-Brown]{Hutter2014}
Hutter, F., Xu, L., Hoos, H.~H., and Leyton-Brown, K.
\newblock {Algorithm runtime prediction: Methods \& evaluation}.
\newblock \emph{Artificial Intelligence}, 206:\penalty0 79--111, 2014.

\bibitem[Japkowicz \& Shah(2011)Japkowicz and Shah]{japkowicz2011evaluating}
Japkowicz, N. and Shah, M.
\newblock \emph{Evaluating learning algorithms: a classification perspective}.
\newblock Cambridge University Press, 2011.

\bibitem[Konyushkova et~al.(2018)Konyushkova, Sznitman, and
  Fua]{KonyushkovaSznitmanFua2018}
Konyushkova, K., Sznitman, R., and Fua, P.
\newblock Discovering general purpose active learning strategies.
\newblock \emph{arXiv preprint arXiv:1810.04114}, 2018.

\bibitem[Kottke et~al.(2016)Kottke, Krempl, Lang, Teschner, and
  Spiliopoulou]{kottke2016mcpal}
Kottke, D., Krempl, G., Lang, D., Teschner, J., and Spiliopoulou, M.
\newblock Multi-class probabilistic active learning.
\newblock In \emph{Proc. of the European Conf. on Artificial Intelligence
  (ECAI)}, pp.\  586--594. IOS Press, 2016.

\bibitem[Lewis \& Gale(1994)Lewis and Gale]{lewis1994uncertainty}
Lewis, D.~D. and Gale, W.~A.
\newblock A sequential algorithm for training text classifiers.
\newblock In \emph{Proc. of the 17th Annual Int. Conf. on Research and
  Development in Information Retrieval (SIGIR)}, pp.\  3--12. Springer, 1994.

\bibitem[Murphy(2006)]{murphy2006binomial}
Murphy, K.~P.
\newblock Binomial and multinomial distributions.
\newblock \emph{University of British Columbia, Tech. Rep}, 2006.

\bibitem[Nguyen \& Smeulders(2004)Nguyen and Smeulders]{NguyenSmeulders2004}
Nguyen, H.~T. and Smeulders, A.
\newblock Active learning using pre-clustering.
\newblock In \emph{Proc. of the 21st Int. Conf. on Machine Learning (ICML)},
  pp.\  79–86. ACM Press, 2004.

\bibitem[Osugi et~al.(2005)Osugi, Kim, and Scott]{OsugiKimScott2005}
Osugi, T., Kim, D., and Scott, S.
\newblock Balancing exploration and exploitation: A new algorithm for active
  machine learning.
\newblock In \emph{Proc. of the 5th Int. Conf. on Data Mining (ICDM)}, pp.\
  8–pp. IEEE, 2005.

\bibitem[Roy \& McCallum(2001)Roy and McCallum]{roy2001eer}
Roy, N. and McCallum, A.
\newblock Toward optimal active learning through monte carlo estimation of
  error reduction.
\newblock \emph{Proc. of the 18th Int. Conf. on Machine Learning (ICML)}, pp.\
  441--448, 2001.

\bibitem[Settles(2009)]{settles2009active}
Settles, B.
\newblock Active learning literature survey.
\newblock Technical report, University of Wisconsin-Madison Department of
  Computer Sciences, 2009.

\bibitem[Settles(2012)]{settles2012}
Settles, B.
\newblock \emph{Active Learning}.
\newblock Number~18 in Synthesis Lectures on Artificial Intelligence and
  Machine Learning. Morgan and Claypool Publishers, 2012.

\bibitem[Seung et~al.(1992)Seung, Opper, and Sompolinsky]{seung1992qbc}
Seung, H.~S., Opper, M., and Sompolinsky, H.
\newblock Query by committee.
\newblock In \emph{Proc. of the 5th Annual Workshop on Computational Learning
  Theory (COLT)}, pp.\  287--294. ACM, 1992.

\bibitem[{Shi} et~al.(2008){Shi}, {Liu}, {Huang}, {Zhu}, and {Liu}]{Shi2008}
{Shi}, S., {Liu}, Y., {Huang}, Y., {Zhu}, S., and {Liu}, Y.
\newblock {Active Learning for kNN Based on Bagging Features}.
\newblock In \emph{2008 Fourth International Conference on Natural
  Computation}, pp.\  61--64, Jinan, China, 2008.

\bibitem[Thrun \& M{\"o}ller(1992)Thrun and M{\"o}ller]{ThrunMoller1992}
Thrun, S.~B. and M{\"o}ller, K.
\newblock Active exploration in dynamic environments.
\newblock In \emph{Advances in neural information processing systems}, pp.\
  531--538, 1992.

\bibitem[Vanschoren et~al.(2013)Vanschoren, van Rijn, Bischl, and
  Torgo]{vanschoren2013openml}
Vanschoren, J., van Rijn, J.~N., Bischl, B., and Torgo, L.
\newblock Openml: Networked science in machine learning.
\newblock \emph{SIGKDD Explorations}, 15\penalty0 (2):\penalty0 49--60, 2013.

\bibitem[Vapnik(1995)]{Vapnik1995}
Vapnik, V.~N.
\newblock \emph{The Nature of Statistical Learning Theory}.
\newblock Springer-Verlag, Berlin, Heidelberg, 1995.
\newblock ISBN 0-387-94559-8.

\bibitem[Wei et~al.(2015)Wei, Iyer, and Bilmes]{WeiIyerBilmes2015}
Wei, K., Iyer, R., and Bilmes, J.
\newblock Submodularity in data subset selection and active learning.
\newblock In \emph{Proc. of the 32rd Int. Conf. on Machine Learning (ICML)},
  pp.\  1954--1963, 2015.

\bibitem[Weigl et~al.(2015)Weigl, Heidl, Lughofer, Radauer, and
  Eitzinger]{WeiglEtal2015}
Weigl, E., Heidl, W., Lughofer, E., Radauer, T., and Eitzinger, C.
\newblock On improving performance of surface inspection systems by online
  active learning and flexible classifier updates.
\newblock \emph{Machine Vision and Applications}, 27\penalty0 (1):\penalty0
  103–127, 2015.
\newblock ISSN 1432-1769.

\bibitem[Xu et~al.(2007)Xu, Akella, and Zhang]{XuAkellaZhang2007}
Xu, Z., Akella, R., and Zhang, Y.
\newblock Incorporating diversity and density in active learning for relevance
  feedback.
\newblock In \emph{Proc. of the European Conf. on Information Retrieval
  (ECIR)}, pp.\  246--257. Springer, 2007.

\bibitem[{\v{Z}}liobait{\.e} et~al.(2014){\v{Z}}liobait{\.e}, Bifet,
  Pfahringer, and Holmes]{zliobaite2014active}
{\v{Z}}liobait{\.e}, I., Bifet, A., Pfahringer, B., and Holmes, G.
\newblock Active learning with drifting streaming data.
\newblock \emph{Transactions on Neural Networks and Learning Systems},
  25\penalty0 (1):\penalty0 27--39, 2014.

\bibitem[Zoller \& Buhmann(2000)Zoller and Buhmann]{ZollerBuhmann2000}
Zoller, T. and Buhmann, J.~M.
\newblock Active learning for hierarchical pairwise data clustering.
\newblock In \emph{Proc. 15th Int. Conf. on Pattern Recognition (ICPR)},
  volume~2, pp.\  186--189. IEEE, 2000.

\end{thebibliography}
\bibliographystyle{icml2020}

\clearpage
\newpage

\onecolumn
\appendix

\section{Proofs}

\subsection{Proof for Theorem 1}
In Sec.~2, \citet{roy2001eer} describe the algorithm: The estimate the expected loss from Eq. (4) using a Monte-Carlo approach over $\mathcal{P}$. They describe to use the unlabeled pool for that. In our work, we call this the candidate set $\US$. Their algorithm consists of 4 steps: In short, they calculate the \textit{average expected loss} for every instance $\x_c \in \US$. Therefor, they consider every possible label $y_c \in \mathcal{Y}$ and add the pair $(\x_c, y_c)$ to the training set $\mathcal{D}$ (here: $\LS$). They call the resulting set $\mathcal{D}^*$ (here $\LS^+$). The resulting expected losses are averaged, weighted with the respective posterior probability $\pr{y_c}{\x_c}$. 
\begin{align} \label{app:eq:posterior}
    \pr{y_c}{\x_c} = \frac{(\kc^\LS)_{y_c}}{||\kc^\LS||_1}
\end{align}
The posterior probabilities for our kernel-based classifier are determined using Eq.~\ref{app:eq:posterior}. \citet{chapelle2005active} proposed to include a beta-prior and thereby extended the approach by \citet{roy2001eer}.
\begin{align} \label{app:eq:posterior-chapelle}
    \pr{y_c}{\x_c} = \normalizedfrequency{\kc^{\LS}}{\epsilon}{y_c}
\end{align}
The resulting equation can be simplified as follows:
\begin{align}
    \mathrm{eer}(\x_c, \LS, \US)
    & =  \sum_{y_c \in \mathcal{Y}} \pr{y_c}{\cx, \LS} \cdot
        \frac{1}{|\US|} \sum_{\x \in \mathcal{\US}} 
            \left( 1 - \max_{y \in \mathcal{Y}} \pr{y}{\x, \LS^+} \right) \\
    & = \sum_{y_c \in \mathcal{Y}} \normalizedfrequency{\kc^\LS}{\epsilon}{y_c} \cdot \frac{1}{|\US|} \sum_{\x \in \US} 
    \left( 1 - \max_{y \in \mathcal{Y}} \normalizedfrequency{\kc^{\LS^+}}{\epsilon}{y} \right) \\
    & = \sum_{y_c \in \mathcal{Y}} \normalizedfrequency{\kc^\LS}{\epsilon}{y_c} \cdot \frac{1}{|\US|} \sum_{\x \in \US} 
    \left( \sum_{y \in \mathcal{Y}} \1\left\{y = f^{\LS^+}(\x)\right\}
    \left(1 - \normalizedfrequency{\kc^{\LS^+}}{\epsilon}{y}\right) \right) \\
    & = \sum_{y_c \in \mathcal{Y}} \normalizedfrequency{\kc^\LS}{\epsilon}{y_c} \cdot \frac{1}{|\US|} \sum_{\x \in \US} 
    \left( \sum_{y \in \mathcal{Y}} \1\left\{y \neq f^{\LS^+}(\x)\right\}
    \cdot \normalizedfrequency{\kc^{\LS^+}}{\epsilon}{y}\right) \\
    & = \sum_{y_c \in \mathcal{Y}} \normalizedfrequency{\kc^\LS}{\epsilon}{y_c} \cdot \frac{1}{|\US|} \sum_{\x \in \US} 
    \left( \sum_{y \in \mathcal{Y}} \normalizedfrequency{\kc^{\LS^+}}{\epsilon}{y} 
    \cdot L(y, f^{\LS^+}(\x)) \right) 
\end{align}
\hfill\ensuremath{\square}

\subsection{Proof of Theorem 2}
Multi-class probabilistic active learning (PAL) by \citet{kottke2016mcpal} describes the expected gain in accuracy. Instead of evaluating this gain on a representative subset, they solely consider the gain locally. To proof Theorem 2, we need to set the $m$ parameter of PAL to $m=1$, which means that we only consider one possible label acquisition in each iteration. \citet{kottke2016mcpal} model the hypothetical labels using a labeling vector $\vec{l} \in \mathbb{N}^C$ which describes the number of potentially added labels for each class. As we only consider one label at a time ($m=1$), these vectors are unit vectors with a $1$ at element of the considered class $y_c$ and $0$ otherwise. Hence, $l \in \{\vec{e}_1, \dots, \vec{e}_C\}$. 
\begin{align}
    l_i &= \begin{cases} 1 & i=y_c \\ 0 & \text{else} \end{cases}
\end{align}

For simplicity, we use $\kx$ instead of writing $\kx_{\cx}$ as PAL solely considers the candidate $\cx$ and no other instance. Moreover, we know that $\kxn = \kx + \vec{e}_{y_c}$, as we increment the frequency estimate of the simulated class $y_c$ by $1$ (the similarity of $\cx$ to $\cx$ is always $1$). 
Additionally to $\vec{l}$, \citet{kottke2016mcpal} model the classifier's decision using a vector $\vec{d}$, which is $1$ for the class of the future decision and $0$ otherwise.
\begin{align}
    d_i &= \begin{cases} 1 & \argmax_y ( \kxn{y} ) = i \\ 0 & \text{else} \end{cases}
\end{align}

For simplicity, we do not write the iterators at sums and products if they iterate from $i=1$ to $C$. 
Based on the old classifier $f^\LS$ and the new classifier $f^{\LS^+}$, we write $\hat{y} = f^\LS(\cx)$ and $\hat{y}^+ = f^{\LS^+}(\cx)$ for the old and the new prediction.

\begin{align}
    \textrm{pal}(\cx, \LS) 
    & = \hat{p}(\cx) \cdot \sum_\vec{l} \left( 
    \underbrace{\left( \prod_{j=\sum(\kx{i} + 1)}^{\sum(\kx{i} + l_i + d_i + 1) - 1} \frac{1}{j} \right)}_{=\text{I}}
    \cdot \underbrace{\prod_{i=1}^C \left( \prod_{j=\kx{i}+1}^{\kx{i}+l_i+d_i} j \right)}_{=\text{II}}
    \cdot \underbrace{\frac{\Gamma((\sum l_i)+1)}{\prod(\Gamma(l_i+1))}}_{=\text{III}} \right)
    - \normalizedfrequency{\kx}{\1}{\hat{y}} \label{eq:pal_start}
\end{align}

\begin{align}
    \text{I} 
    &= \left( \prod_{j=\sum(\kx{i} + 1)}^{\sum(\kx{i} + l_i + d_i + 1) - 1} \frac{1}{j} \right)
    = \left( \prod_{j=0}^{1} \frac{1}{\sum(\kx{i} + 1) + j} \right)
    = \frac{1}{\sum(\kx{i} + 1)} \cdot \frac{1}{\sum(\kx{i} + 1) + 1} \\
    &= \frac{1}{||\kx + 1||_1 \cdot ||\kxn + 1||_1}  \\
    \vspace{.5em} \nonumber \\
    \text{II} 
    &=  \prod_{i=1}^C \left( \prod_{j=\kx{i}+1}^{\kx{i}+l_i+d_i} j \right)
    = \prod_{i=1}^C \left( \prod_{j=1}^{l_i+d_i} \kx{i} + j \right)
    = \prod_{i=1}^C \begin{cases}
    1 & l_i + d_i = 0 \\
    (\kx{i} + 1) & l_i + d_i = 1 \\
    (\kx{i} + 1)(\kx{i} + 2) & l_i + d_i = 2
    \end{cases}\\
    &= \prod_{i=1}^C \begin{cases}
    (\kx{i} + 1) & l_i + d_i = 1 \\
    (\kx{i} + 1)(\kx{i} + 2) & l_i + d_i = 2
    \end{cases}
    = \begin{cases} \label{eq:b}
    (\kx + 1)_{y_c} \cdot (\kx + 2)_{y_c} & \hat{y}^+ = y_c \\
    (\kx + 1)_{y_c} \cdot (\kx + 1)_{\hat{y}^+} & \text{else}
    \end{cases}  \\
    \vspace{.5em} \nonumber \\
    \text{III} 
    &= \frac{\Gamma((\sum l_i)+1)}{\prod(\Gamma(l_i+1))} = \frac{\Gamma(2)}{1} = 1
\end{align}

%

We now insert I, II, III back into Eq.~\ref{eq:pal_start}.
\begin{align} 
    \textrm{pal}(\cx, \LS) 
    & = \hat{p}(\cx) \cdot \sum_\vec{l} \frac{1}{||\kx + \1||_1 \cdot ||\kxn + \1||_1} \cdot
    \begin{cases}
        (\kx + \1)_{y_c} \cdot (\kx + \2)_{y_c} & \hat{y}^+ = y_c \\
        (\kx + \1)_{y_c} \cdot (\kx + \1)_{\hat{y}^+} & \text{else}
    \end{cases} - \normalizedfrequency{\kx}{\1}{\hat{y}} \\
    & = \hat{p}(\cx) \cdot \sum_{y_c \in \mathcal{Y}} \frac{1}{||\kx + \1||_1 \cdot ||\kxn + \1||_1} \cdot
    \begin{cases}
        (\kx + \1)_{y_c} \cdot (\kx + \2)_{y_c} & \hat{y}^+ = y_c \\
        (\kx + \1)_{y_c} \cdot (\kx + \1)_{\hat{y}^+} & \text{else}
    \end{cases} - \normalizedfrequency{\kx}{\1}{\hat{y}} \\
    & = \hat{p}(\cx) \cdot \sum_{y_c \in \mathcal{Y}} \frac{(\kx + 1)_{y_c}}{||\kx + \1||_1 \cdot ||\kxn + \1||_1} \cdot
    \begin{cases}
        (\kx + \2)_{y_c} & \hat{y}^+ = y_c \\
        (\kx + \1)_{\hat{y}^+} & \text{else}
    \end{cases} - \normalizedfrequency{\kx}{\1}{\hat{y}} \label{eq:cases}
\end{align}

We divide the sum into two parts: (A) The subset of all labels ($\mathcal{Y}_{\neq}$) that change the decision, (B) and the labels ($\mathcal{Y}_=$) that do not change the decision. Please remember that a new label $y_c$ could change the decision of $\hat{y}^+$ as it includes the new label. They are defined as follows:
\begin{align}
    \mathcal{Y} = \mathcal{Y}_{\neq} \,\dot\cup\, \mathcal{Y}_=
    = \{y_c \in \mathcal{Y}: \hat{y} \neq \hat{y}^+\}
    \,\dot\cup\, \{y_c \in \mathcal{Y}: \hat{y} = \hat{y}^+\}
\end{align}

Now, we consider both cases independently.

\paragraph{A) Labels that change the decision}
For all $y_c \in \mathcal{Y}$ with $\hat{y} \neq \hat{y}^+$, we know that $\hat{y}^+ = y_c$. \\
It follows that $L(y_c, \hat{y}^+) - L(y, \hat{y}) = -1$.
\begin{align}
    & \sum_{y_c \in \mathcal{Y}_{\neq}} \frac{(\kx + \1)_{y_c}}{||\kx + \1||_1 \cdot ||\kxn + \1||_1} \cdot
    \begin{cases}
        (\kx + \2)_{y_c} & \hat{y}^+ = y_c \\
        (\kx + \1)_{\hat{y}} & \text{else}
    \end{cases}
    = \sum_{y_c \in \mathcal{Y}_{\neq}}
    \normalizedfrequency{\kx}{\1}{y_c} \cdot 
    \frac{(\kx + \2)_{y_c}}{||\kxn + \1||_1} \\
    & = \sum_{y_c \in \mathcal{Y}_{\neq}}
    \normalizedfrequency{\kx}{\1}{y_c} \cdot \normalizedfrequency{\kxn}{\1}{y_c}
    = -\sum_{y_c \in \mathcal{Y}_{\neq}}
    \normalizedfrequency{\kx}{\1}{y_c} \sum_{y \in \{y_c\}} 
    \normalizedfrequency{\kxn}{\1}{y} \cdot (L(y, \hat{y}^+) - L(y, \hat{y}))
\end{align}

\paragraph{B) Labels that do not change the decision}
Here, we can use the following implications to rewrite the cases from Eq.~\ref{eq:cases} into the sum:
\begin{itemize}
    \item $y_c = \hat{y} \Rightarrow \hat{y}^+ = y_c$
    \item $y_c \neq \hat{y} \Rightarrow \hat{y}^+ \neq y_c$
\end{itemize}

\begin{align}
    & \sum_{y_c \in \mathcal{Y}_=} \frac{(\kx + \1)_{y_c}}{||\kx + \1||_1 \cdot ||\kxn + \1||_1} \cdot
    \begin{cases}
        (\kx + \2)_{y_c} & \hat{y}^+ = y_c \\
        (\kx + \1)_{\hat{y}} & \text{else}
    \end{cases} - \normalizedfrequency{\kx}{\1}{\hat{y}} \\
    & = \left( \sum_{y_c \in \mathcal{Y}_= \setminus \{\hat{y}\}}
    \frac{(\kx + \1)_{y_c} \cdot (\kx + \1)_{\hat{y}}}{||\kx + \1||_1 \cdot ||\kxn + \1||_1} \right)
    + \frac{(\kx + \1)_{\hat{y}} \cdot (\kx + \2)_{\hat{y}}}{||\kx + \1||_1 \cdot ||\kxn + \1||_1}
    - \normalizedfrequency{\kx}{\1}{\hat{y}} \\
    & = \left( \sum_{y_c \in \mathcal{Y}_= \setminus \{\hat{y}\}}
    \frac{(\kx + \1)_{y_c} \cdot (\kx + \1)_{\hat{y}}}{||\kx + \1||_1 \cdot ||\kxn + \1||_1} \right)
    + \frac{(\kx + \1)_{\hat{y}} \cdot (\kx + \2)_{\hat{y}}}{||\kx + \1||_1 \cdot ||\kxn + \1||_1}
    - \frac{(\kx + \1)_{\hat{y}} \cdot ||\kxn + \1||_1}{||\kx + \1||_1 \cdot ||\kxn + \1||_1} \\
    & = \frac{(\kx + \1)_{\hat{y}}}{||\kx + \1||_1 \cdot ||\kxn + \1||_1}
    \left( \sum_{y_c \in \mathcal{Y}_= \setminus \{\hat{y}\}}
    (\kx + \1)_{y_c}\right) + (\kx + \2)_{\hat{y}} - (||\kx + \1||_1 + 1)
\end{align}
\begin{align}
    & = \frac{(\kx + \1)_{\hat{y}}}{||\kx + \1||_1 \cdot ||\kxn + \1||_1}
    \left( \sum_{y_c \in \mathcal{Y}_= \setminus \{\hat{y}\}}
    (\kx + \1)_{y_c}\right) + (\kx + \1)_{\hat{y}} - ||\kx + \1||_1 \\
    & = \frac{(\kx + \1)_{\hat{y}}}{||\kx + \1||_1 \cdot ||\kxn + \1||_1}
    \left( \sum_{y_c \in \mathcal{Y}_=}(\kx + \1)_{y_c} \right)
    - \left( \sum_{y_c \in \mathcal{Y}}(\kx + \1)_{y_c} \right) \\
    & = -\sum_{y_c \in \mathcal{Y}_{\neq}}
    \normalizedfrequency{\kx}{\1}{y_c} \cdot
    \frac{(\kx + \1)_{\hat{y}}}{||\kxn + \1||_1} \\
    & = -\sum_{y_c \in \mathcal{Y}_{\neq}} \normalizedfrequency{\kx}{\1}{y_c}
    \sum_{y \in \{\hat{y}\}} \normalizedfrequency{\kxn}{\1}{y} \cdot (L(y, \hat{y}^+) - L(y, \hat{y}))
\end{align}

In the last step, we use that $\hat{y} \neq \hat{y}^+ \Longrightarrow L(\hat{y}, \hat{y}^+) - L(\hat{y}, \hat{y}) = 1$. Additionally, we use that $y_c \neq \hat{y}$ applies and thus $\kx{\hat{y}} = \kxn{\hat{y}}$.
Next, we combine both cases:

\begin{align}
    \textrm{pal}(\cx, \LS) 
    & = - \hat{p}(\cx) \cdot \bigg( \sum_{y_c \in \mathcal{Y}_{\neq}} \normalizedfrequency{\kx}{\1}{y_c}
    \sum_{y \in \{y_c\}} \normalizedfrequency{\kxn}{\1}{y}
    \cdot (L(y, \hat{y}^+) - L(y, \hat{y})) \\
    & \qquad\qquad\quad + \sum_{y_c \in \mathcal{Y}_{\neq}} \normalizedfrequency{\kx}{\1}{y_c}
    \sum_{y \in \{\hat{y}\}} \normalizedfrequency{\kxn}{\1}{y} 
    \cdot (L(y, \hat{y}^+) - L(y, \hat{y}))\bigg) \\
    & = - \hat{p}(\cx) \cdot \sum_{y_c \in \mathcal{Y}_{\neq}} \normalizedfrequency{\kx}{\1}{y_c}
    \sum_{y \in \{y_c, \hat{y}\}} \normalizedfrequency{\kxn}{\1}{y} \cdot (L(y, \hat{y}^+) - L(y, \hat{y}))
\end{align}

Because of $L(y, \hat{y}^+) - L(y, \hat{y}) = 0 \text{ for } y \notin \{y_c, \hat{y}\}$ and for $y_c \in \mathcal{Y}_=$, we can change this equation to

\begin{align}
    \textrm{pal}(\cx, \LS) 
    = - \hat{p}(\cx) \cdot \sum_{y_c \in \mathcal{Y}} \normalizedfrequency{\kx}{\1}{y_c}
    \sum_{y \in \mathcal{Y}} \normalizedfrequency{\kxn}{\1}{y}
    \cdot (L(y, \hat{y}^+) - L(y, \hat{y})).
\end{align}
\hfill\ensuremath{\square}

\subsection{Proof of Theorem 3}

According to \citet{settles2009active}, the usefulness score for ``least confidence uncertainty sampling'' is determined by the following equation and can easily be rewritten. We denote: $\hat{y} = f^\LS(\cx)$.
\begin{align}
    \mathrm{us}(\cx, \LS) 
    & = 1 - \pr{\hat{y}}{\cx}
    = \sum_{y_c \in \mathcal{Y}} \indicator{y_c = f^{\LS}(\x)}
    \left(1 - \pr{y_c}{\cx} \right) 
    = \sum_{y_c \in \mathcal{Y}} \indicator{y_c = f^{\LS}(\x)} 
    \left(1 - \normalizedfrequencyshort{\kc^\LS + \vec{0}}{y_c}\right) \\
    & = \sum_{y_c \in \mathcal{Y}} \indicator{y_c \not= f^{\LS}(\x)}
    \left(\normalizedfrequencyshort{\kc^\LS + \vec{0}}{y_c}\right) 
    = \sum_{y_c \in \mathcal{Y}} 
    \normalizedfrequencyshort{\kc^\LS + \vec{0}}{y_c} \cdot L(y_c, f^\LS(\cx)
\end{align}
\hfill\ensuremath{\square}

\clearpage
\newpage

\section{Description of Datasets}
A detailed description of the datasets is available in Tab.~\ref{tab:datasets}. We provide the openML identifier\footnote{\url{https://www.openml.org/}}, the dataset's name, the number of instances and features, and the distribution of classes (the list describes the fraction of class 1 in the first element, the fraction of class 2 in the second element, etc).

\begin{table*}[h]
    \small
    \centering
    \begin{tabular}{rlccc}
        \toprule
        openML id &                              name & instances & features &                               class distribution \\
        \midrule
        61   &                              iris &       150 &        4 &                               [0.33, 0.33, 0.33] \\
        187  &                              wine &       178 &       13 &                                [0.33, 0.4, 0.27] \\
        1488 &                        parkinsons &       195 &       22 &                                     [0.25, 0.75] \\
        446  &                        prnn\_crabs &       200 &        7 &                                       [0.5, 0.5] \\
        40   &                             sonar &       208 &       60 &                                     [0.53, 0.47] \\
        1500 &                     seismic-bumps &       210 &        7 &                               [0.33, 0.33, 0.33] \\
        1499 &                             seeds &       210 &        7 &                               [0.33, 0.33, 0.33] \\
        41   &                             glass &       214 &        9 &             [0.33, 0.36, 0.06, 0.14, 0.04, 0.08] \\
        1523 &                   vertebra-column &       310 &        6 &                               [0.19, 0.32, 0.48] \\
        39   &                             ecoli &       336 &        7 &  [0.43, 0.23, 0.01, 0.01, 0.1, 0.06, 0.01, 0.15] \\
        59   &                        ionosphere &       351 &       34 &                                     [0.36, 0.64] \\
        1508 &                    user-knowledge &       403 &        5 &                    [0.25, 0.32, 0.3, 0.06, 0.06] \\
        814  &                chscase\_vine2 (v2) &       468 &        2 &                                     [0.45, 0.55] \\
        1063 &                               kc2 &       522 &       21 &                                       [0.8, 0.2] \\
        1510 &                              wdbc &       569 &       30 &                                     [0.63, 0.37] \\
        11   &                     balance-scale &       625 &        4 &                               [0.08, 0.46, 0.46] \\
        1464 &  blood-transfusion-service-center &       748 &        4 &                                     [0.76, 0.24] \\
        37   &                          diabetes &       768 &        8 &                                     [0.65, 0.35] \\
        54   &                           vehicle &       846 &       18 &                         [0.26, 0.25, 0.26, 0.24] \\
        1494 &                       qsar-biodeg &      1055 &       41 &                                     [0.66, 0.34] \\
        1462 &           banknote-authentication &      1372 &        4 &                                     [0.56, 0.44] \\
        1504 &                steel-plates-fault &      1941 &       33 &                                     [0.65, 0.35] \\ \midrule
        40669 & corral              & 160   & 6 &  [0.56, 0.44] \\
        1495 & bankruptcy          & 250   & 6 &  [0.43, 0.57] \\
        333 & monks               & 556   & 6 &  [0.5, 0.5] \\
        50 & tic                 & 958   & 9 &  [0.35, 0.65] \\
        40664 & car                 & 1728  & 21 &  [0.7, 0.22, 0.04, 0.04] \\ \midrule
        - & reports-mozilla     & 675   & 100 &  [0.23, 0.09, 0.43, 0.25] \\
        - & reports-compendium  & 962   & 56 &  [0.09, 0.33, 0.23, 0.35] \\
        \bottomrule
    \end{tabular}
	\medskip
    \caption{Description of datasets.}
    \label{tab:datasets}
\end{table*}

\clearpage
\newpage

\section{More Experimental Results}
In this section, we provide more plots from our experimental evaluation. Please refer to the original paper for the detailed explanation of the experimental setup and the discussion of the results.

\subsection{Usefulness Plots With Randomly Selected Labels}
\begin{figure*}[h!]
    \centering
    \includegraphics[width=.31\textwidth]{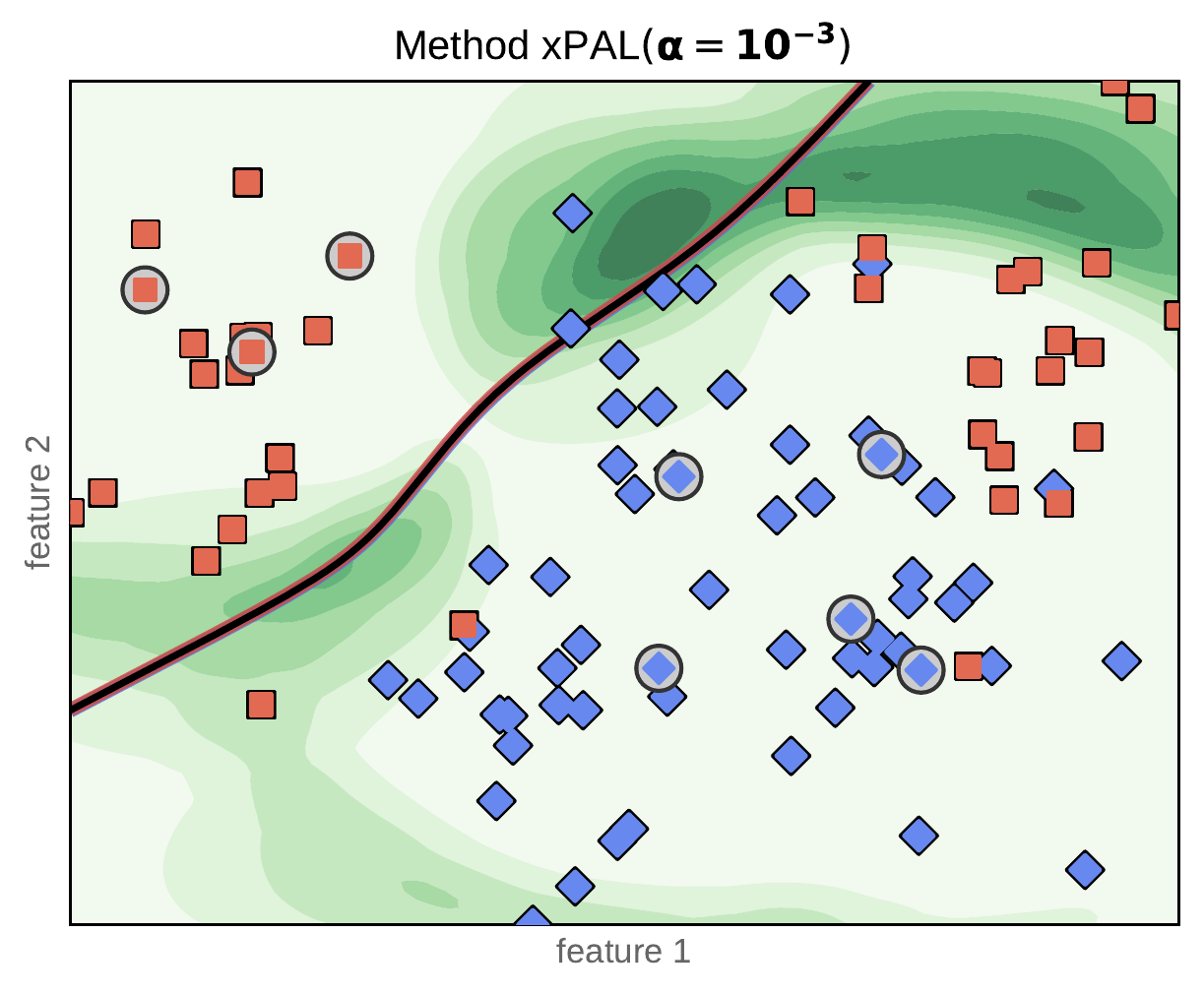}
    \includegraphics[width=.31\textwidth]{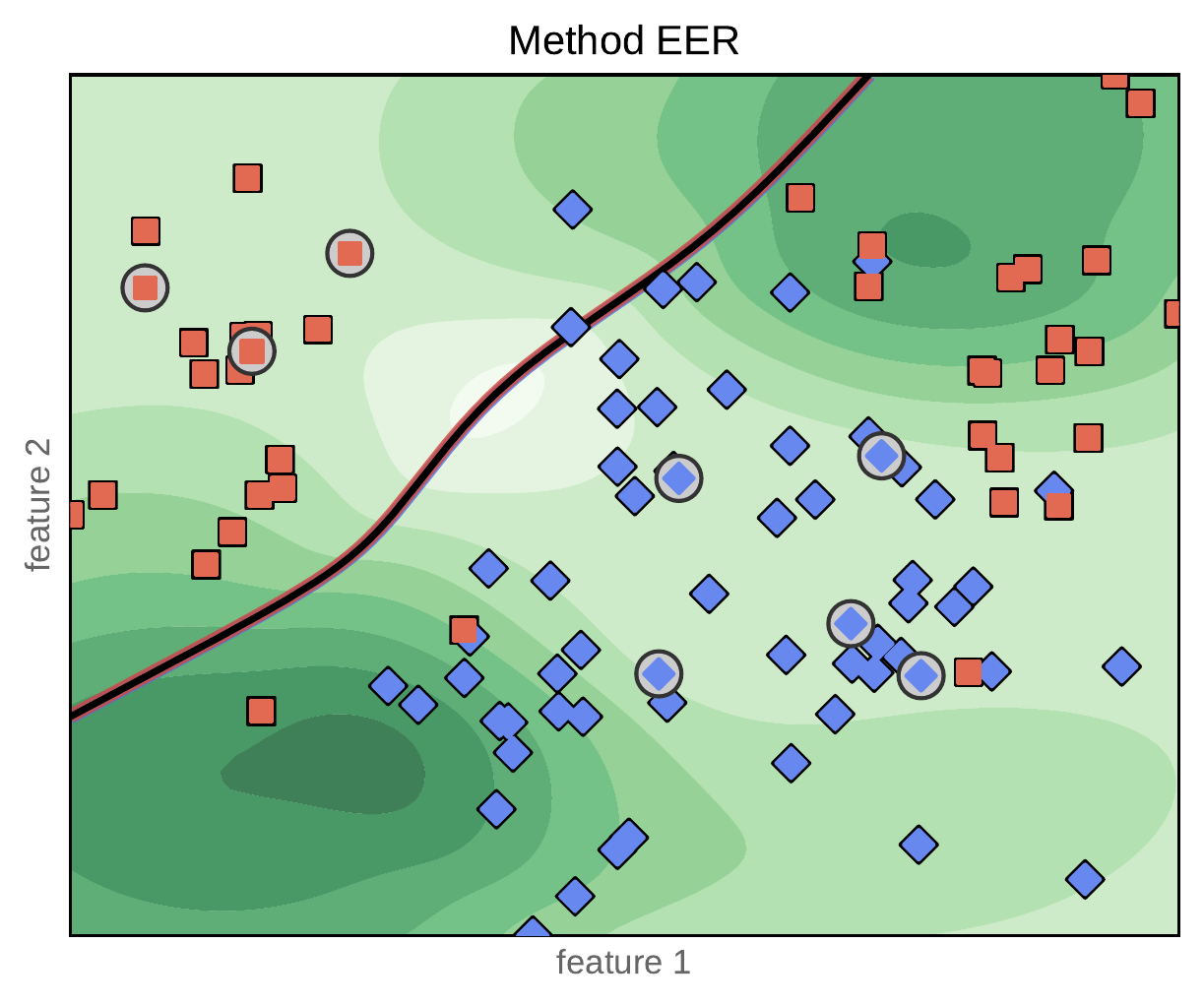}
    \includegraphics[width=.31\textwidth]{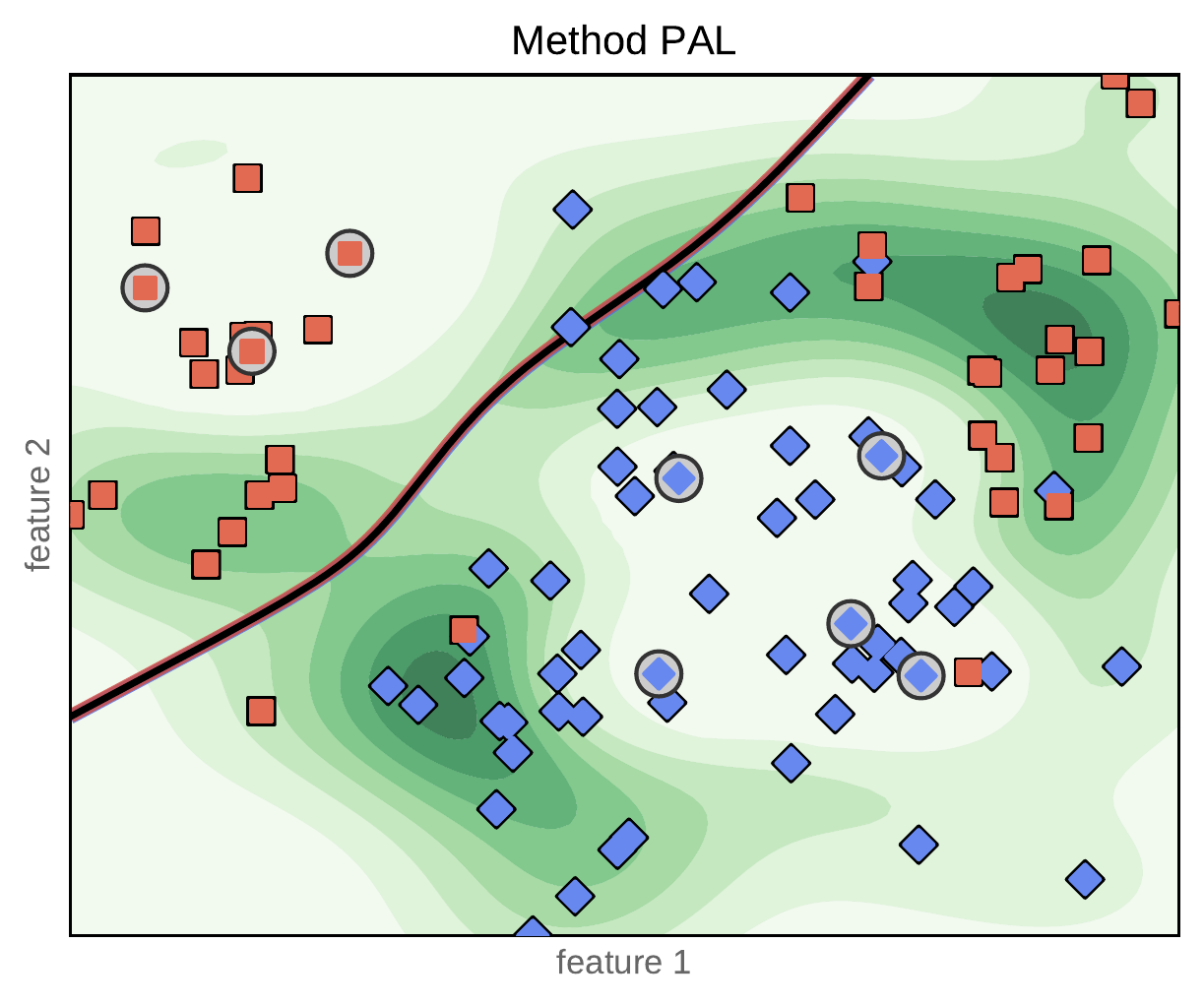}
    \includegraphics[width=.31\textwidth]{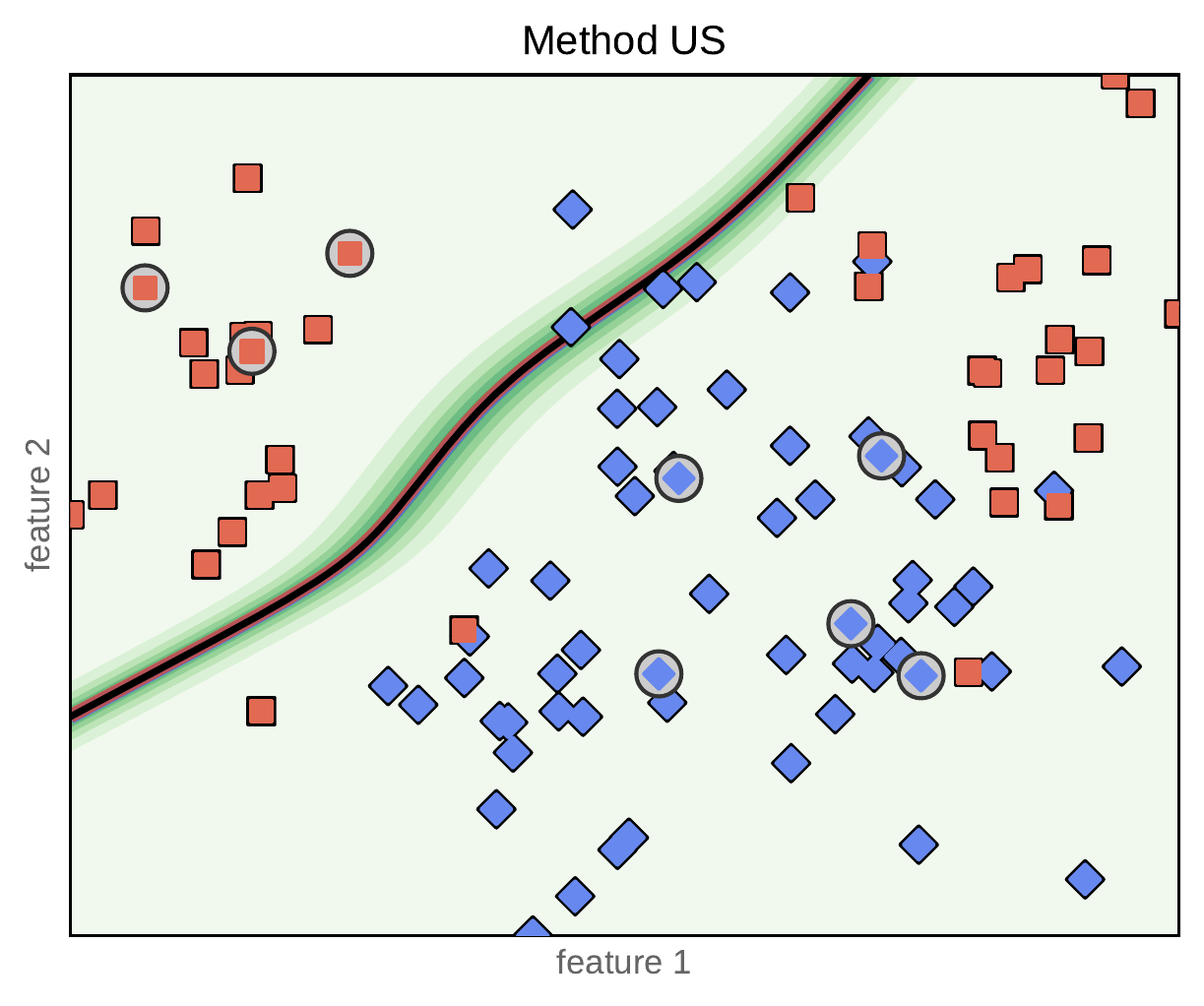}
    \includegraphics[width=.31\textwidth]{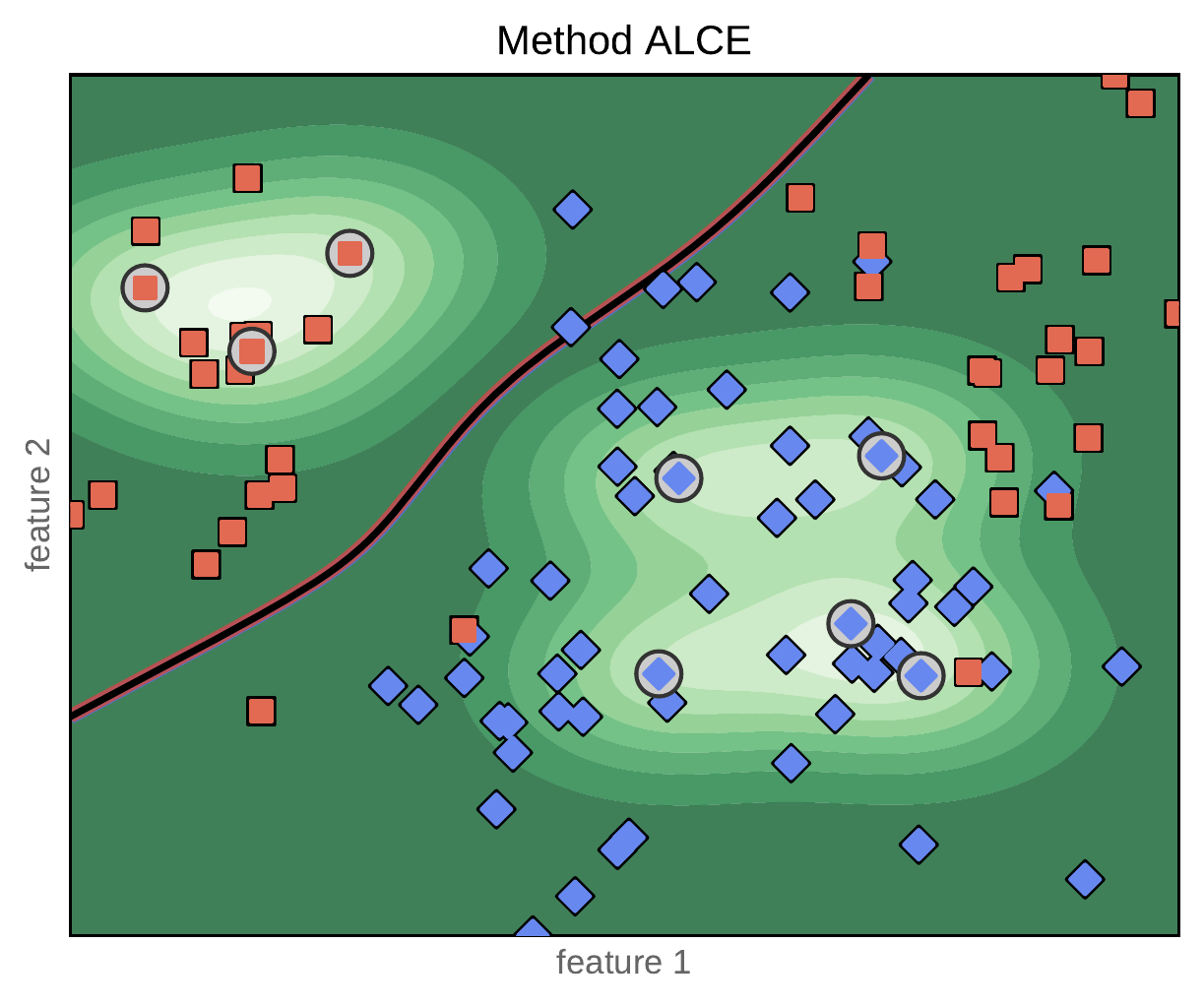}
    \includegraphics[width=.31\textwidth]{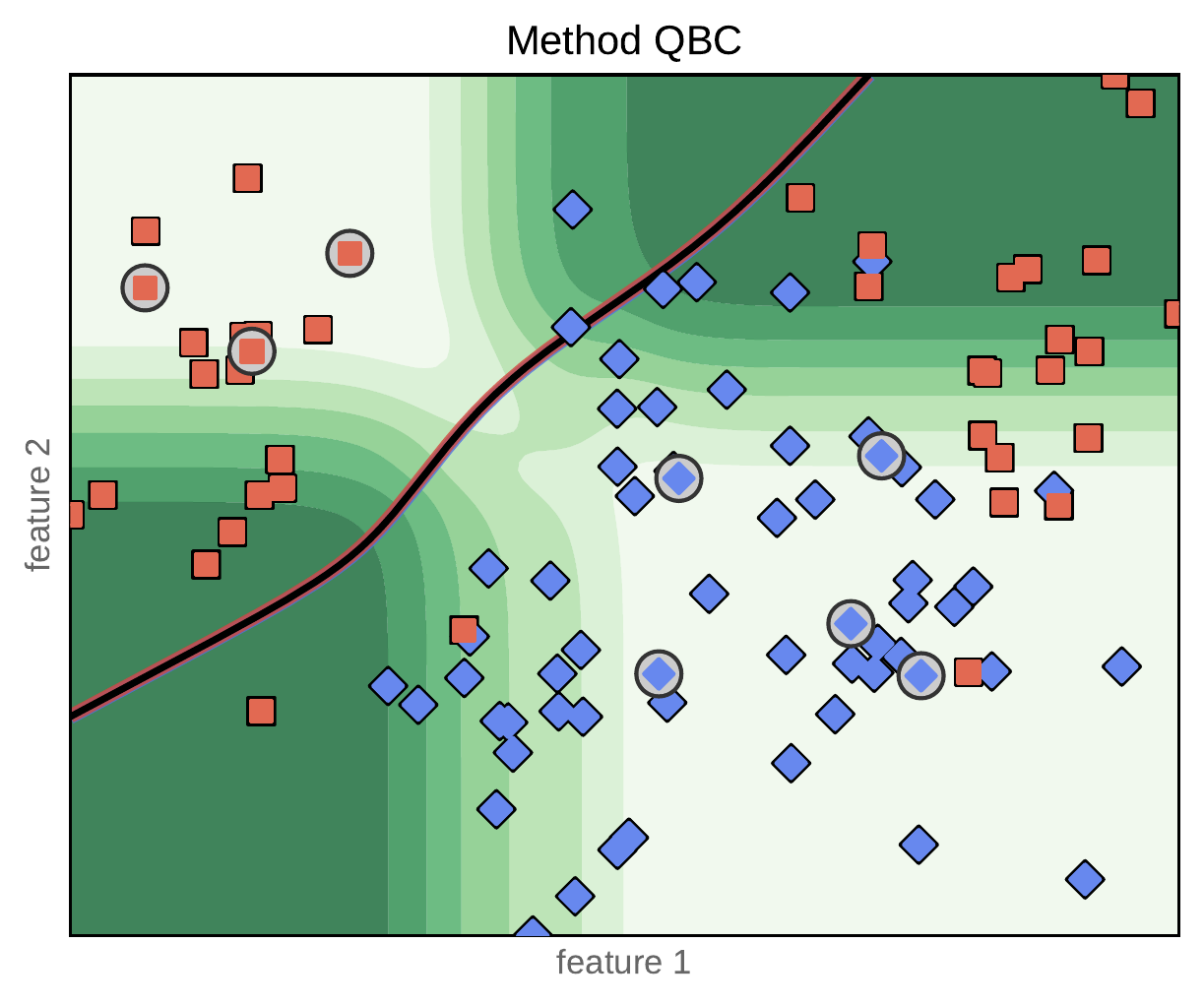}
    \includegraphics[width=.8\textwidth, trim=-3.7mm 0 -1.5mm 0]{fig/usefulness_plots_legend.pdf}
    \caption{Visualization of acquisition behavior for different selection strategies. The green color indicates how useful a selection strategy considers a region. The usefulness depends on the selection criterion of the strategy. The eight labels have been randomly selected and are similar for all strategies to emphasize how different selection strategies assess the usefulness of different regions.}
    \label{fig:usefulness-rand}
\end{figure*}

\clearpage
\newpage
\subsection{Learning Curves}
\begin{figure*}[h!]
	\centering
	\includegraphics[width=.4\textwidth]{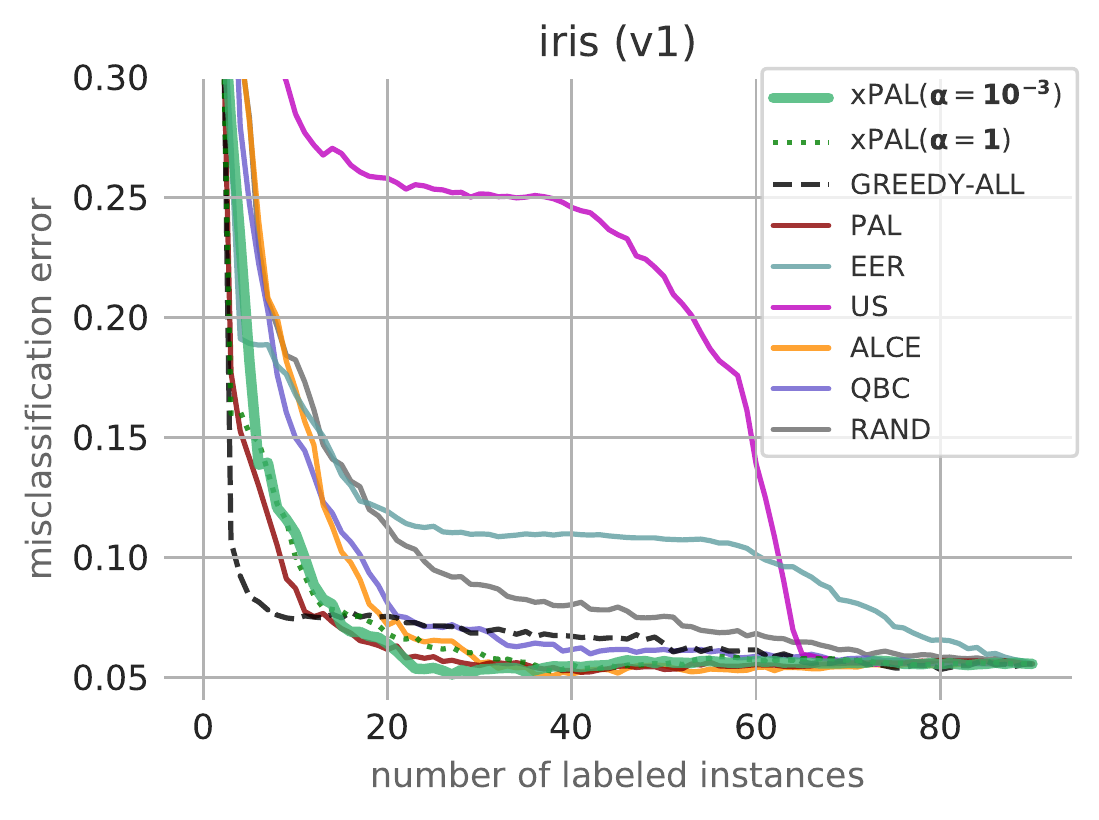}
	\includegraphics[width=.4\textwidth]{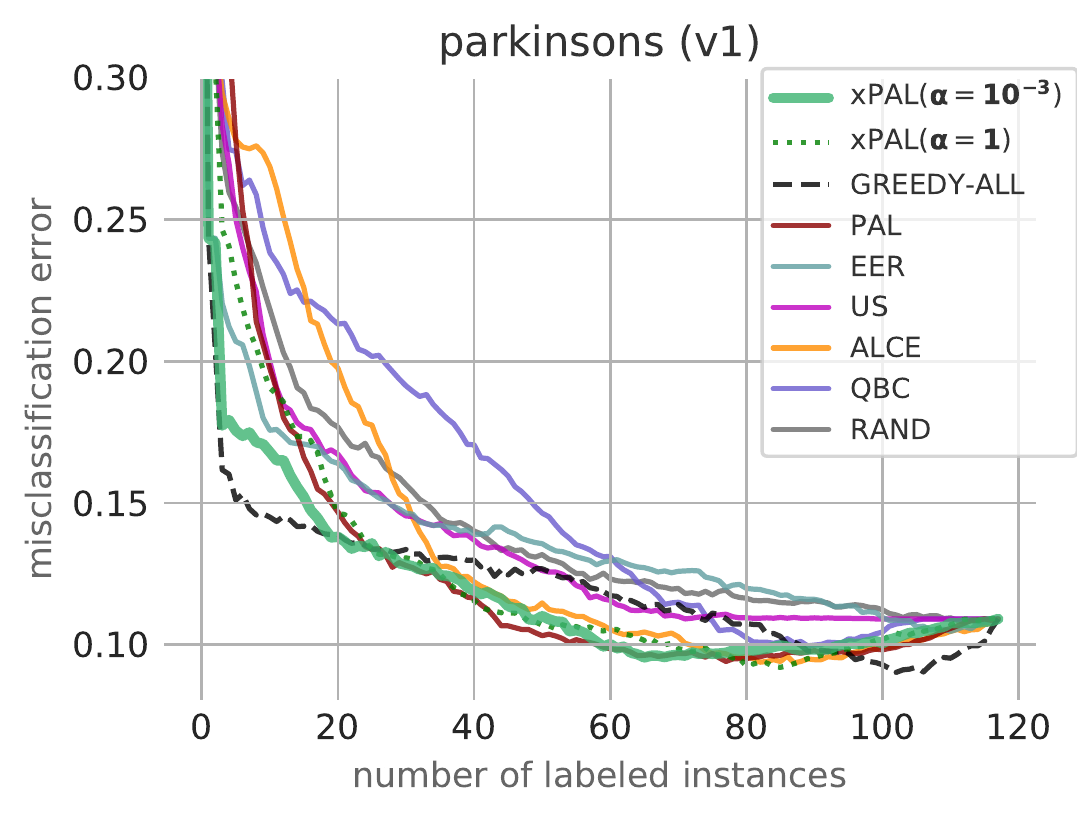}
	\includegraphics[width=.4\textwidth]{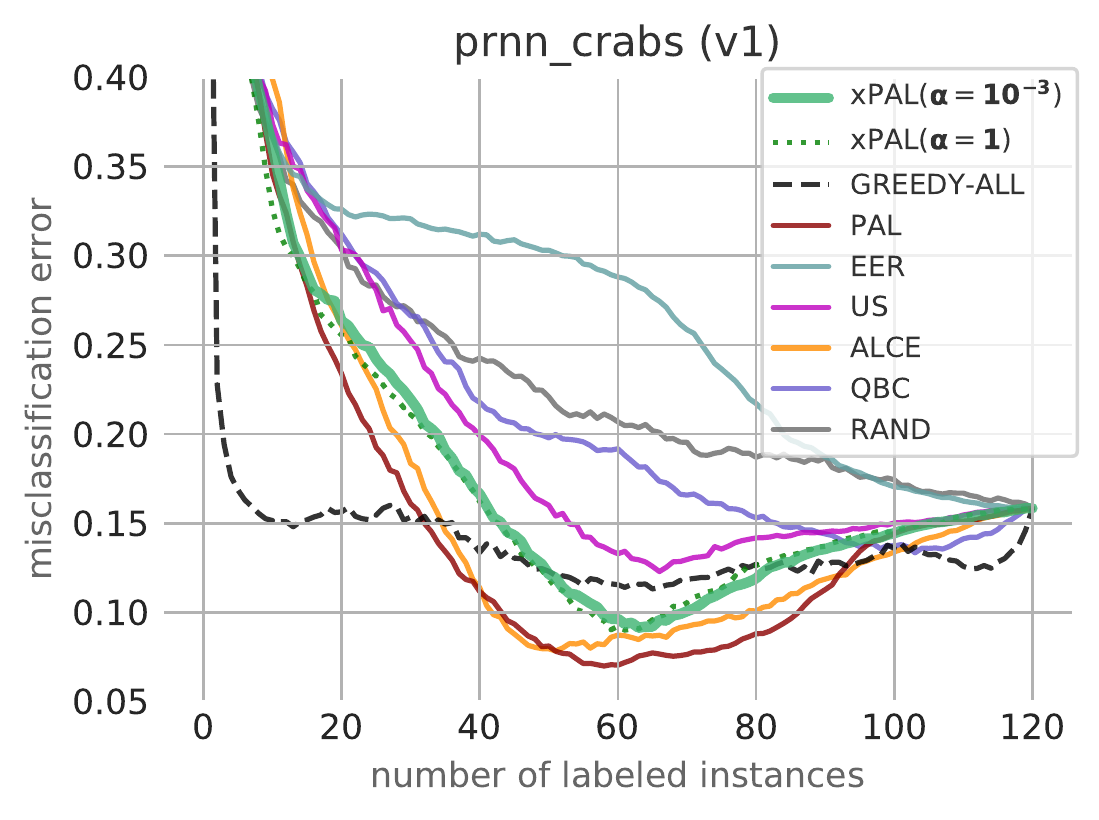}
	\includegraphics[width=.4\textwidth]{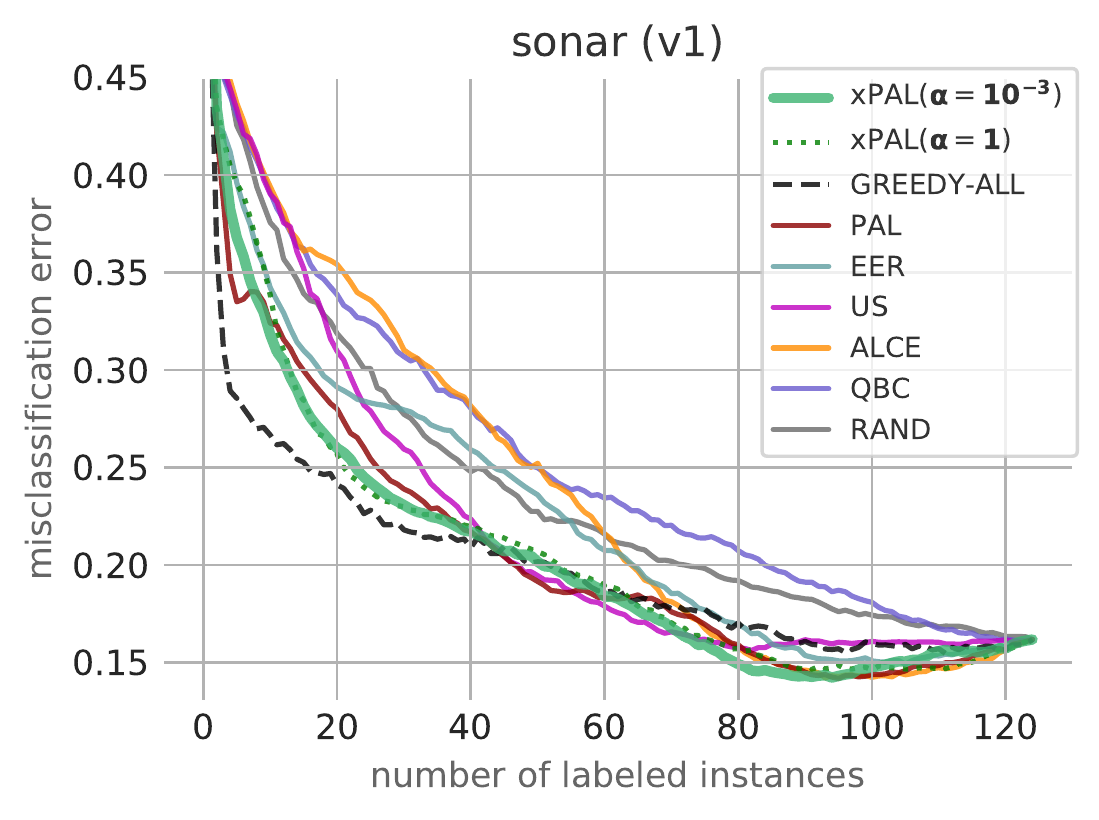}
	\includegraphics[width=.4\textwidth]{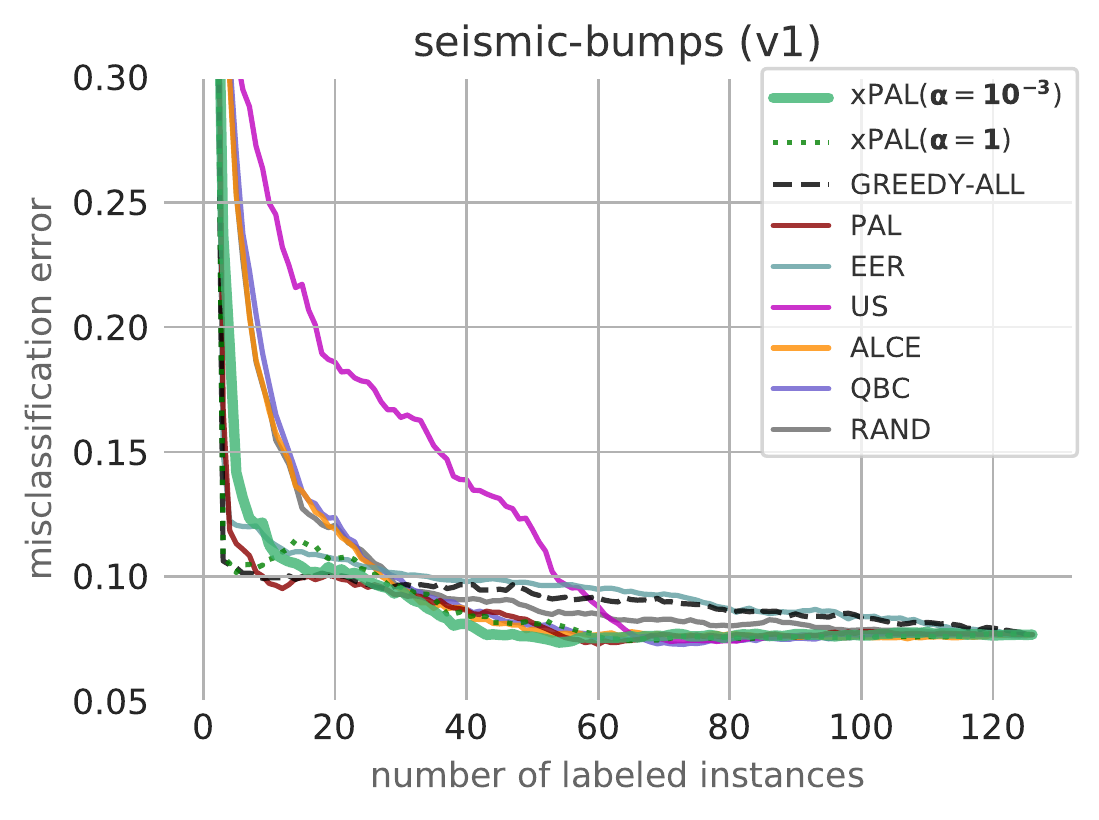}
	\includegraphics[width=.4\textwidth]{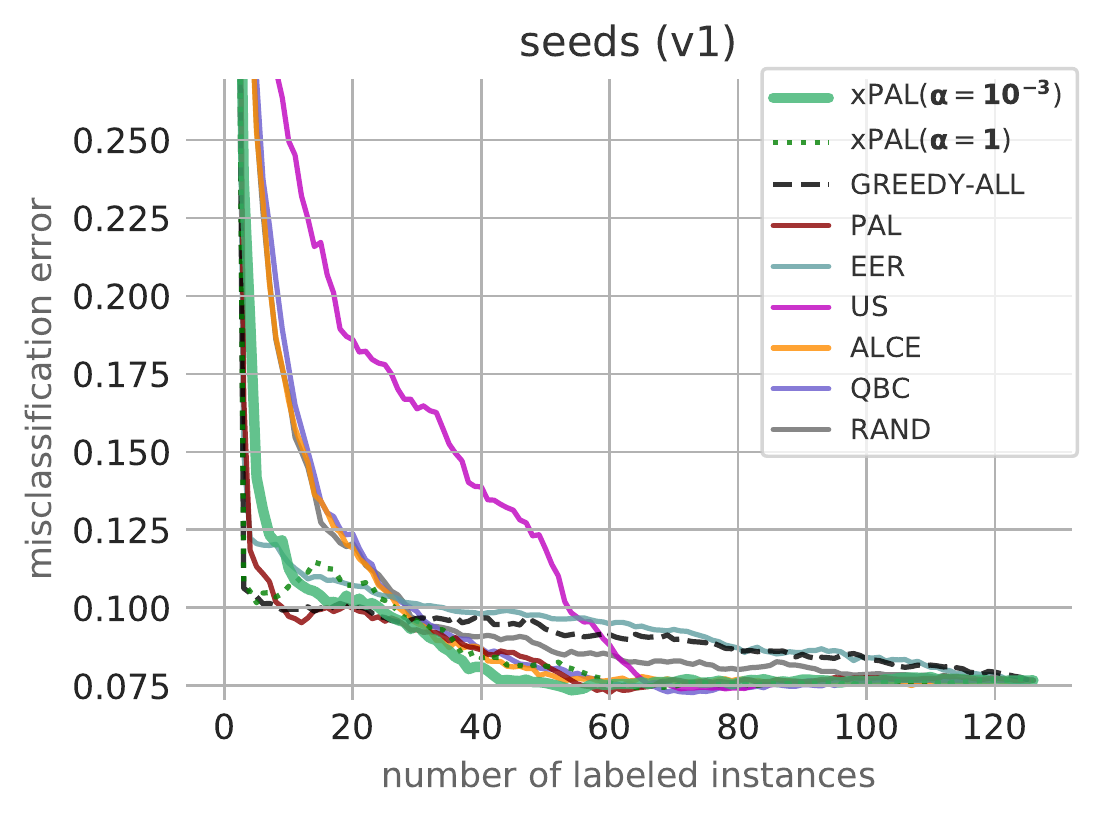}
	\includegraphics[width=.4\textwidth]{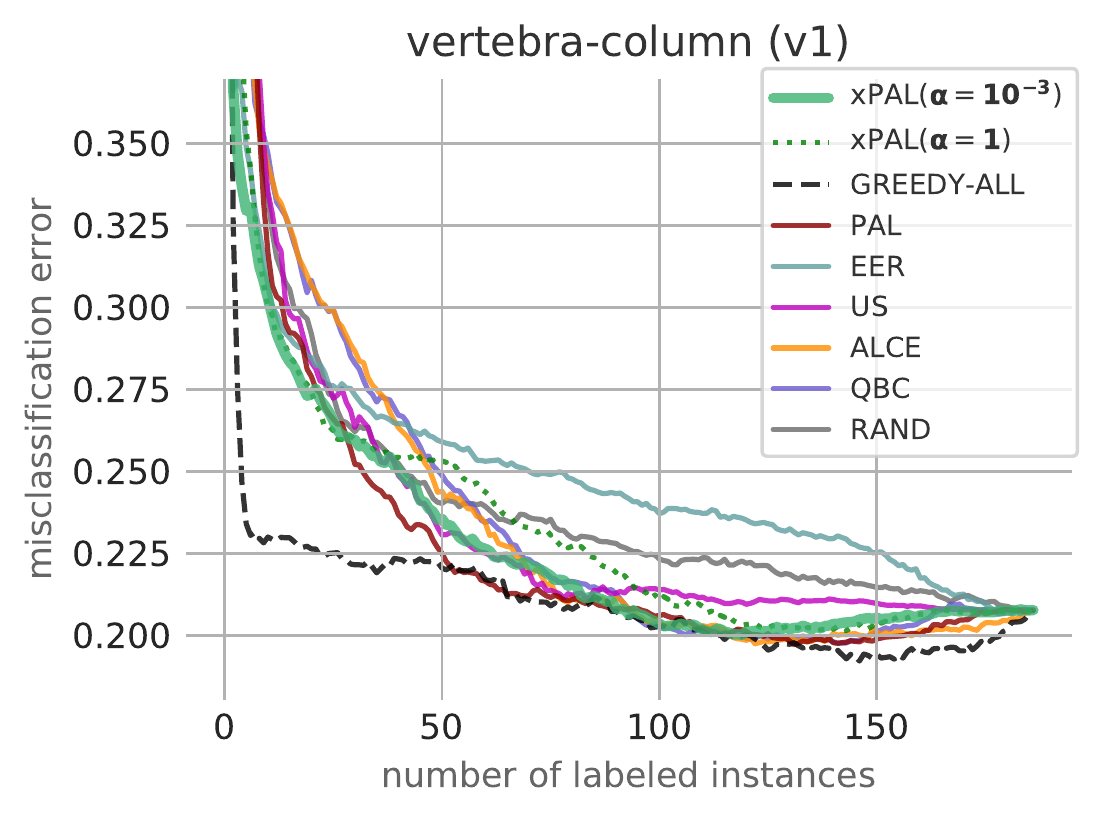}
	\includegraphics[width=.4\textwidth]{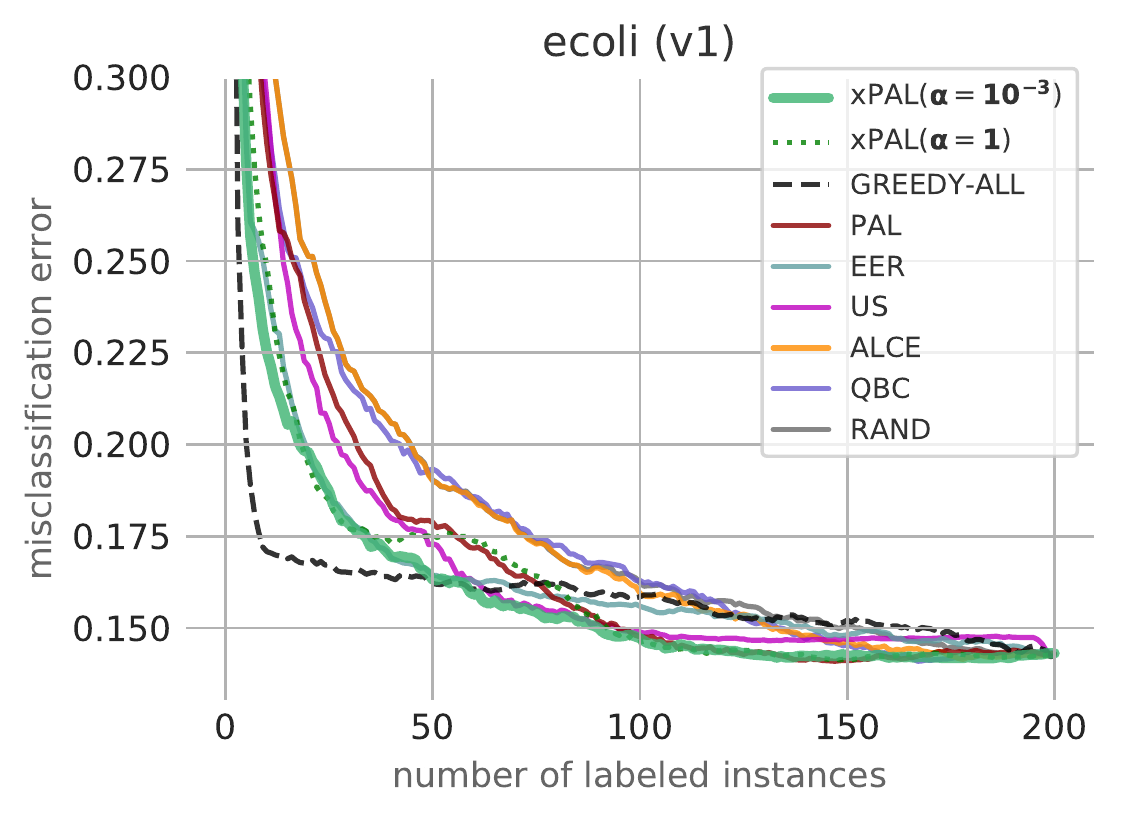}
	\caption{Mean accuracy learning curves comparing xPAL to its competitors. High values and fast convergence is considered best.}
	\label{fig:learning-curves2}
\end{figure*}

\begin{figure*}[p!]
	\centering
	\includegraphics[width=.4\textwidth]{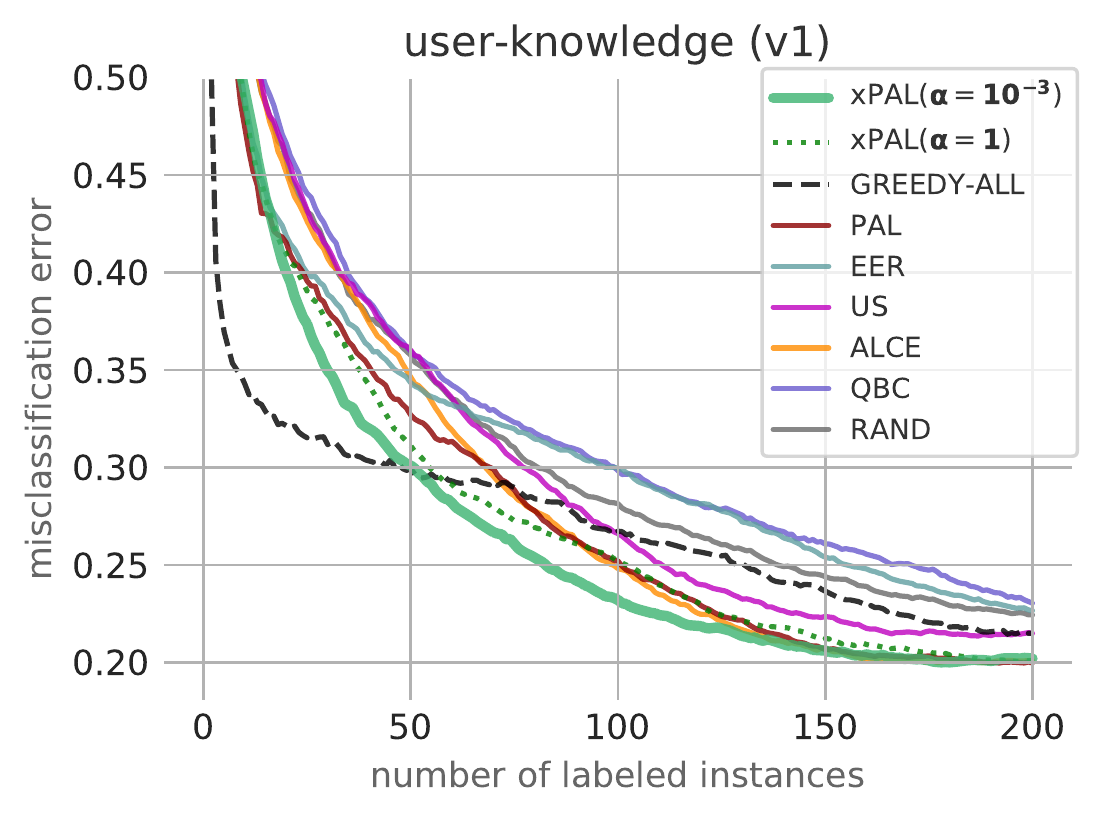}
	\includegraphics[width=.4\textwidth]{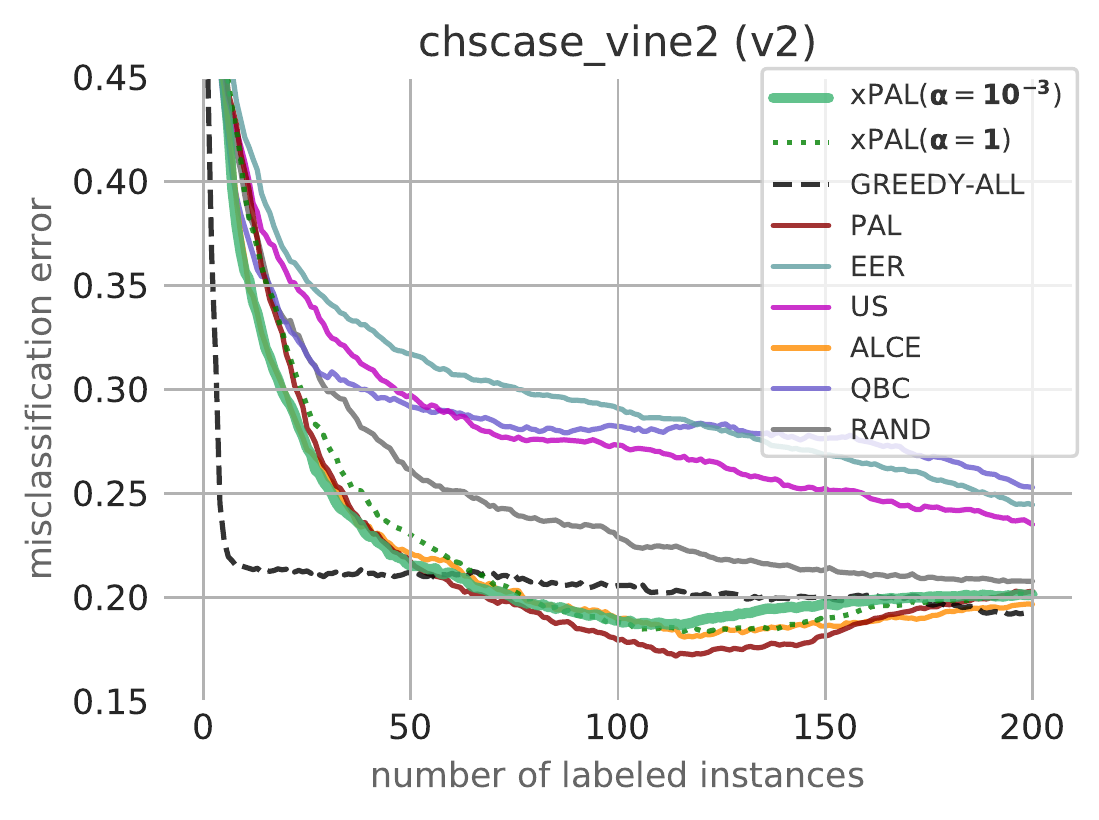}
	\includegraphics[width=.4\textwidth]{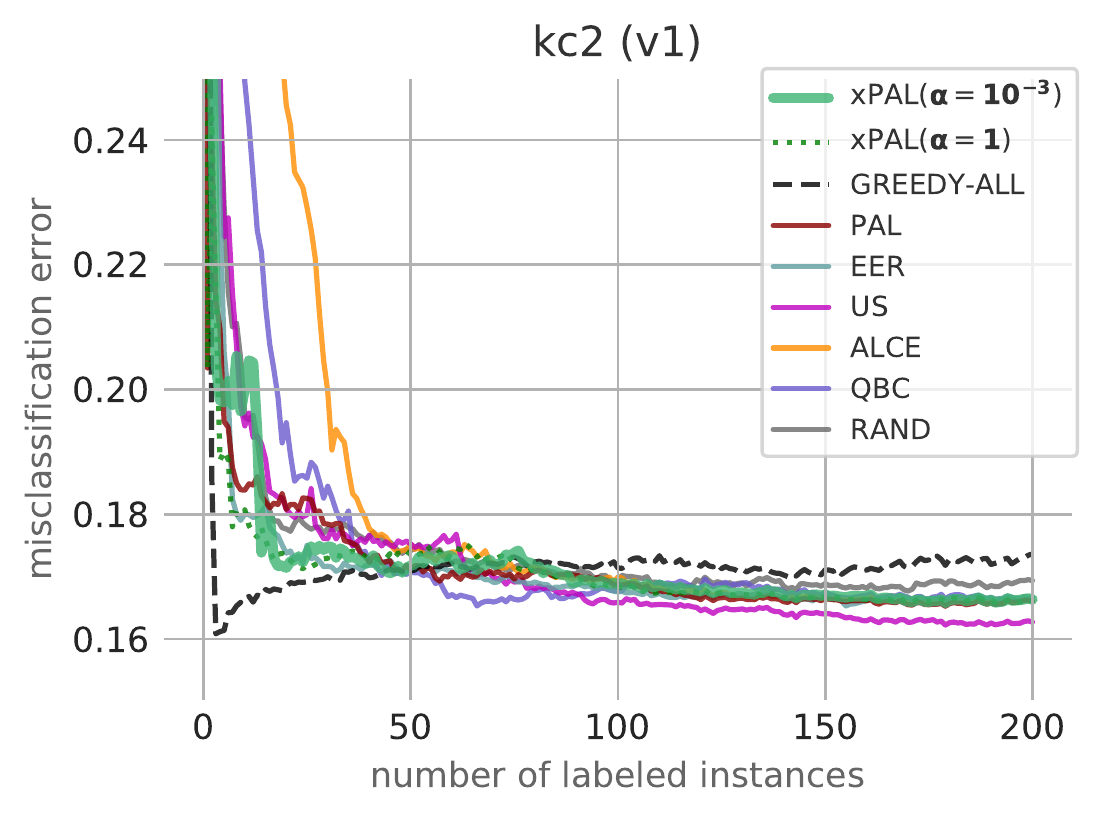}
	\includegraphics[width=.4\textwidth]{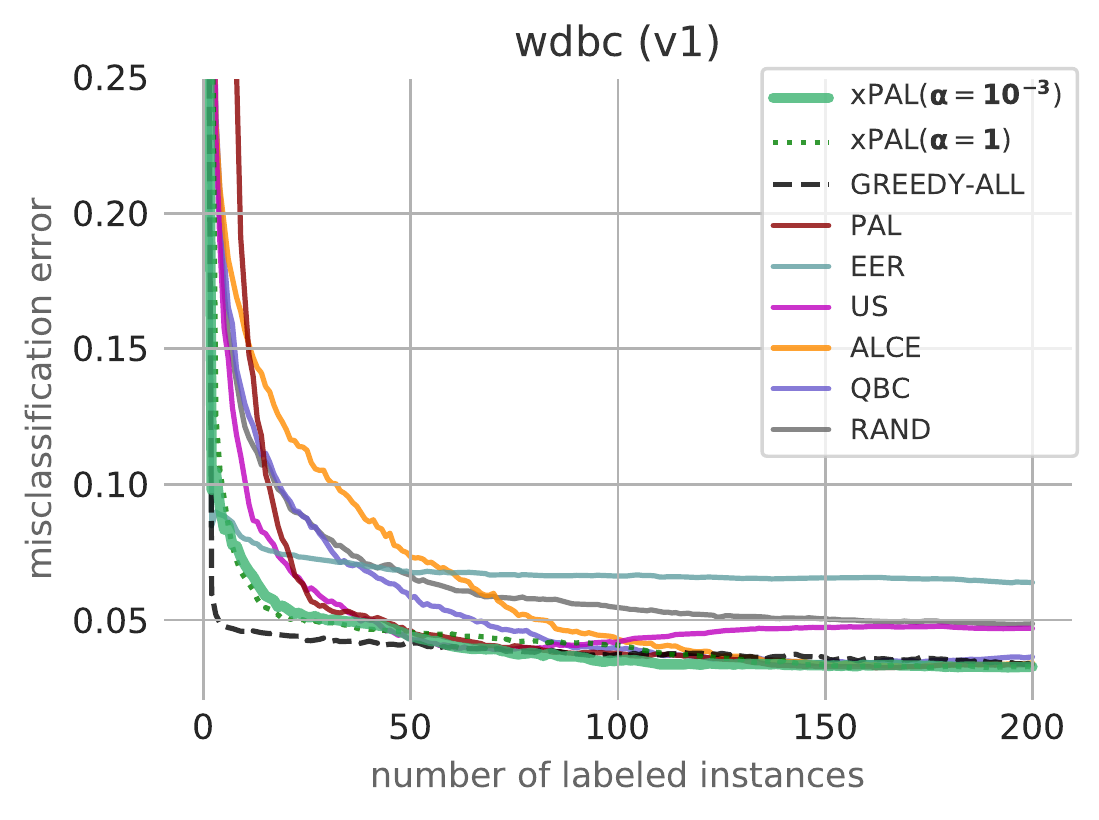}
	\includegraphics[width=.4\textwidth]{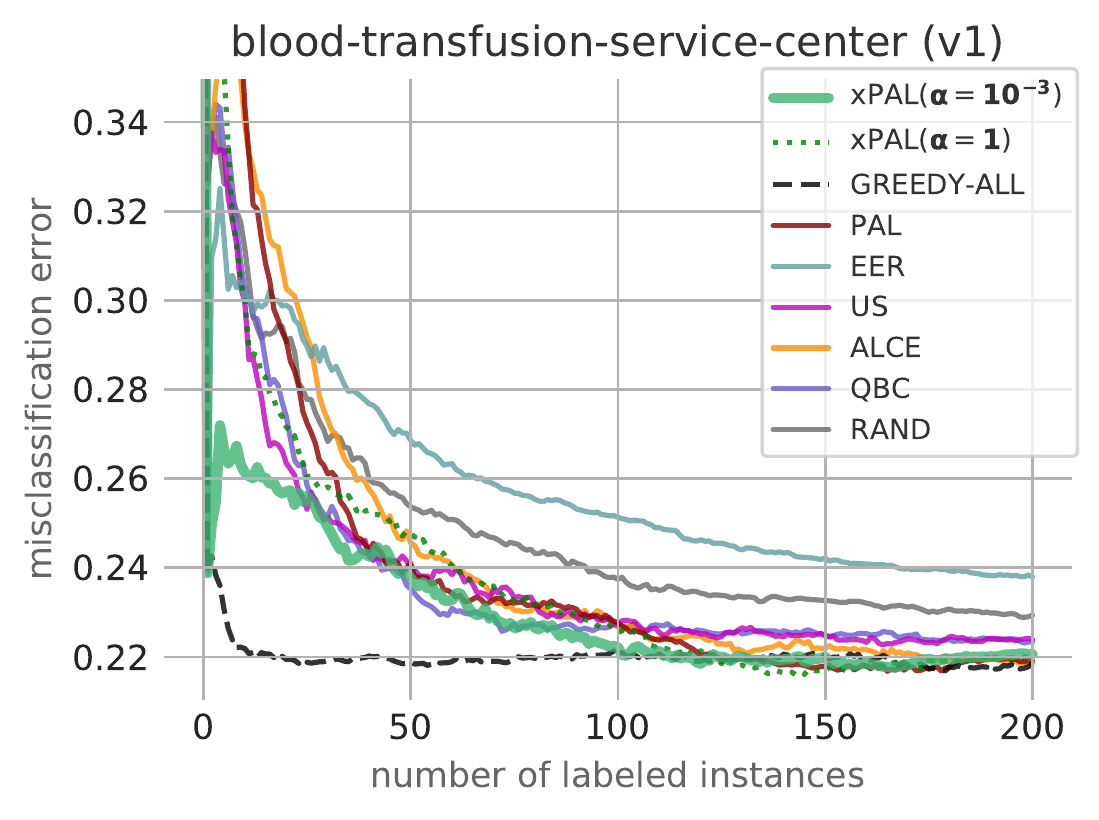}
	\includegraphics[width=.4\textwidth]{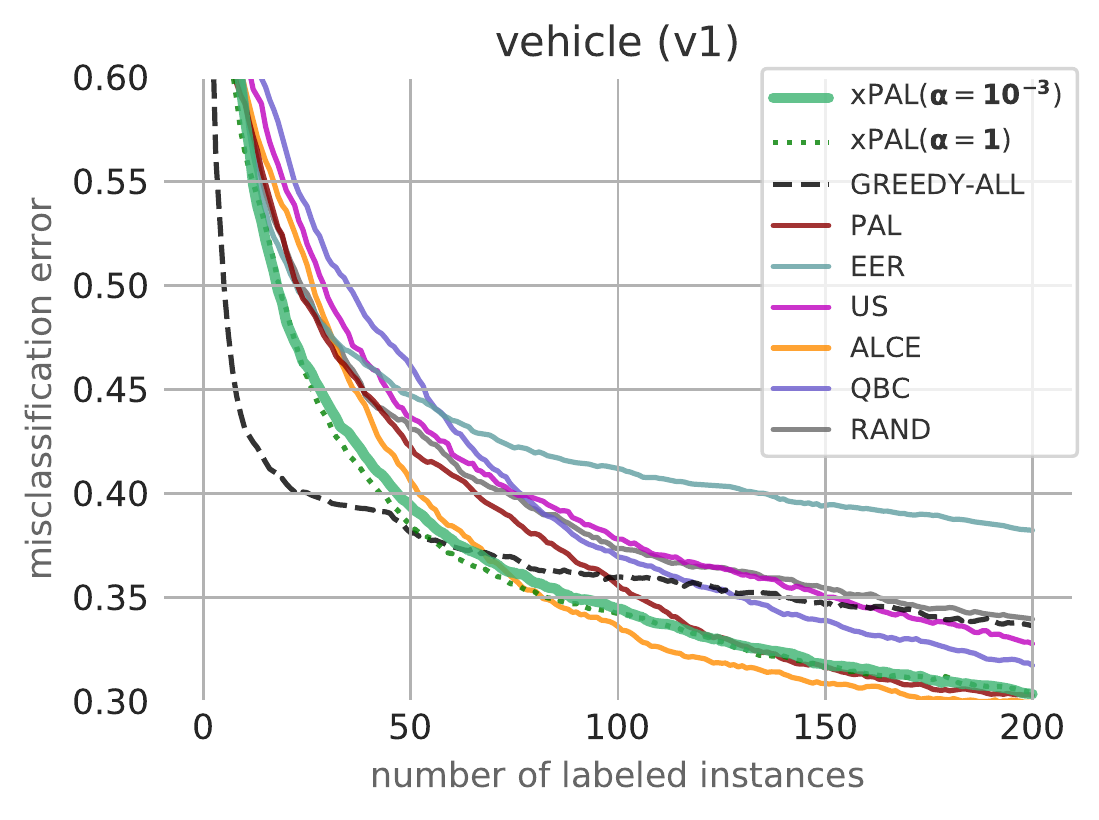}
	\includegraphics[width=.4\textwidth]{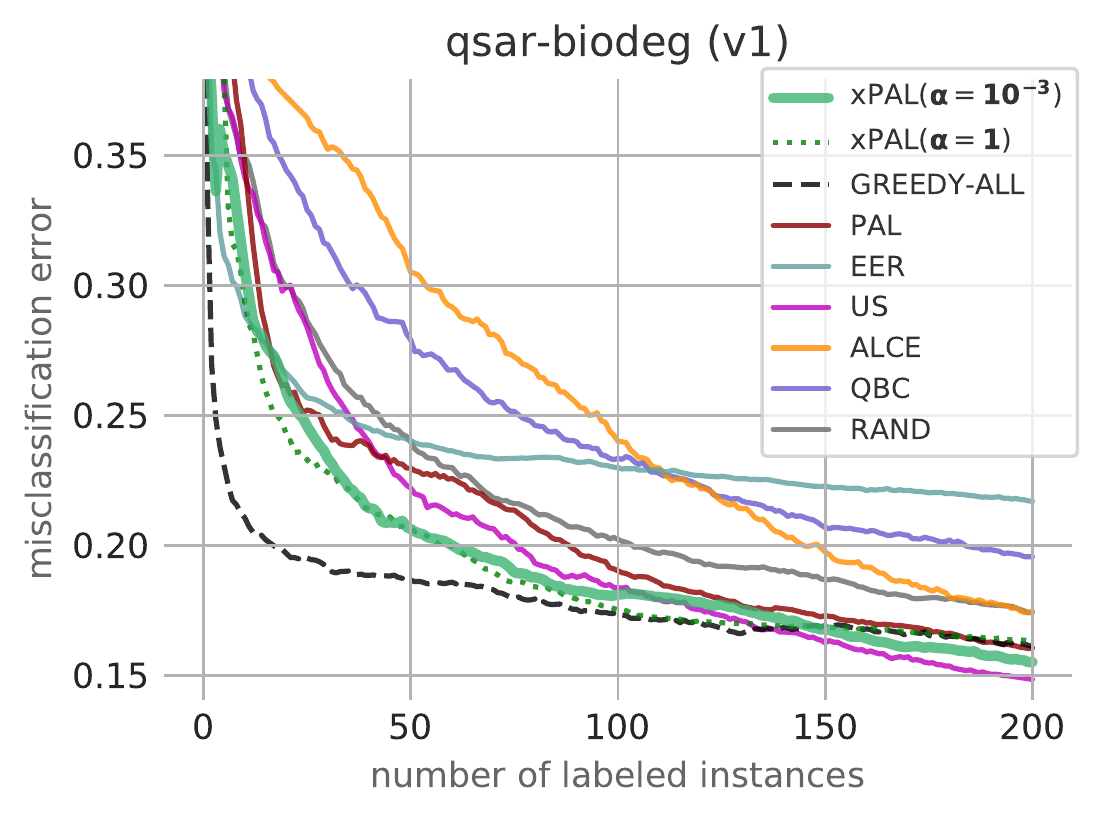}
	\includegraphics[width=.4\textwidth]{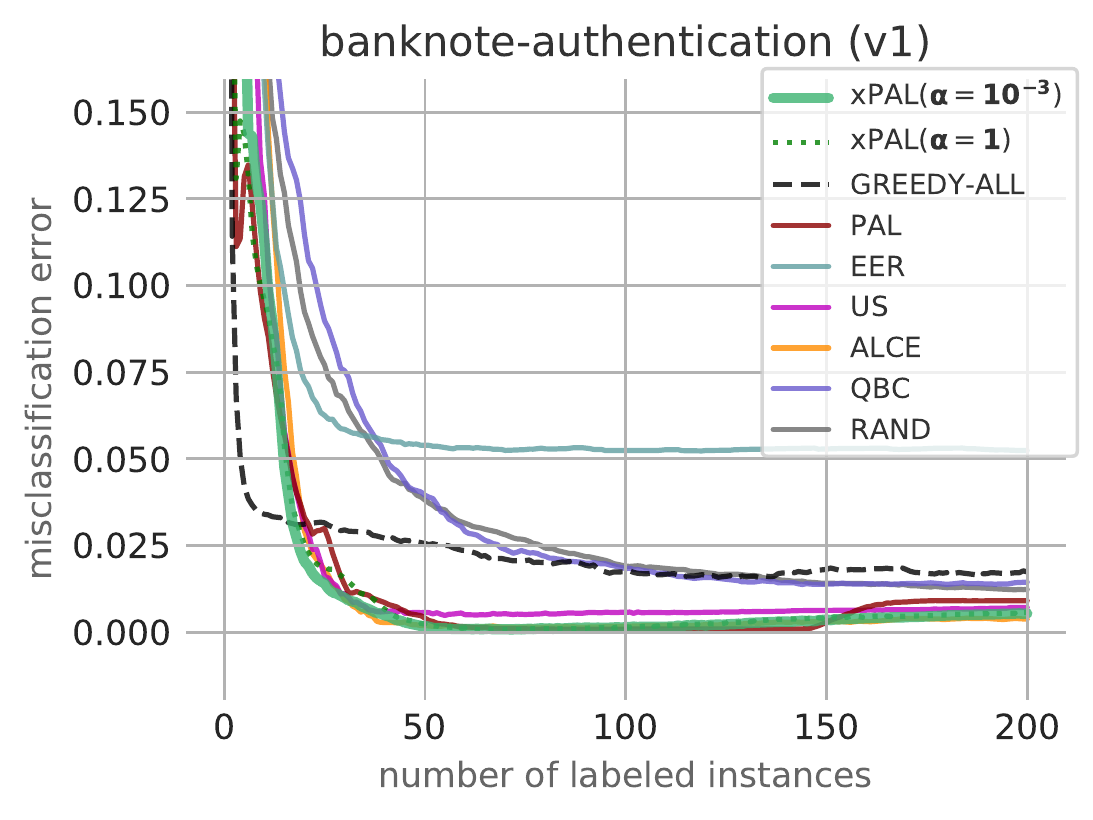}
	\caption{Mean accuracy learning curves comparing xPAL to its competitors. High values and fast convergence is considered best.}
	\label{fig:learning-curves3}
\end{figure*}
\begin{figure*}[p!]
	\centering
    \includegraphics[width=.4\textwidth]{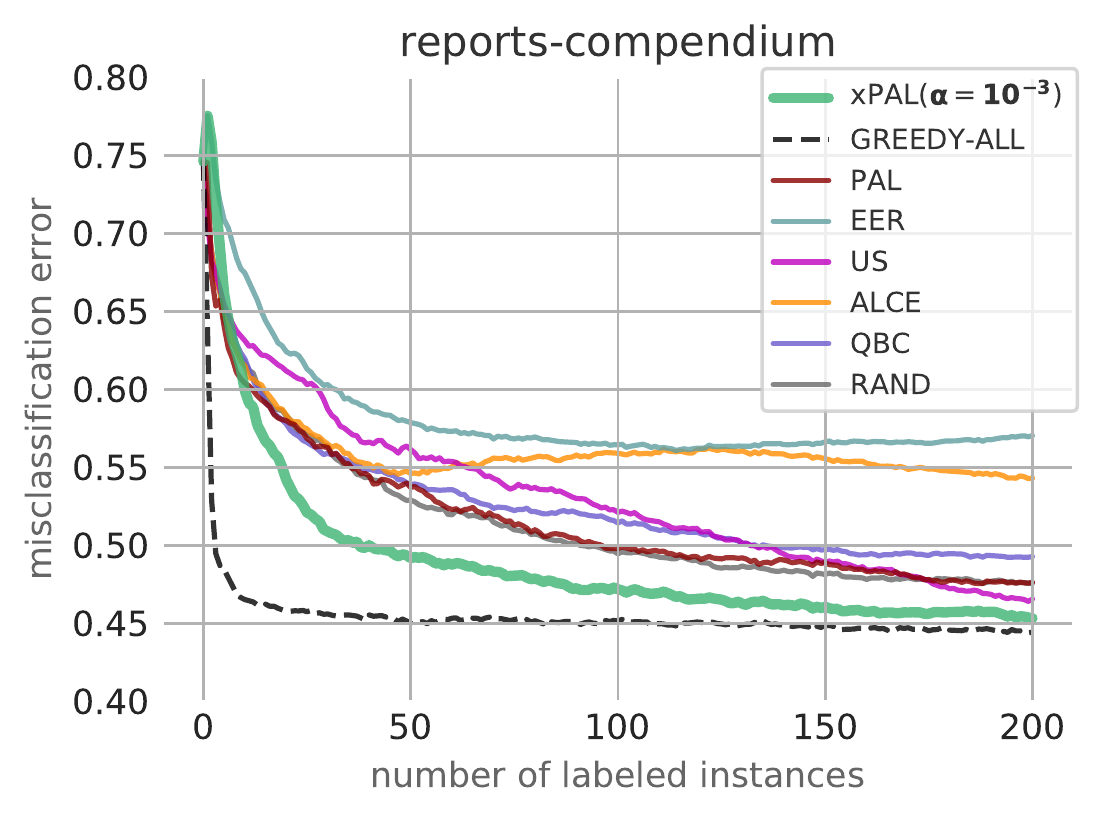}
    \includegraphics[width=.4\textwidth]{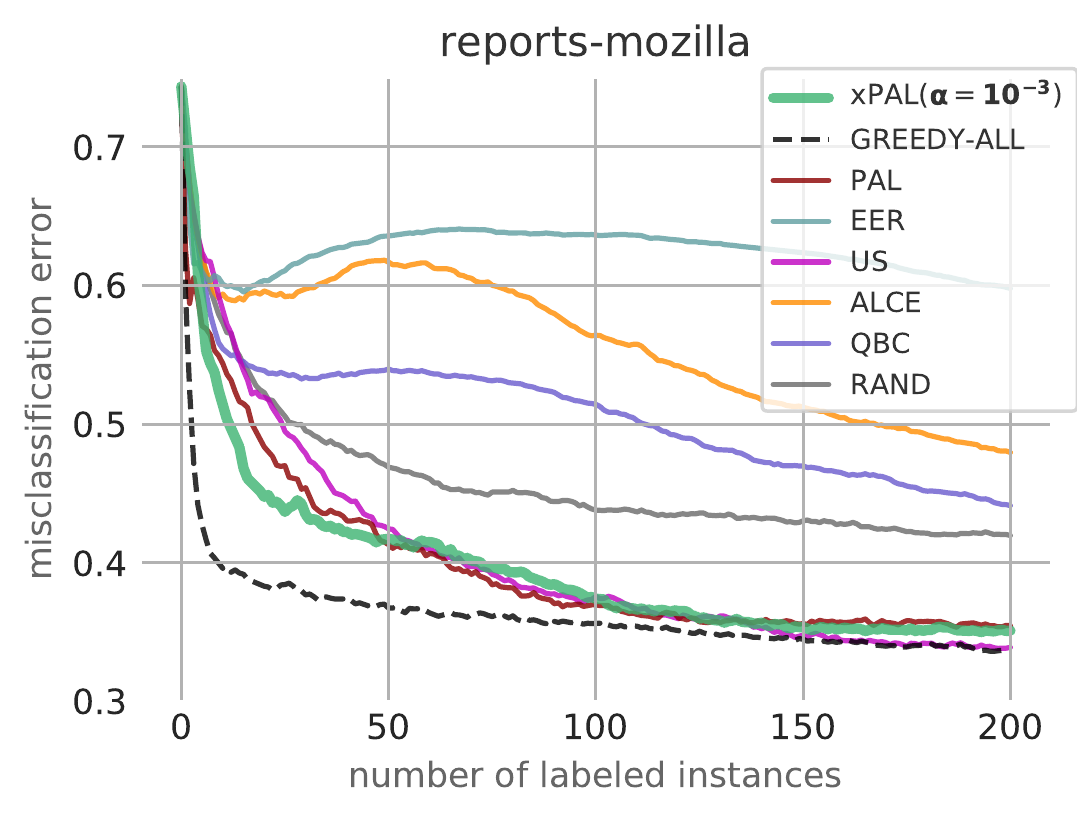}
    \includegraphics[width=.4\textwidth]{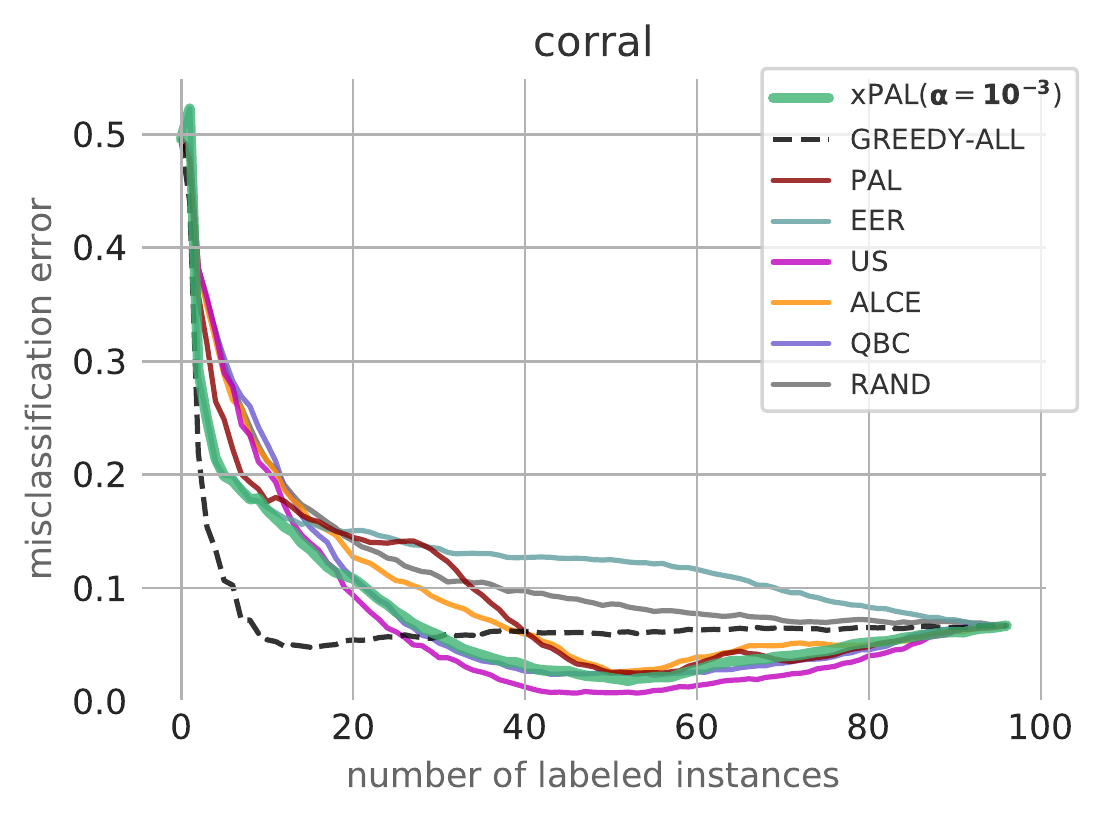}
    \includegraphics[width=.4\textwidth]{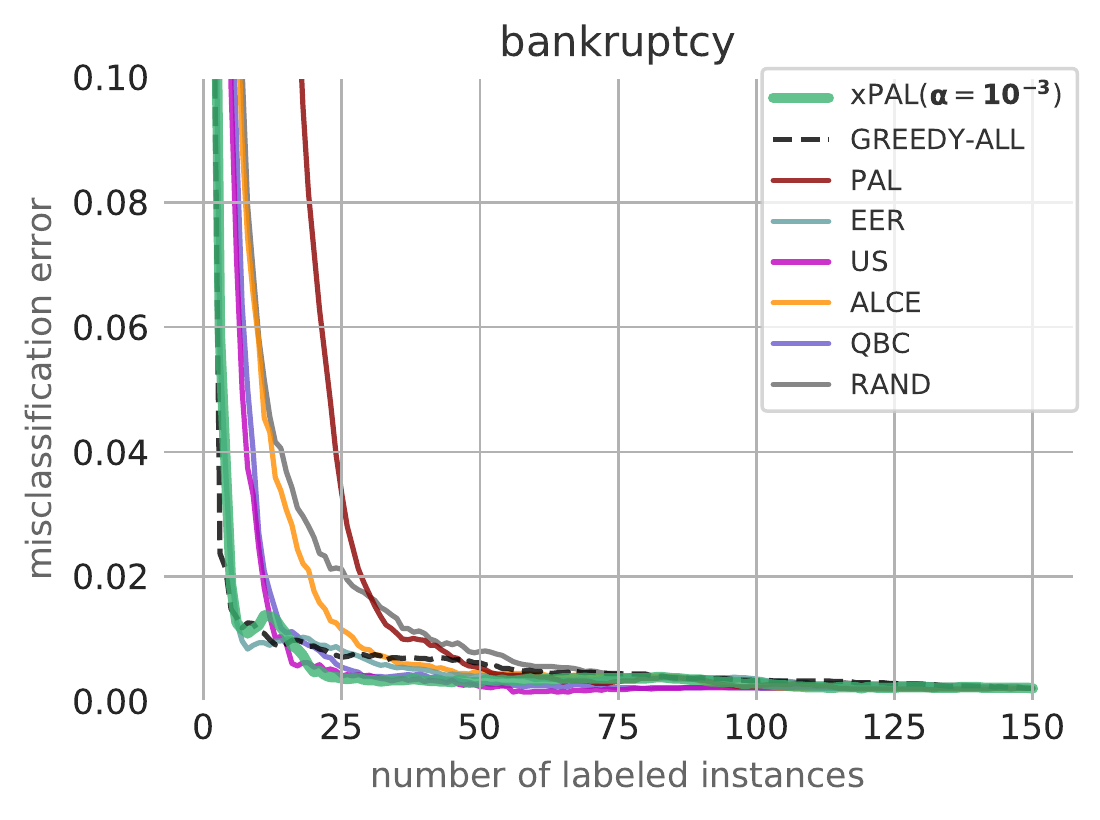}  %
    \includegraphics[width=.4\textwidth]{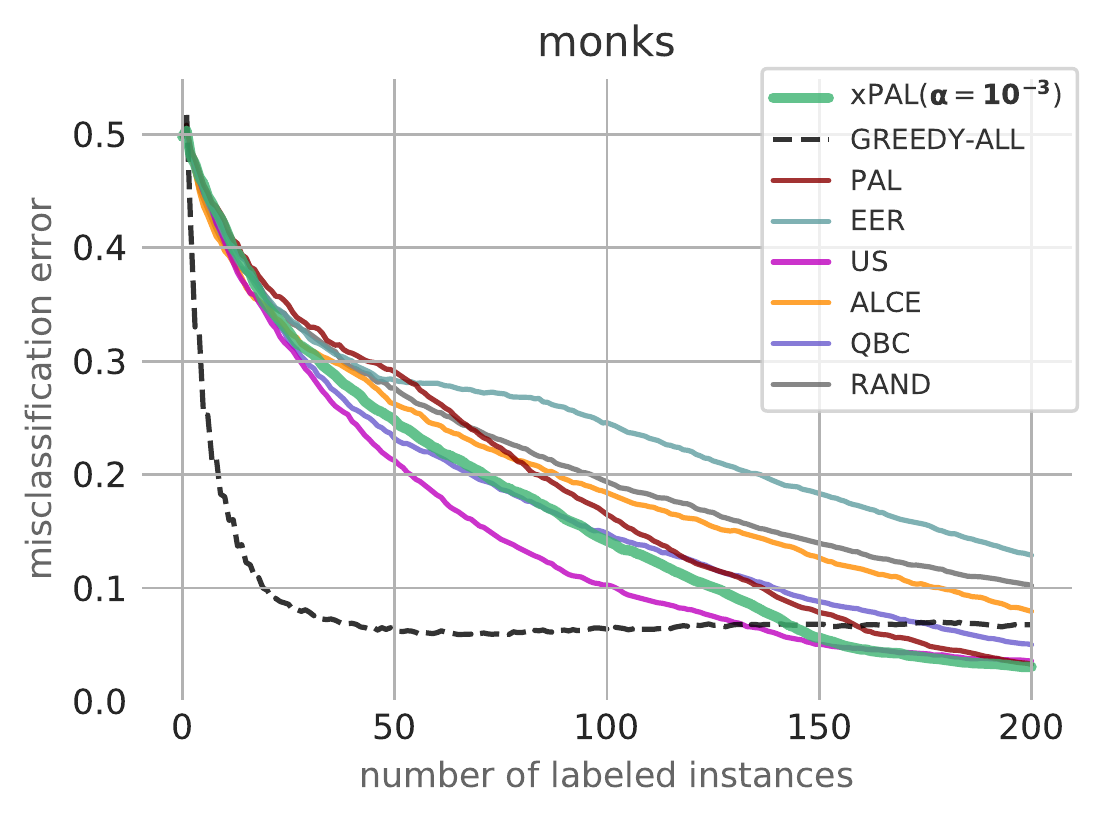}
    \includegraphics[width=.4\textwidth]{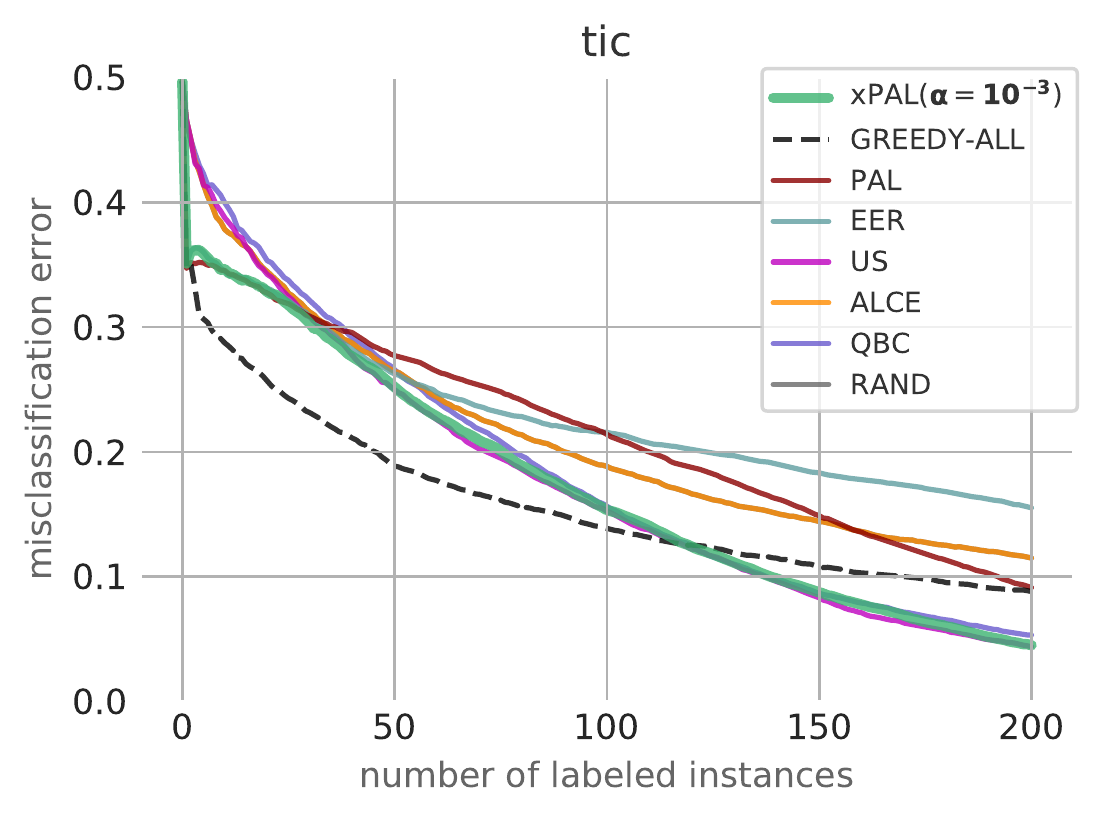}
    \includegraphics[width=.4\textwidth]{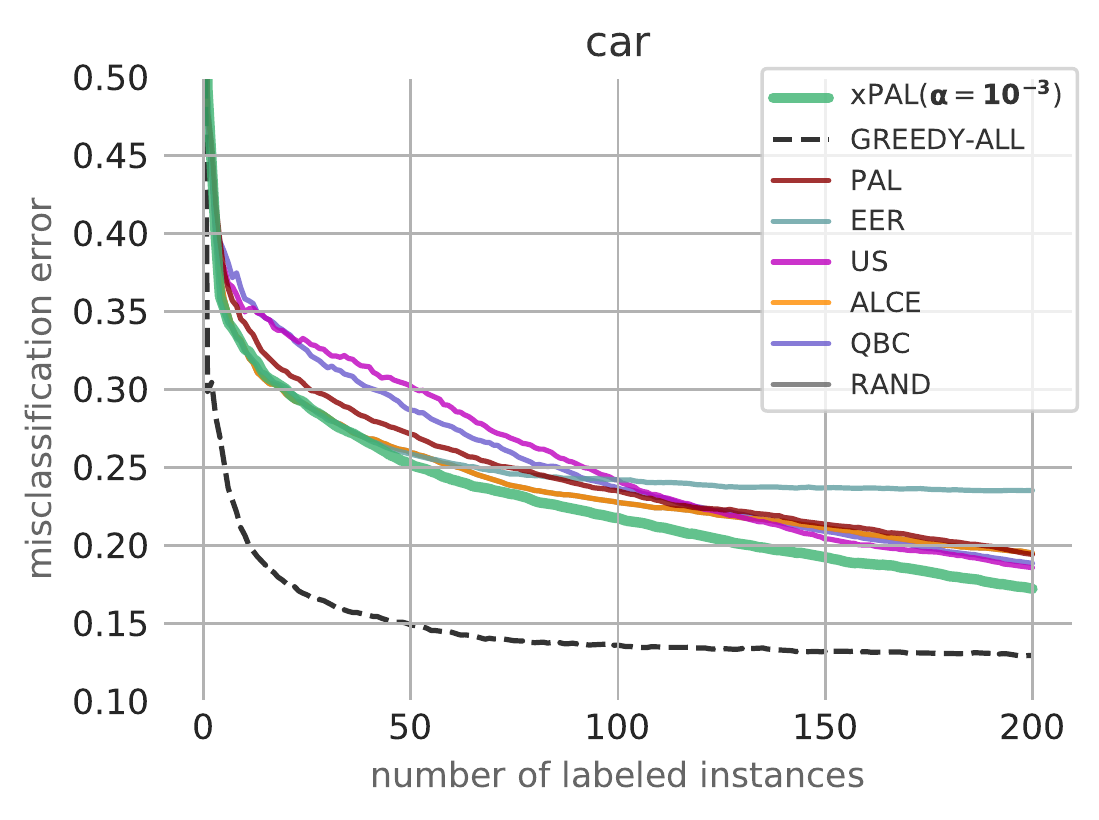}
	\caption{Mean accuracy learning curves comparing xPAL to its competitors. High values and fast convergence is considered best.}
	\label{fig:learning-curves4}
\end{figure*}

\clearpage
\newpage
\subsection{Area Under The Learning Curve}
Table~\ref{tab:aulc} describes the averaged area under the learning curve including standard deviations and significance testing with the Wilcoxon signed rank test. The notation is similar to the one from the paper.

\begin{table}[h]
    \centering
    \tiny
    \begin{tabular}{llllllll}
        \toprule
        {} &          xPAL($\mathbf{10^{-3}}$) &                                        PAL &                     US &                                       ALCE &                                 QBC &                              EER &                       RAND \\
        \midrule
        iris (v1)                             &  .084 (\tpm.022) &  .080 (\tpm.022)\(\dagger\dagger\dagger\) &  .199 (\tpm.096)*** &                        .096 (\tpm.024)*** &                 .099 (\tpm.024)*** &                        .123 (\tpm.050)*** &          .113 (\tpm.029)*** \\
        wine (v1)                             &  .067 (\tpm.017) &                        .079 (\tpm.019)*** &  .077 (\tpm.020)*** &                        .077 (\tpm.017)*** &                 .082 (\tpm.017)*** &                           .066 (\tpm.019) &          .084 (\tpm.022)*** \\
        parkinsons (v1)                       &  .122 (\tpm.028) &                        .131 (\tpm.025)*** &  .141 (\tpm.032)*** &                        .141 (\tpm.025)*** &                 .156 (\tpm.030)*** &                        .140 (\tpm.030)*** &          .147 (\tpm.027)*** \\
        prnn\_crabs (v1)                       &  .184 (\tpm.038) &  .162 (\tpm.030)\(\dagger\dagger\dagger\) &  .207 (\tpm.042)*** &  .170 (\tpm.030)\(\dagger\dagger\dagger\) &                 .220 (\tpm.034)*** &                        .270 (\tpm.047)*** &          .237 (\tpm.040)*** \\
        sonar (v1)                            &  .206 (\tpm.031) &                           .207 (\tpm.029) &  .222 (\tpm.028)*** &                        .239 (\tpm.028)*** &                 .255 (\tpm.029)*** &                        .227 (\tpm.040)*** &          .240 (\tpm.033)*** \\
        seeds (v1)                            &  .097 (\tpm.021) &                .096 (\tpm.021)\(\dagger\) &  .133 (\tpm.053)*** &                        .107 (\tpm.023)*** &                 .108 (\tpm.023)*** &                        .106 (\tpm.026)*** &          .111 (\tpm.022)*** \\
        seismic-bumps (v1)                    &  .097 (\tpm.021) &                .096 (\tpm.021)\(\dagger\) &  .133 (\tpm.053)*** &                        .107 (\tpm.023)*** &                 .108 (\tpm.023)*** &                        .106 (\tpm.026)*** &          .111 (\tpm.022)*** \\
        glass (v1)                            &  .378 (\tpm.037) &                        .406 (\tpm.036)*** &  .411 (\tpm.041)*** &                        .411 (\tpm.037)*** &                 .414 (\tpm.038)*** &                        .437 (\tpm.042)*** &          .423 (\tpm.041)*** \\
        vertebra-column (v1)                  &  .231 (\tpm.031) &                         .235 (\tpm.032)** &  .239 (\tpm.033)*** &                        .239 (\tpm.031)*** &                 .240 (\tpm.029)*** &                        .252 (\tpm.033)*** &          .246 (\tpm.033)*** \\
        ecoli (v1)                            &  .166 (\tpm.024) &                        .177 (\tpm.022)*** &  .178 (\tpm.023)*** &                        .189 (\tpm.026)*** &                 .187 (\tpm.021)*** &                          .172 (\tpm.025)* &          .191 (\tpm.026)*** \\
        ionosphere (v1)                       &  .152 (\tpm.028) &                        .160 (\tpm.029)*** &  .172 (\tpm.028)*** &                        .180 (\tpm.035)*** &                 .168 (\tpm.027)*** &  .140 (\tpm.023)\(\dagger\dagger\dagger\) &          .194 (\tpm.036)*** \\
        user-knowledge (v1)                   &  .273 (\tpm.021) &                        .286 (\tpm.024)*** &  .307 (\tpm.032)*** &                        .294 (\tpm.021)*** &                 .330 (\tpm.024)*** &                        .317 (\tpm.026)*** &          .317 (\tpm.027)*** \\
        chscase\_vine2 (v2)                    &  .223 (\tpm.021) &                           .221 (\tpm.021) &  .287 (\tpm.063)*** &                           .221 (\tpm.023) &                 .294 (\tpm.030)*** &                        .303 (\tpm.039)*** &          .253 (\tpm.023)*** \\
        kc2 (v1)                              &  .174 (\tpm.018) &                          .173 (\tpm.017)* &     .174 (\tpm.019) &                        .185 (\tpm.024)*** &                 .179 (\tpm.019)*** &                           .173 (\tpm.019) &           .177 (\tpm.019)** \\
        wdbc (v1)                             &  .045 (\tpm.009) &                        .058 (\tpm.009)*** &  .058 (\tpm.011)*** &                        .066 (\tpm.014)*** &                 .059 (\tpm.012)*** &                        .072 (\tpm.019)*** &          .069 (\tpm.014)*** \\
        balance-scale (v1)                    &  .188 (\tpm.017) &                         .194 (\tpm.016)** &  .199 (\tpm.017)*** &                        .200 (\tpm.015)*** &  .182 (\tpm.016)\(\dagger\dagger\) &                        .229 (\tpm.022)*** &           .195 (\tpm.018)** \\
        blood-transfusion-service-center (v1) &  .231 (\tpm.017) &                        .245 (\tpm.021)*** &  .239 (\tpm.017)*** &                        .244 (\tpm.022)*** &                 .239 (\tpm.018)*** &                        .259 (\tpm.029)*** &          .250 (\tpm.019)*** \\
        diabetes (v1)                         &  .303 (\tpm.020) &                        .311 (\tpm.018)*** &     .301 (\tpm.021) &                          .310 (\tpm.026)* &  .296 (\tpm.030)\(\dagger\dagger\) &                         .309 (\tpm.021)** &  .298 (\tpm.021)\(\dagger\) \\
        vehicle (v1)                          &  .375 (\tpm.018) &                        .387 (\tpm.016)*** &  .412 (\tpm.026)*** &                           .378 (\tpm.018) &                 .412 (\tpm.019)*** &                        .436 (\tpm.026)*** &          .409 (\tpm.023)*** \\
        qsar-biodeg (v1)                      &  .198 (\tpm.016) &                        .214 (\tpm.014)*** &  .206 (\tpm.019)*** &                        .261 (\tpm.033)*** &                 .254 (\tpm.030)*** &                        .239 (\tpm.025)*** &          .224 (\tpm.021)*** \\
        banknote-authentication (v1)          &  .019 (\tpm.003) &  .018 (\tpm.002)\(\dagger\dagger\dagger\) &  .025 (\tpm.005)*** &                        .024 (\tpm.005)*** &                 .048 (\tpm.010)*** &                        .070 (\tpm.022)*** &          .046 (\tpm.010)*** \\
        steel-plates-fault (v1)               &  .056 (\tpm.005) &                        .084 (\tpm.008)*** &  .085 (\tpm.017)*** &                        .113 (\tpm.013)*** &                 .192 (\tpm.030)*** &                        .127 (\tpm.013)*** &          .128 (\tpm.016)*** \\
        \midrule
        corral (v1)                 &  .079 (\tpm.023) &  .098 (\tpm.028)*** &                .076 (\tpm.020)\(\dagger\) &  .098 (\tpm.027)*** &  .087 (\tpm.022)*** &  .131 (\tpm.032)*** &  .121 (\tpm.032)*** \\
        qualitative-bankruptcy (v1) &  .012 (\tpm.004) &  .048 (\tpm.012)*** &                        .016 (\tpm.006)*** &  .021 (\tpm.007)*** &  .017 (\tpm.006)*** &    .013 (\tpm.005)* &  .023 (\tpm.009)*** \\
        monks-problems-1 (v1)       &  .169 (\tpm.014) &  .190 (\tpm.020)*** &  .148 (\tpm.019)\(\dagger\dagger\dagger\) &  .206 (\tpm.022)*** &  .178 (\tpm.019)*** &  .248 (\tpm.026)*** &  .218 (\tpm.022)*** \\
        tic-tac-toe (v1)            &  .175 (\tpm.012) &  .217 (\tpm.019)*** &                           .177 (\tpm.014) &  .213 (\tpm.016)*** &  .186 (\tpm.015)*** &  .231 (\tpm.023)*** &  .213 (\tpm.016)*** \\
        car-evaluation (v1)         &  .232 (\tpm.010) &  .251 (\tpm.013)*** &                        .258 (\tpm.016)*** &  .243 (\tpm.014)*** &  .256 (\tpm.017)*** &  .258 (\tpm.014)*** &  .243 (\tpm.014)*** \\
        \midrule
        reports-mozilla    &  .397 (\tpm.027) &     .399 (\tpm.026) &   .404 (\tpm.020)** &  .559 (\tpm.068)*** &  .508 (\tpm.030)*** &  .625 (\tpm.070)*** &  .461 (\tpm.043)*** \\
        reports-compendium &  .490 (\tpm.021) &  .517 (\tpm.019)*** &  .532 (\tpm.022)*** &  .563 (\tpm.032)*** &  .528 (\tpm.022)*** &  .583 (\tpm.030)*** &  .516 (\tpm.023)*** \\
        \bottomrule
    \end{tabular}   
    \caption{Averaged area under the learning curve performances including standard deviations. Low values are considered best.}
    \label{tab:aulc}
\end{table}

\subsection{Detailed Ranking Plots for Different Parameters}
\begin{figure}[h]
    \centering
    \includegraphics[width=.9\textwidth]{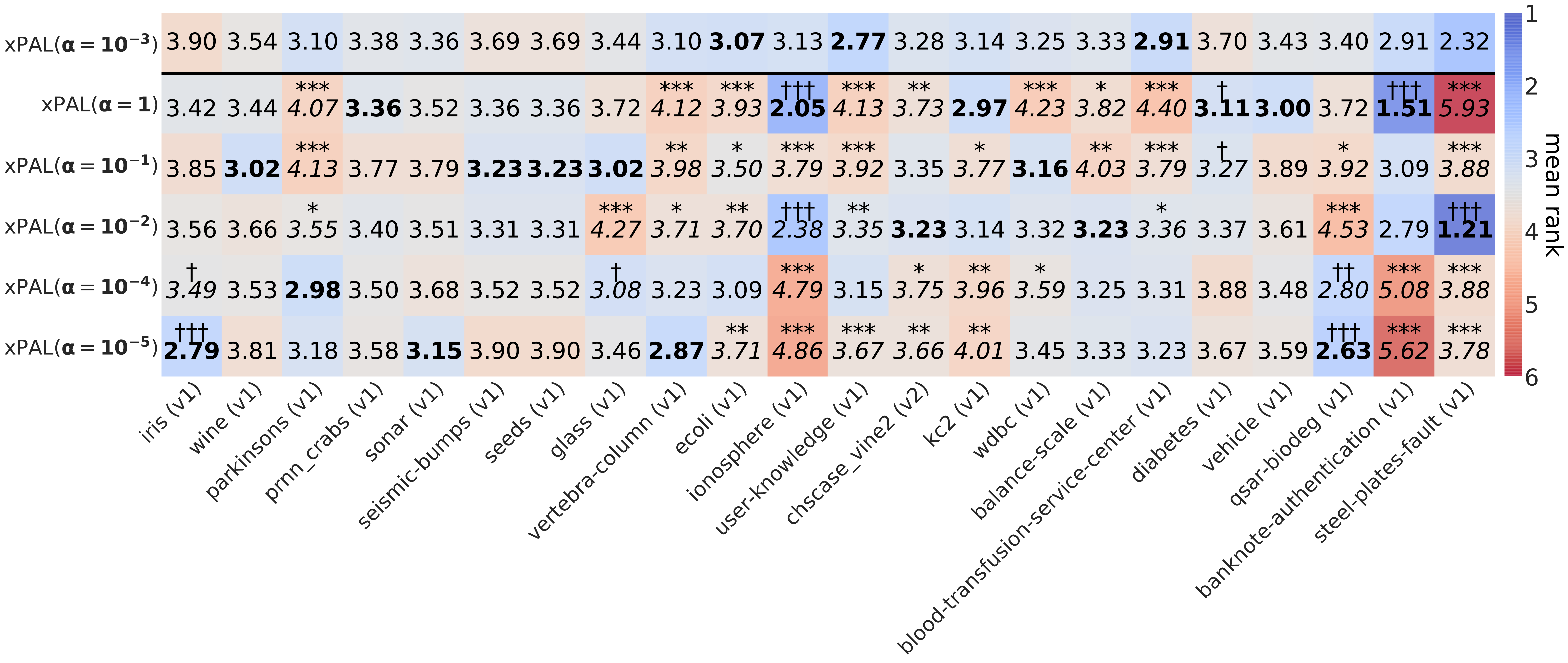}
    \caption{The mean rank for xPAL with different parameters and datasets across 100 repetitions. The best parameter is printed in bold. The Wilcoxon signed rank test shows pairwise significance between xPAL with $\prior = \vec{10^{-3}}$ and its competitor.}
    \label{fig:rank}
\end{figure}

Figure~\ref{fig:rank} is the detailed version of Fig.~\ref{fig:ranking-prior} (right) in the original paper.

\newpage
\section{Execution Times and Computing Infrastructure}
Table~\ref{tab:exec_times} provides an overview of the execution times over the different selection strategies and datasets. The execution times are averaged over 100 repeated runs each with a maximum number of 200 instance selections. A single execution time entry indicates the average time in seconds to select a single instance for a given dataset and selection strategy. The execution times are primarily depended on the number of instances but also on aspects like the number of features and classes as calculations might become more complex.

All experiments were run on an heterogeneous computer cluster which might lead to irregular results as the speed between the cluster nodes vary. 

\begin{table}[h!]
    \centering
    \small
    \begin{tabular}{lc|cccccccc}
        \toprule
        dataset & instances & xPAL & PAL & OPT & US & ALCE & QBC & EER & RAND \\
         &  & $(\boldsymbol{\alpha}=\mathbf{10^{-3}})$ &  &  &  &  &  &  &  \\
        \midrule
        iris (v1) & 150 & 0.0763 & 0.0089 & 0.0124 & 0.0009 & 0.0034 & 0.0329 & 0.0961 & 0.0001 \\
        wine (v1) & 178 & 0.0824 & 0.0105 & 0.0160 & 0.0010 & 0.0038 & 0.0349 & 0.1184 & 0.0001 \\
        parkinsons (v1) & 195 & 0.0726 & 0.0101 & 0.0234 & 0.0020 & 0.0029 & 0.0354 & 0.0994 & 0.0002 \\
        prnn\_crabs (v1) & 200 & 0.0717 & 0.0063 & 0.0193 & 0.0010 & 0.0030 & 0.0349 & 0.0919 & 0.0001 \\
        sonar (v1) & 208 & 0.0865 & 0.0137 & 0.0459 & 0.0018 & 0.0032 & 0.0365 & 0.1631 & 0.0002 \\
        seismic-bumps (v1) & 210 & 0.0941 & 0.0113 & 0.0179 & 0.0012 & 0.0042 & 0.0342 & 0.0305 & 0.0001 \\
        seeds (v1) & 210 & 0.0991 & 0.0126 & 0.0186 & 0.0010 & 0.0041 & 0.0352 & 0.1459 & 0.0001 \\
        glass (v1) & 214 & 0.3196 & 0.0616 & 0.0193 & 0.0011 & 0.0067 & 0.0355 & 0.2980 & 0.0001 \\
        vertebra-column (v1) & 310 & 0.1737 & 0.0189 & 0.0308 & 0.0013 & 0.0056 & 0.0369 & 0.2515 & 0.0001 \\
        ecoli (v1) & 336 & 0.5243 & 0.2307 & 0.0466 & 0.0028 & 0.0064 & 0.0388 & 0.8007 & 0.0002 \\
        ionosphere (v1) & 351 & 0.2497 & 0.0525 & 0.1307 & 0.0095 & 0.0046 & 0.0404 & 0.7586 & 0.0002 \\
        user-knowledge (v1) & 403 & 0.4468 & 0.0858 & 0.0524 & 0.0015 & 0.0094 & 0.0425 & 0.7438 & 0.0001 \\
        chscase\_vine2 (v2) & 468 & 0.2468 & 0.0182 & 0.0730 & 0.0017 & 0.0046 & 0.0457 & 0.4156 & 0.0001 \\
        kc2 (v1) & 522 & 0.4760 & 0.1127 & 0.2549 & 0.0173 & 0.0053 & 0.0491 & 2.9201 & 0.0003 \\
        wdbc (v1) & 569 & 0.4808 & 0.1655 & 0.3257 & 0.0258 & 0.0054 & 0.0523 & 4.8052 & 0.0005 \\
        balance-scale (v1) & 625 & 0.6702 & 0.0664 & 0.1512 & 0.0045 & 0.0074 & 0.0531 & 0.4069 & 0.0003 \\
        blood-transfusion-s. (v1) & 748 & 0.8058 & 0.0801 & 0.3079 & 0.0128 & 0.0056 & 0.0574 & 1.3400 & 0.0006 \\
        diabetes (v1) & 768 & 0.8174 & 0.1688 & 0.4503 & 0.0287 & 0.0059 & 0.0594 & 8.3075 & 0.0009 \\
        vehicle (v1) & 846 & 2.2947 & 0.5192 & 1.9738 & 0.0384 & 0.0122 & 0.0664 & 35.5409 & 0.0010 \\
        qsar-biodeg (v1) & 1055 & 1.9238 & 0.3374 & 1.2020 & 0.0712 & 0.0077 & 0.0789 & 38.2082 & 0.0016 \\
        banknote-auth. (v1) & 1372 & 1.9130 & 0.3116 & 5.4191 & 0.0996 & 0.0085 & 0.0868 & 69.2271 & 0.0023 \\
        steel-plates-fault (v1) & 1941 & 5.8124 & 0.6423 & 26.3854 & 0.2050 & 0.0118 & 0.1242 & 38.3739 & 0.0022 \\
        \bottomrule
    \end{tabular} 
    \caption{Execution times in seconds for one single instance averaged over all repetitions and acquisitions.}
    \label{tab:exec_times}
\end{table}



\end{document}